%% file: main.tex
\def\isarxiv{1} %%% for icml submission version, we comment this line

\ifdefined\isarxiv
\documentclass[11pt]{article}

\usepackage[numbers]{natbib}

\else

\documentclass[letterpaper]{article} % DO NOT CHANGE THIS
\usepackage[submission]{aaai24}  

\fi

\usepackage{amsmath}
\usepackage{amsthm}
\usepackage{amssymb}
\usepackage{xcolor}
\usepackage{algpseudocode}
\ifdefined\isarxiv

\usepackage{algorithm}
\usepackage{subfig}

\usepackage{graphicx}
\usepackage{grffile}%%% AAAI doesn't allow this
\usepackage{wrapfig,epsfig}
\usepackage{url}

\usepackage{epstopdf}

\usepackage{bbm} %%% AAAI doesn't allow this
\usepackage{dsfont}

\else

\usepackage{times}  % DO NOT CHANGE THIS
\usepackage{helvet}  % DO NOT CHANGE THIS
\usepackage{courier}  % DO NOT CHANGE THIS
\usepackage[hyphens]{url}  % DO NOT CHANGE THIS
\usepackage{graphicx} % DO NOT CHANGE THIS
\usepackage{natbib}  % DO NOT CHANGE THIS AND DO NOT ADD ANY OPTIONS TO IT
\usepackage{caption} % DO NOT CHANGE THIS AND DO NOT ADD ANY OPTIONS TO IT
\usepackage{algorithm}
\usepackage{newfloat}
\usepackage{listings}
\DeclareCaptionStyle{ruled}{labelfont=normalfont,labelsep=colon,strut=off} % DO NOT CHANGE THIS
\floatstyle{ruled}
\newfloat{listing}{tb}{lst}{}
\floatname{listing}{Listing}
\fi 

\ifdefined\isarxiv
\allowdisplaybreaks

\usepackage{tikz}
\usepackage{hyperref}  %%% arxiv don't allow this.
\hypersetup{colorlinks=true,citecolor=blue,linkcolor=blue} %%% Zhao : maybe we should comment this in submission.
\usetikzlibrary{arrows}
\usepackage[margin=1in]{geometry}

\else

\fi
\newtheorem{theorem}{Theorem}%[section]
\newtheorem{lemma}[theorem]{Lemma}
\newtheorem{definition}[theorem]{Definition}

\newtheorem{assumption}[theorem]{Assumption}

\newtheorem{fact}[theorem]{Fact}

\newcommand{\wt}{\widetilde}

\newcommand{\R}{\mathbb{R}}

\newcommand{\new}{\mathrm{new}}
\renewcommand{\d}{\mathrm{d}}

\DeclareMathOperator{\poly}{poly}

\DeclareMathOperator{\diag}{diag}

\makeatletter
\newcommand*{\RN}[1]{\expandafter\@slowromancap\romannumeral #1@}
\makeatother

\ifdefined\isarxiv

\else
\newcommand{\texorpdfstring}[1]{{#1}} 
\fi

\usepackage{lineno}

\begin{document}

\ifdefined\isarxiv

\date{}

\title{Convergence of Two-Layer Regression with Nonlinear Units}
\author{
Yichuan Deng\thanks{\texttt{ethandeng02@gmail.com}. The University of Washington.}
\and
Zhao Song\thanks{\texttt{zsong@adobe.com}. Adobe Research.}
\and
Shenghao Xie\thanks{\texttt{xsh1302@gmail.com}. The Chinese University of Hong Kong, Shenzhen.}
}

\else

\title{Convergence of Two-Layer Regression with Nonlinear Units} 
\maketitle 
\fi

\ifdefined\isarxiv
\begin{titlepage}
  \maketitle
  \begin{abstract}
\input{abstract}

  \end{abstract}
  \thispagestyle{empty}
\end{titlepage}

{\hypersetup{linkcolor=black}
%\tableofcontents
}
\newpage

\else

\begin{abstract}
\input{abstract}
\end{abstract}

\fi

\input{intro} %%% Section 1. Introduction

\input{preli_short}

\input{model_overvie}

\input{cov_variable}

\input{cov_loss}

\input{related_work}

\input{conclusion}

\ifdefined\isarxiv
%\section*{Acknowledgments}
\bibliographystyle{alpha}
\bibliography{ref}
\else
\bibliography{ref}

\fi

\newpage
\onecolumn
\appendix

\section*{Appendix}
\input{preli}

\input{gradient}

\input{hessian}

\input{bound}

\input{lip}

\input{psd}

\input{newton}

\input{sophia_related}

%%%% Cut-line between first 10 pages and appendix

%%% some writing rules

%% Writing rule for creating tags.
%% Tags :
%% Theorem    \ref{thm:bla_bla}
%% Lemma      \ref{lem:bla_bla}
%% Claim      \ref{cla:bla_bla}
%% Corollary  \ref{cor:bla_bla}
%% Fact       \ref{fac:bla_bla}
%% Definition \ref{def:bla_bla}
%% Section    \ref{sec:bla_bla}
%% Subsection \ref{sub:bla_bla}
%% Equation   \ref{eq:bla_bla}

\end{document}

%% file: abstract.tex
Large language models (LLMs), such as ChatGPT and GPT4, have shown outstanding performance in many human life task. Attention computation plays an important role in training LLMs.

Softmax unit and ReLU unit are the key structure in attention computation. Inspired by them, we put forward a softmax ReLU regression problem. Generally speaking, our goal is to find an optimal solution to the regression problem involving the ReLU unit. 

% Formally speaking, given matrices $A_1 \in \R^{n \times d}, A_2 \in \R^{m \times n}$, vectors $b \in \R^n$, $w \in \R^m$, let $R \in \R$ be a scalar. Our goal is to find the optimal point of the following optimization question
% $
%      \min_{x\in \{ \|x\|_2 \leq R, x \in \R^d \} }  \| (\langle \exp(A_2 \cdot \phi(A_1 \cdot x)), {\bf 1}_m \rangle)^{-1} \cdot \exp(A_2 ~\cdot \phi(A_1 \cdot x))- b \|_2^2  
% $.

In this work, we calculate a close form representation for the Hessian of the loss function. Under certain assumptions, we prove the Lipschitz continuous and the PSDness of the Hessian. Then, we introduce an greedy algorithm based on approximate Newton method, which converges in the sense of the distance to optimal solution. Last, We relax the Lipschitz condition and prove the convergence in the sense of loss value.

%\Zhao{Please try to expand the abstract here.}

%% file: intro.tex
\section{Introduction}
{\bf Large Language Models (LLMs).} The development of Large Language Models (LLMs) \cite{vsp+17, rns+18, dclt18, rwc+19, bmr+20, cnd+22, zrg+22, cha22, o23} represents a groundbreaking milestone in the realm of artificial intelligence and natural language processing. LLMs are a type of deep learning model that has significantly advanced the capabilities of language understanding and generation. This remarkable progress can be attributed to several key factors: the availability of vast amounts of text data, advancements in computing power, and breakthroughs in deep learning algorithms. The development of LLMs can be traced back to the early efforts of language modeling with neural networks. As researchers experimented with different architectures and techniques, LLMs gradually evolved into more sophisticated models capable of handling complex linguistic patterns and generating coherent text. These models have demonstrated extraordinary achievements in a wide range of applications, including machine translation \cite{hwl21}, sentiment analysis \cite{uas20}, text generation \cite{zrg+22,cha22,o23}, and question-answering systems \cite{cha22, o23}, revolutionizing the way we interact with language-based technologies and opening up new possibilities for the future of AI-driven communication. As the scale of language models continues to rise, some recent works are focused on efficient training of LLMs \cite{mwy+23, pmxa23, mgn+23}.

\paragraph{Non-linear Units.}

We consider the canonical softmax unit and ReLU unit in this work. 

The softmax function is a fundamental tool in data science and computer science. It was first applied in statistical mechanism \cite{b68, g02} and decision theory \cite{tl59}. Then, it became a widely used activation function in machine learning, having significant influence on the development of LLMs \cite{b90,  b06, bsf06, gbc16, lbh15}. Taking a real-valued vector as input, the softmax function transforms it into a probabilistic distribution summing up to one. This property makes it suitable for multi-class classification tasks. It now plays an important role in the formulation of transformer models. We present its renowned definition as follows.
\begin{definition}[Softmax function]
Suppose $x \in \R^d$, let $\exp(x) \in \R^d$ be the vector that $i$-th entry is $\exp(x_i)$. Let ${\bf 1}_d$ denote the length-$d$ vector where all entries are ones.
The Softmax function is defined as
\begin{align*}
    \mathrm{Softmax}(x) = (\langle \exp(x), {\bf 1}_d \rangle)^{-1} \cdot \exp(x)
\end{align*}
\end{definition}

%\paragraph{ReLU function}

%\mbox{}

The Rectified Linear Unit (ReLU) function was introduced by in the context of visual feature extraction in hierarchical neural networks \cite{f69, f80}. Later, it was discussed that the utilization of this approach was supported by robust biological underpinnings and sound mathematical rationales \cite{hsm+00, hs00}. In 2011, researchers discovered that this technique facilitated more effective training of deeper networks \cite{gbb11} when compared to the activation functions commonly employed before that time, such as the logistic sigmoid (which draws inspiration from probability theory, as seen in logistic regression) and its more practical counterpart, the hyperbolic tangent \cite{lbom02}. The widely-used definition of ReLU is as follows. 
\begin{definition}[ReLU function]
Let $z \in \R$. 
    The ReLU function is defined as
    \begin{align*}
        \mathrm{ReLU}(z) := \max\{0, z\}. 
    \end{align*}
\end{definition}
We use the vectorized version of ReLU function in our regression model.
\begin{definition}[Vectorized version]
Let $x \in \R^d$, we defined $\phi(x)$ to be a vector which has its $i$-th entry being $\mathrm{ReLU}(x_i)$.
\end{definition}
ReLU is now a fundamental element extensively utilized in Large Language Models (LLMs) and various deep learning architectures \cite{LL18, DZPS19, azls19a, azls19b, JT20, LSS+20, HLSY21, ZPD+20, BPSW21, SZZ21, Zha22}. Introduced as a remedy to address the vanishing gradient problem, ReLU has gained immense popularity due to its simplicity and effectiveness. This activation function ensures that the output of a neuron is non-linear, allowing neural networks to learn complex patterns and representations. The ReLU function replaces all negative values in the input with zero, while preserving positive values unchanged, making it computationally efficient and easy to implement. Its ability to prevent vanishing gradients facilitates faster and more efficient training of LLMs, enabling them to achieve remarkable performance in various natural language processing tasks.

\paragraph{Attention-motivated regression.}

Attention computation is a key aspect when constructing LLMs \cite{vsp+17, rns+18, dclt18,rwc+19, bmr+20, cha22}. The attention mechanism allows the model to assign different weights to different elements of the input sequence, enabling it to capture important information and make accurate inference. Self-attention is a commonly-used attention scheme in transformer model. Model with self-attention unit can handle long sequences effectively, capture contextual information, and generate more coherent outputs. Several works demonstrated the benefit of self-attention in attending In-Context learning \cite{szks21, gtlv22, zfb23, wzw23}. We present the definition of softmax self-attention as follows.
\begin{definition} [Self-attention module]
Given matrices $Q, K, V \in \R^{n \times d}$, (which represent the query, key and value weight matrices). We have
\begin{align*}
    \mathrm{Attention} (Q, K, V) := \diag(A \cdot {\bf 1}_n)^{-1} A V
\end{align*}
where $A := \exp(QK^\top)$.
\end{definition}
Previous work put forward softmax regression based on the definition of self-attention.
\begin{definition}[Softmax regression \cite{dls23}]
    Let $A \in \R^{n \times d}$ be a matrices, let $b \in \R^n$ be a vector. The loss function is defined as \begin{align*} L(x) := \| (\langle \exp(A x)), {\bf 1}_n \rangle)^{-1}  \exp(A x))- b \|_2^2\end{align*}
\end{definition}

Inspired by the importance ReLU function, we consider a two-layer regression problem involved it. In this work, we study the following regression model
\begin{definition}[Two-Layer Regression with Nonlinear Units] \label{def:overview_brief}
    Let $A_1 \in \R^{n \times d}, A_2 \in \R^{m \times n}$ be two matrices, let $b \in \R^m$ be a vector, let $R \in \R$ be a scalar. Our goal is to minimize the objective function \begin{align*}\| (\langle \exp(A_2  \phi(A_1 x)), {\bf 1}_m \rangle)^{-1}  \exp(A_2    \phi(A_1 x))- b \|_2^2\end{align*} under conditions  $\min_{x\in \{ \|x\|_2 \leq R, x \in \R^d \} }  $.
\end{definition}
where its first layer is activated by a ReLU function, the second layer is by softmax function. 

The core idea of this model formulation echos with \cite{dls23, lsz23, gsy23, gsx23}.

This paper gives a convergence analysis. We verify the feasibility of applying a Hessian-based greedy algorithm to find the optimal point of above regression problem. Our contribution can be summarized as follows
\begin{itemize}
    \item We calculate the Hessian of the loss function and reformulate it into a close-form.
    \item We prove the Lipschitz continuous and Positive Semi-definite Definite (PSD) lower bound for Hessian of loss function. We add a regularization function to the loss function so that its Hessian is Positive Definite (PD). Then, we prove the convergence of an approximation Newton method.
    \item To relax an assumption on ReLU function, we present a analysis framework which replaces the Lipschitz of Hessian by a milder condition. Then we prove the convergence of Newton method in the sense of descending loss.
\end{itemize}

\paragraph{Roadmap.} In Section~\ref{sec:preli:informal}, we present some notations used in this paper. We introduce our model in Section~\ref{sec:model_overview}. We study the convergence of an approximation Newton method to solve the regression problem in Section~\ref{sec:cov_variable:informal}. We apply a novel analysis framework in Section~\ref{sec:cov_loss:informal} to prove the convergence in loss value. We discuss related work on similar topic in Section~\ref{sec:related_work}. We state our conclusion and potential future research area in Section~\ref{sec:conclusion}.

%% file: preli_short.tex
\section{Preliminary} \label{sec:preli:informal}
We define some notations in this section.
%%%Zhao: The correct label: {sec:preli:notation}

Let $z$ be a scalar, we define
$
 {\bf 1}[ z > 0 ] :=   \begin{cases}
        0, & z \leq 0 \\
        1, & z > 0
    \end{cases}
$. Let $x \in \R^n$ be a vector, we use ${\bf 1}[x]$ to denote a vector $y \in \R^n$ such that $y_i := {\bf 1}[x_i > 0]$. Let $x \in \R^n$ be a vector, $\exp(x) \in \R^n$ denotes the vector that $\exp(x)_i = \exp(x_i)$. Let $f: \R^d \to \R^n $, let $t$ be a scalar, $\frac{\d f}{\d t} \in \R^n$ denotes a vector whose $i$-th entry is $\frac{\d f_i}{\d t}$. Let $x, y$ be two vectors $\in \R^n$, let $\langle x,y \rangle$ be $\sum_{i=1}^n x_i \cdot y_i$. Let $x, y$ be two vectors $\in \R^n$, let $x \circ y := z \in \R^n$ where $z_i = x_i \cdot y_i$. Let ${\bf 1}_n$ denotes a vector with n entries and all of them are 1. Let ${\bf 0}_n$ denotes a vector with n entries and all of them are 0. Let ${\bf I}_n$ denotes a ${n \times n}$ identity matrix. Let $[n] := \{1, 2, \cdots n \}$.

%% file: model_overvie.tex
\section{Model Overview} \label{sec:model_overview}
This section provides a formal statement of our regression model. In section~\ref{sec:model}, we provide a model overview. We present the Hessian of $L(x)$ in Section~\ref{sec:hessian}.

\subsection{Our Model} \label{sec:model}
Our model consists of two parts:
\begin{itemize}
    \item $L(x)$: the loss function inspired by softmax-ReLU regression
    \item $R(x)$: the regularization term $R(x)$ which ensures the strongly convexity of the objective function
\end{itemize}
The following Definition~\ref{def:overview} is a detailed restatement of Definition~\ref{def:overview_brief}, together with introduction of basic functions that we discuss in this paper.

\begin{definition} \label{def:overview}
Let $A_1 \in \R^{n \times d}, A_2 \in \R^{m \times n}, b, w \in \R^m, R \in \R$. We consider the following regression model
\begin{align*}
    \min_{x \in \{ \|x\|_2 \leq R, x \in \R^d \} } L_{\mathrm{reg}}(x) := L(x) + R(x)
\end{align*}
which satisfies
\begin{itemize}
    \item $L(x) := \frac{1}{2} \cdot \| (\langle \exp(A_2 \cdot \phi(A_1 \cdot x)), {\bf 1}_m \rangle)^{-1} \cdot \exp(A_2 \cdot \phi(A_1 \cdot x))- b \|_2^2$
    \item $R(x) := \frac{1}{2} \cdot \| WCx \|_2^2$
    \item $C := A_2 \cdot \diag({\bf 1}[A_1 x]) \cdot A_1$
    \item $W := \diag(w)$
\end{itemize}
\end{definition}

\subsection{Hessian of Loss function} \label{sec:hessian}
To simplify the calculation of Hessian, we define several auxiliary functions and calculate their Hessian separately (see in Section~\ref{sec:hess}). Then we combine them by chain rule and linear algebra lemmas. To analyze the complicated expression of Hessian, we exert a close-form decomposition technique to give a reformulation \cite{ans00, hjs+22}. The formula of each second-order derivative is as follow
\begin{align*}
    \frac{\d^2 L(x)}{\d x_i \d x_j} = A_{1,i}^\top \diag({\bf 1} [A_1 x]) A_2^\top  B(x) A_2 \diag({\bf 1}[A_1 x]) A_{1,j}
\end{align*}
where $A_{1,i}$ is defined to be the $i$-th column of matrix $A_1$.

In above, $B(x)$ is a $m \times m$ symmetric matrix which is independent of the entry index ($i$, $j$). The explicit expression of $B(x)$ can be found in Section~\ref{sec:hess_L_decompose}. It follows that we can decompose the Hessian into multiplication of matrices.

\begin{lemma} [Hessian of loss function] \label{lem:decomposition_L:informal}
Let $C := A_2 \cdot \diag({\bf 1}[A_1 x]) \cdot A_1$. Let $W^2$ be the matrix which each diagonal entry is the square of that of $W$, then we have
\begin{itemize}
    \item $\nabla^2 L(x) = C^\top B(x) C$
    \item $\nabla^2 R(x) = C^\top W^2 C$
    \item $\nabla^2 L_\mathrm{reg}(x) = C^\top (B(x) + W^2) C$
\end{itemize}

\end{lemma}
 Notice that $B(x)$ have both PSD lower bound and upper bound. Therefore, we can straightforwardly improve it to a PD matrix by adding the regularization term $W$.

%% file: cov_variable.tex
\section{Convergence of distance to optimal point} \label{sec:cov_variable:informal}
In this section, we introduce an approximation Newton method to solve the softmax-ReLU regression. We first prove the objective function has Lipschitz continuous Hessian (see Section~\ref{sec:lip_hess:informal}) and it is strongly convex (see Section~\ref{sec:pd_hess:informal}), then we prove the convergence of the algorithm (see Section~\ref{sec:main_result:informal}).

\subsection{Hessian is Lipschitz} \label{sec:lip_hess:informal}
In this section, we compute the Lipschitz constant for Hessian. The core idea of this proof echos with previous analysis on recurrent neural network (RNN) and transformer-inspired regression models \cite{azls19a,azls19b,dls23,gsx23}. 

Now, we introduce two assumptions to bound the Hessian. First, we assume the upper bound for each parameter.
\begin{assumption}[Bounded parameters] \label{ass:bounded_parameters}
We have $\| A_1 \| \leq R$, $\| A_2 \| \leq R$, $\|b\|_2 \leq 1$, $R > 4$.
\end{assumption}

Next, we assume that the ReLU function doesn't change state for all $x$ in the set of decision variable. Although this assumption is strong, it is necessary because the term ${\bf 1}[A_1 x]$ is unstable. There is a high probability that ${\bf 1}[A_1 x]$ and ${\bf 1}[A_1 y]$ differs by at least one entry even when $x$ and $y$ become relatively close. We relax this assumption later in a proof of convergence which does not use the Lipschitz condition of Hessian.

\begin{assumption}[Fixed ReLU state] \label{ass:fixed_ReLU}
For all $ x \in \{ \|x\|_2 \leq R, x \in \R^d \}$, ${\bf 1}[A_1 x]$ has the same value.
\end{assumption}

Next we present the Lipschitz for Hessian.
\begin{lemma} [Lipschitz of $\nabla^2 L(x)$, informal version of Lemma~\ref{lem:lipschitz_hessian_L}] \label{lem:lipschitz_hessian_L:informal}
If Assumption~\ref{ass:bounded_parameters} and Assumption~\ref{ass:fixed_ReLU} hold, we have $\| \nabla^2 L(x) - \nabla^2 L(y) \| \leq M \cdot \| x - y \|_2$, where $M = n^{1.5} m^{1.5} \sqrt{d} \exp(5 R^3)$.
\begin{proof}
See Section~\ref{sec:bound} for bounds on basic functions. See Section~\ref{sec:lipschitz_basic} for Lipschitz of basic functions. See Section~\ref{sec:lipschitz_hess} for Lipschitz of Hessian.
\end{proof}
\end{lemma}
Notice that, under Assumption~\ref{ass:fixed_ReLU}, $\nabla^2 R(x)$ is fixed for all $x \in \{ \|x\|_2 \leq R, x \in \R^d \}$. Therefore, the Lipschitz constant of $\nabla^2 L_\mathrm{reg}(x)$ is the same as $\nabla^2 L(x)$.

\subsection{Hessian is PD} \label{sec:pd_hess:informal}
In the convergence analysis of approximation Newton method, PSDness of Hessian is insufficient to generate a proof. The Hessian needs to be PD, hence we verify this property in this section. As we mentioned before, the Hessian can be reformulated to a close-form expression $C^\top B(x) C$ \cite{ans00, hjs+22, dls23,gsy23}. By bounding the symmetric matrix $B(x)$ and adding the regularization term, we prove the PD property. First, let us present a PSD bound for $B(x)$ (see its definition in Lemma~\ref{lem:decomposition_L:informal}).
\begin{lemma} \label{lem:psd_B:informal}
If $\|b\|_2 \leq 1$, we have $-20 {\bf I}_m \preceq B(x) \preceq 20 {\bf I}_m$.
\end{lemma}

To compute a PD lower bound for the Hessian, we need $C$ to have only non-zero singular value, or it is possible that the Hessian has a zero eigenvalue. This condition can be achieved if $C$ has full rank. Therefore, we have following assumption about the structure of $C$.
\begin{assumption}[Rank of $C$] \label{ass:rank_C}
Let $A_1, A_2$ have full rank. Let $n \geq \xi \cdot \max\{m, d\}$, where $\xi > 1$. Let $\| {\bf 1}[A_1 x] \|_1 \geq \theta n$, where $1 > \theta > \frac{1}{\xi}$.
\end{assumption}

Assumption~\ref{ass:rank_C} implies that the Rank of $\diag({\bf 1}[A_1 x]))$, which equals to the number of non-zero entry of it, is bigger than or equal to $\max \{m,d \} $. Then, $\mathrm{Rank}(C) = \min \{m, \mathrm{Rank}(\diag({\bf 1}[A_1 x])), d \} = \min \{m,d \}$. Please note that $C \in \R^{m \times d}$, as a consequence, $C$ has full rank. 

Therefore, $\sigma_{\min}(C)^2 > 0$, where $\sigma_{\min}(C)$ denote the singular value of $C$ that has the smallest absolute value. Then, it is naturally to define $W$ to be such that it makes $L_\mathrm{reg}(x)$ strongly convex. The condition for $W$ is given below in accordance with $B(x)$ and $\sigma_{\min}(C)$.
\begin{assumption}[Condition of regularization term $W$] \label{ass:w}
Let . Let $l > 0$ be a scalar, for all $i \in [m]$, $w_i^2 \geq 20 + \frac{l}{\sigma_{\min}(C)^2}$.
\end{assumption}

With Assumption~\ref{ass:rank_C} and Assumption~\ref{ass:w}, we can easily verify that the Hessian is PD, which implies the strongly convexity of the objective function $L_\mathrm{reg}(x)$. We provide the formal statement as below.

\begin{lemma} [Hessian is PD, informal version of Lemma~\ref{lem:pd_L_reg}] \label{lem:pd:informal}
If the Assumption~\ref{ass:rank_C} and Assumption~\ref{ass:w} are satisfied, we have $\nabla^2 L_{\mathrm{reg}}(x) \succeq l \cdot {\bf I}_d$.
\end{lemma}

\subsection{Convergence analysis} \label{sec:main_result:informal}
In this section, we present the convergence analysis of the algorithm.

Note that it may be hard to explicitly compute the Hessian or its inverse in real-word scenarios. Thus, it is natural to introduce an approximation algorithm for Hessian to precipitate the computation (ex. \cite{cls19, lsz19, Son19, ccly19, Bra20, jswz21, sy21, hjs+22,
gs22, dsw22, syyz22, jlsz23, gsz23, syyz23, bs23, lsz+23}).
Given a PD matrix in close-form, we are able to wield a lemma from \cite{dsw22} to efficiently generate an estimation.
\begin{lemma} [Informal version of Lemma~\ref{lem:approx_hessian}] \label{lem:approx_hessian:informal}
Under the following conditions:
\begin{itemize}
\item Consider a constant precision parameter $\epsilon_0 = 0.01$
\item Let $C$ be a real matrix of size $m \times d$
\item Let $D$ be a positive definite matrix of size $m \times m$
\end{itemize}
there exists an algorithm that can be executed in time
\begin{align}
O(( \mathrm{nnz}(C) + d^\omega)\poly(\log(m/\delta)))
\end{align}
The algorithm outputs a sparse diagonal matrix $\Tilde{D}$ of size $m \times m$, and it satisfies the condition
\begin{align}
(1 - \epsilon_0) C^\top D C \preceq C^\top \Tilde{D} C \preceq (1 + \epsilon_0) C^\top D C
\end{align}
Notice parameter $\omega$ represents the exponent of matrix multiplication, which is recently estimated to be $2.372$ \cite{wil12,lg14,aw21,dwz22}.
\end{lemma}

\paragraph{$Remark$} Recall that $\nabla^2 L_\mathrm{reg}(x) = C^\top (B(x) + W^2) C$. We have $D := B(x)+W^2$ to be PD. Thus, Lemma~\ref{lem:approx_hessian:informal} is well-fitted to our algorithm framework.

Based on above assumptions and lemmas, we provide the convergence theorem for the approximation Newton method. We state the update rule in each iteration as follows.
\begin{definition}[Update rule] \label{def:update_rule:informal}
In each step, we sample an approximation $\wt{H}(x)$ of $H(x) := \nabla^2 L_\mathrm{reg}(x)$. Next, we update the decision value by $x_{t+1} = x_t - \wt{H}(x)^{-1} \cdot \nabla L_\mathrm{reg}(x)$
\end{definition}

We need more assumptions to compute the convergence rate of the algorithm. First, we assume that the initial point is relatively close to the optimal point. Second, we assume the optimal point to have a slightly smaller upper bound, which ensures the norm of decision value is smaller than $R$ in each update.
\begin{assumption}[Initial point] \label{ass:good_initial_pt}
We choose an initial point $x_0$ such that $M \cdot \|x_0 - x^*\|_2 \leq 0.1l$, where $M = m^{1.5} \sqrt{nd} \exp(5 R^3)$.
\end{assumption}
\begin{assumption}[Optimal point] \label{ass:good_optimal_pt}
Let $x^*$ represents the optimal solution, we have $\|x^* \|_2 \leq R - 0.1l/M$ and $\nabla L(x^*) = {\bf 0}_d$.
\end{assumption}

The formal statement of the theorem is listed as below. The correctness of Theorem~\ref{thm:main:informal} is given by the combination of lemmas and assumptions in this section.
\begin{theorem} [Convergence of decision variable, informal version of Theorem~\ref{thm:main}] \label{thm:main:informal}
If Assumption~\ref{ass:bounded_parameters}, Assumption~\ref{ass:fixed_ReLU}, Assumption~\ref{ass:rank_C}, Assumption~\ref{ass:w}, 
Assumption~\ref{ass:good_initial_pt}, and Assumption~\ref{ass:good_optimal_pt} are satisfied,
for any accuracy parameter $\epsilon$ in the range of (0, 0.1) and a failure probability $\delta$ in the range of (0, 0.1), an algorithm (refer to Algorithm~\ref{alg:main_algorithm:informal}) can be employed. This algorithm guarantees, with a probability of at least $1 - \delta$, that it will execute $T = O(\log(|x_0 - x^*|_2 / \epsilon))$ iterations and produce a vector $\Tilde{x} \in \mathbb{R}^d$ satisfying $|\Tilde{x} - x^*|_2 \leq \epsilon$.

The execution time for each iteration is
\begin{align*}
    O(( \mathrm{nnz}(C) + d^\omega)\poly(\log(m/\delta)))
\end{align*}
Here $\omega$ represents the exponent of matrix multiplication, which is currently estimated to be $\omega \approx 2.372$ \cite{wil12,lg14,aw21,dwz22}.
\begin{proof}
 See detailed proof in Theorem~\ref{thm:main}.   
\end{proof}
\end{theorem}

\begin{algorithm}[!ht]\caption{Approximation Newton method}\label{alg:main_algorithm:informal}
\begin{algorithmic}[1]
\Procedure{IterativeSoftmaxReLURegression}{$x_0$} \Comment{Theorem~\ref{thm:main:informal}} 
    \State
    We choose an initial point $x_0$ satifying Assumption~\ref{ass:good_initial_pt}.
    \State
    Let $T = \eta^{-1} \log(\frac{\| x_0 - x^* \|_2}{\epsilon})$ be the number of iterations.
    \For{$t=0 \rightarrow T$}
        \State
        $g \leftarrow \nabla L_\mathrm{reg}(x)$
        \State
        $H \leftarrow \nabla^2 L_\mathrm{reg}(x)$
        \State
        $\wt{H} \leftarrow \mathrm{subsample}(H)$
        \Comment{Lemma~\ref{lem:approx_hessian:informal}}
        \State
        $x_{t+1} \leftarrow x_t - \eta \cdot \wt{H}^{-1} g$
    \EndFor
    \State
    $\Tilde{x} \leftarrow x_{T+1}$
    \State
    \Return $\Tilde{x}$
\EndProcedure
\end{algorithmic}
\end{algorithm}

%% file: cov_loss.tex
\section{Convergence of Loss value} \label{sec:cov_loss:informal}
To relax Assumption~\ref{ass:fixed_ReLU}, we present a method from \cite{llh+23}. We modified their lemmas and proofs to a generalized version (see Section~\ref{sec:preli_sophia:informal}). Then, we utilize them to prove the convergence of Newton method in the sense of shrinking loss value (see Section~\ref{sec:cov_sophia:informal}).

In this chapter, we also omit the condition that $\|x\|_2 \leq R$ since it is not necessary in the proof. Therefore, we are considering a optimization problem which is $\min_{x \in \R^d} L_\mathrm{reg}(x)$.

\subsection{Preliminary} \label{sec:preli_sophia:informal}
Now, we consider $L(x)$ as a general loss function, and we aim to minimize it. We have following definitions:
\begin{itemize}
    \item Let $x^*$ denote the optimal solution
    \item Let $L_{\min} := L(x^*)$
\end{itemize}
We verify the convergence of Newton Method in this section based on following Assumption~\ref{ass:L:informal} and Assumption~\ref{ass:lip_hess:informal}.
\begin{assumption} \label{ass:L:informal}
$L(x)$ is twice continuously differentiable almost everywhere and strictly convex.
\end{assumption}
Assumption~\ref{ass:L:informal} focuses on the smoothness and convexity of the loss function. The smoothness ensures that we can write $L(x) - L_{\min}$ as an integration. Then, it can be bounded by bounding its gradient and Hessian. The convexity is used to prove Lemma~\ref{lem:ode:informal}.

\begin{assumption} \label{ass:lip_hess:informal}
Let $N \in \R^d$ be a scalar, $\forall x, y \in \R^d$, we have $\| \nabla^2 L(x)^{-1} \nabla^2 L(y) \| \leq N$
\end{assumption}

Assumption~\ref{ass:lip_hess:informal} still considers about the local situation of Hessian. The difference is that it replaces a Lipschitz bound by a constant bound, which makes the constrain on the loss function looser. A straightforward corollary from Assumption~\ref{ass:lip_hess:informal} is as below
\begin{align} \label{eq:psd_bound:informal}
    N(\nabla^2 L(y))^{-1} \succeq \nabla^2 L(x)^{-1} \succeq \frac{1}{N}(\nabla^2 L(y))^{-1}
\end{align}
which means that the inverse of $\nabla^2 L(x)$ can be upper bounded and lower bounded by that of an arbitrary variable.

If Assumption~\ref{ass:L:informal} and Assumption~\ref{ass:lip_hess:informal} hold, we have following lemmas.

\begin{lemma}[Informal version of Lemma~\ref{lem:ode}] \label{lem:ode:informal}
For any $z \in \R^d$, the following differential equation system has at least one solution on interval $[0, 1]$:
\begin{align} \label{eq:ode:informal}
    \frac{\d x(t)}{\d t} = & ~ -(\nabla^2 L(x(t)))^{-1} \nabla L(z), \\
    x(0) = & ~ z \notag
\end{align}
and the solution satisfies that $\nabla L(x(t)) = (1 - t) \nabla L(z)$ and $x(1) = x^*$.
\end{lemma}
Lemma~\ref{lem:ode:informal} introduces a tool from Ordinary Differential Equation (ODE) which gives us a way to relate decision variable $z$ and optimal point $x^*$. Let $x(t)$ be the solution of Eq.~\eqref{eq:ode:informal} on interval $[0,1]$. We have $z, x^*$ to be its two endpoints, i.e., $x(0) = z$, $x(1) = x^*$.

As a result of Lemma~\ref{lem:ode:informal}, we are able to use an integral to represent the value of $L(z) - L_{\min}$, which equals $\int_0^1 (1-t)(\nabla L(z))^\top (\nabla^2 L(x(t)))^{-1} \nabla L(z) \d t$. Together with Eq.~\eqref{eq:psd_bound:informal}, the distance between $L(z)$ and $ L_{\min}$ can be controlled by $\nabla L(z)^\top (\nabla^2 L(z))^{-1} \nabla L(z)$. Formally speaking, we have following lemma.

\begin{lemma}[Informal version of Lemma~\ref{lem:bound_hess}] \label{lem:bound_hessian:informal}
For any $ z \in \R^d$, it holds that
$\frac{2}{N}(L(z) - L_{\min}) \leq \nabla L(z)^\top (\nabla^2 L(z))^{-1} \nabla L(z) \leq 2N (L(z) - L_{\min})$
\begin{proof}
See detailed proof in Section~\ref{sec:preli_sophia}.
\end{proof}
\end{lemma}

We state the update rule for Newton method as follows
\begin{definition}[Update rule] \label{def:update_rule_sophia:informal}
    We define the update rule as $x_{\new} = x - \eta \cdot (\nabla^2 L(x))^{-1} \nabla L(x)$.
\end{definition}
Lemma~\ref{lem:bound_hessian:informal} connects the error of loss value (i.e., $L(z) - L_{\min}$) to the change in each updation, enabling us to derive the shrinking lemma. By an integral decomposition of $L(x)$, we derive the following descent lemma. We omit the proof here for space limitation.

\begin{lemma} [Informal version of Lemma~\ref{lem:descent}] \label{lem:descent:informal} 
Using the update rule as Definition~\ref{def:update_rule_sophia:informal}, we have
\begin{align*}
    L(x_{\new}) - L(x) \leq -(\eta - \frac{N}{2} \eta^2) \cdot \nabla L(x) (\nabla^2 L(x))^{-1} \nabla L(x)
\end{align*}
\begin{proof}
See detailed proof in Section~\ref{sec:preli_sophia}.
\end{proof}
\end{lemma}

Combining Lemma~\ref{lem:bound_hess:informal} and Lemma~\ref{lem:descent:informal}, we have the following shrinking lemma. By setting a proper number of iteration $T$, we can prove convergence.
\begin{lemma} [Informal version of Lemma~\ref{lem:shrinking_loss}] \label{lem:shrinking_loss:informal}
Let the update rule be as Definition~\ref{def:update_rule_sophia:informal}. Then, for $t, T \in \mathbb{N}$, $t < T$, we have
\begin{align*}
    L(x_T) - L_{\min} \leq (\eta^2 - \frac{2}{N} \eta + 1)^{T-t} \cdot (L(x_t) - L_{\min})
\end{align*}
If we take $\eta = \frac{1}{N}$, then
\begin{align*}
    L(x_T) - L_{\min} \leq (1 - \frac{1}{N^2})^{T-t} \cdot (L(x_t) - L_{\min})
\end{align*}

\begin{proof}
See detailed proof in Section~\ref{sec:preli_sophia}.
\end{proof}
\end{lemma}

\subsection{Convergence analysis} \label{sec:cov_sophia:informal}
In this section, we apply the shrinking lemma to prove the convergence when the loss function is our $L_\mathrm{reg}(x)$. We observe that $L_\mathrm{reg}(x)$ satisfy both above assumptions, therefore it is well-fitted to above analysis framework. 

Notice that $L_\mathrm{reg}$ is twice continuously differentiable everywhere except at $x = 
{\bf 0}_d$ since $A_1$ has full rank. Furthermore, it is strictly convex by Lemma~\ref{lem:pd:informal}, and hence Assumption~\ref{ass:L:informal} is achieved. 

Next, recall that $\nabla^2 L_\mathrm{reg}(x)$ has a PD lower bound, so its inverse has an upper bound. The term $\|\nabla^2 L_\mathrm{reg}(x)\|$ can be upper bounded using assumed upper bound on each parameter (see Assumption~\ref{ass:bounded_parameters}). Therefore, we have the lemma below, implying that Assumption~\ref{ass:lip_hess:informal} is satisfied.

\begin{lemma}[Informal version of Lemma~\ref{lem:bound_hess}] \label{lem:bound_hess:informal}
If Assumption~\ref{ass:bounded_parameters}, Assumption~\ref{ass:rank_C}, and Assumption~\ref{ass:w} hold. Let $f \in \R$ be a scalar such that $w_i^2 \leq f$ for all $i \in [n]$. Then for $\forall x, y \in \R^d$, we have
\begin{align*}
    \| \nabla^2 L_\mathrm{reg}(x)^{-1} \nabla^2 L_\mathrm{reg}(y)\| \leq \frac{1}{l} \cdot R^4 \cdot (16 + f)
\end{align*}
\end{lemma}

Last, we propose a convergence theorem for Newton method with loss function being $L_\mathrm{reg}(x)$.
\begin{theorem} [Convergence of loss, informal version of Theorem~\ref{lem:cov_L}] \label{thm:cov_L:informal} 
If the following conditions hold
\begin{itemize}
    \item Assumption~\ref{ass:bounded_parameters}, Assumption~\ref{ass:rank_C}, and Assumption~\ref{ass:w} are satisfied
    \item Let $x_0$ be the initial point of the algorithm. Let $x^*$ be the optimal point of $\min_{x \in \R^d} L_\mathrm{reg}(x)$, we define $L_{\min} := L_\mathrm{reg}(x^*)$
    \item we define the update rule as $x_{t+1} = x_t - \eta \cdot (\nabla^2 L_\mathrm{reg}(x_t))^{-1} \nabla L_\mathrm{reg}(x_t)$
    \item Let $N = \frac{1}{l}\cdot R^4 \cdot (16+f)$, we take $\eta = \frac{1}{N}$
\end{itemize}   
Then for any accuracy parameter $\epsilon > 0$, the algorithm outputs a vector $\wt{x} \in \R^d$ with $L_\mathrm{reg}(\wt{x}) - L_{\min} \leq \epsilon$ in $T = O(  N^2 \log( (L_{\mathrm{reg}}(x_0) - L_{\min}) / \epsilon) )$ iterations.

\begin{proof}
 See detailed discussion in Theorem~\ref{lem:cov_L}.

\end{proof}
\end{theorem}

%% file: related_work.tex
\section{Related Work} \label{sec:related_work}
{\bf Transformer theory.} 
Since the explosion of LLM, many research have focused on analyzing the learning ability of LLM, adding to the theoretical background of transformer theory. One important topic is how transformer achieves In-Context learning. \cite{szks21} proposed a way to understand how a single-head attention module learns in a seq2seq translation experiment. They defined a model property: knowing to translate individual word (KTIW), which means the model knows a word translates to another. They claimed that KTIW is dominated in the learning of attention, since it can be learned from word co-occurence statistics before the attention is learned, but the opposite direction is not available. \cite{gtlv22} empirically showed that transformer can in-context learn linear function classes, with performence close to optimal least square estimator. They testified its ability that, when some specific functions are fed to the model, transformer can infer information about most of the functions in the class. \cite{asa+22} established a implicit relation between in-context learner based on transformer and conventional learning algorithms. This is achieved by encoding smaller models in their activation and updating them corresponding with emergence of new examples in the context. They verify their hypothesis using a prototypical linear regression. \cite{zfb23} examined the in-context learning ability of a single head self-attention layer in linear regression tasks. When training by gradient flow, the transformer succeed in finding global optimum. However, they pointed out that, the model is not capable even when moderate shift happens in covariant distribution. \cite{wzw23} considered real-world LLMs as implicit topic models to test in-context learning by Bayesian lens. They viewed transformer in-context learning as a process of Bayesian selection, which secretly deduce information related to designated tasks.

Furthermore, research groups are motivated to gain more knowledge about the intrinsic structure of transformer. \cite{zpga23} discovered the parsing mechanism inside transformer in linguistic tasks. They verified their statement by designed attention module and probing experiments. They also related this process to a Inside-Outside algorithm. \cite{psza23} aimed to discover the location where newly learned skills reside in a fine-tuned transformer model, when they are assigned to specific tasks. They put forward a new concept named "skill location", which means that a few amount of updated parameter in fine-tune is sufficient for the performance of the whole model. They also demonstrated the implicit effect of localization on continual learning.

Others investigated the ability of LLMs established by transformer. \cite{sht23} empirically and analytically discussed the representational strength and limit of transformer. For example, it has logarithmically growing scale with input size when it learns a sparse average function, which performs better than other neural networks. However, it has linear growth rate in a triple detection task. \cite{ag23} seek a mathematical understanding about scaling up the training data and parameter size in LLM training. They established a strong connection between Scaling Law and inductive bias that makes pre-trained model to learn efficient. They proposed a potential conceptual framework for next AI generation. \cite{bce+23} provided a thorough investigation of GPT-4 \cite{o23}. They gave affirmative evaluation on its performance in several tasks, including coding, mathematics, vision, psychology and more. They also pointed out the potential future research direction that a novel model beyond next-word prediction is in need. Our work switch the attention to a specific 2-layer regression problem with nonlinear units inspired by transformer model.

\paragraph{Fast attention computation.}

Fine-tuning pre-trained LLMs is challenging due to the huge-scale parameter set. Researchers have been worked on finding efficient method to calculate attention module. Several articles discussed the usage of locality sensitive hashing (LSH) technique in attention approximation, including \cite{kkl20}, \cite{clp+21}, and \cite{zhdk23}. \cite{kkl20} provided two techniques to increase efficiency. They utilize a LSH to substitute dot product attention, which significantly lowers time complexity. They also uses a reversible residual layer to replace standard residual. \cite{clp+21} improved the approximation based on the observation that LSH does not experience constant requirement in updating the model parameters. \cite{zhdk23} proposed a novel estimator based on Kernel Density Estimation (KDE), supported by the observation that KDE can be exerted to accelerate softmax function computing and matrix multiplication. \cite{pmxa23} introduced approximation techniques that use a transformer in transformer (TinT) model to simulate the forward pass and back-propagation of a transformer, which significantly improved the parameter-efficiency. \cite{mgn+23} researched about efficiently fine-tuning LLMs with huge memory consumption. Improving classical ZO-SCD optimizer, they designed a memory-efficient MeZO gradient estimator with only forward pass. 
\cite{as23} provide tight bound for static attention and \cite{bsz23} prove results for dynamic attention problem. \cite{gsyz23} present quantum algorithm for attention computation.

\paragraph{LLM-related optimizer.} 

Gradient-based algorithm is a fundamental topic in machine learning. Recently, researchers have been dedicated to put forward efficient optimizer to solve LLM-related optimization problems. \cite{clmy21} studied huge-scale optimization problems, where conducting basic vector operation on decision variable is impossible. They made use of block gradient estimator and derived an algorithm that significantly reduces the query complexity and computation complexity per iteration. \cite{rsm+23} proposed Direct Preference Optimization algorithm, which directly fine-tunes a LLM based on a given human preference dataset, without explicit utilization of reward model or reinforcement learning. \cite{llh+23} proposed an efficient second-order optimizer of LLM based on diagonal Hessian approximation and clipping mechanism. Also, they relaxed the assumption that Hessian is Lipschitz in their proof of the convergence. Inspired by their novel argument, we apply a similar proof to handle ReLU function in our regression problem.

\paragraph{Convergence of ReLU Neural Network through Over-Parameterization.}
Over-parameterization is a commonly-used technique in deep learning which means that the trainable parameters are much larger than input dataset. It gives a explanation about the human-like performance of neural networks in different tasks. Previous work focused on convergence analysis of over-parameterized models, many of them discussed the ReLU network, e.g., \cite{LL18, DZPS19, azls19a, azls19b, ADH+19a, ADH+19b, SY19, CGH+19, ZMG19, CG19, ZG19, OS20, JT20, LSS+20, HLSY21, ZPD+20, BPSW21, SZZ21, Zha22, als+22, gqsw22, gms23, qsy23, sy23}. \cite{JGH18} had an important observation about over-parameterized neural network, which is the neural network is equivalent to a neural tangent kernel (NTK) when its width tends to infinity. Therefore, technical tools from kernel methods can be applied on analysis of deep neural networks. \cite{azls19a} demonstrated that over-parameterization guarantee converges for Stochastic Gradient Method (SGD). It attains linear convergence speed in a regression problem. Their result can be extended to non-smooth function like ReLU function. \cite{syz21} introduced a shifted ReLU activation and a shifted NTK technique with exact high-dimensional search data structure. These mechanisms significantly accelerate the training of over-parameterized neural networks, demonstrating the ability of pre-processing data. \cite{als+22} further improved the time-complexity of pre-processing by the implementation of tree data structure. They also provided a lower bound for time complexity, emphasising the optimality of their method.

\paragraph{LLM-inspired regression model.} 

Some recent works are done to analyze regression models by commonly-used activation functions in neural network. \cite{gms23} did a convergence analysis on over-parameterized neural networks setting exponential activation function as the activation function. \cite{lsz23} improved the formulation in \cite{gms23} by adding a regularization term on the exponential function to ensure strongly convexity. They introduced an efficient algorithm to simulate PD matrix explored in \cite{dsw22}, which is commonly-used to achieve calculation-efficient (ex. \cite{cls19, lsz19, Son19, Bra20, jswz21, sy21, hjs+22,
gs22, dsw22, syyz22, jlsz23, qszz23, gsz23}). Based on such estimator, they proposed a convergent approximation Newton method algorithm. Also, they discussed the case that sinh and cosh function are the activation function. \cite{dls23} investigated the softmax function, which is fundamental in attention computation. \cite{gsy23} probed the rescaled version of softmax function and hyperbolic functions. To do so, they multiplied the loss function by the sum of the regression function value of each entry. \cite{gsx23} leveraged a tensor-trick to reformulate a matrix expression of regression model into a vectorized version. They studied the lipschitz condition of several loss function, including softmax, sparse, entropy, and cross-entropy functions in vectorized version. Then, they provided an application of Lipschitz analysis in the field of in-context learning. \cite{dlms23} exerted a zero-th order algorithm to solve softmax regression which is efficient for large-scale LLMs. This work moves one step further from previous results. We analyze a two-layer regression unit, and we introduce new techniques to handle the situation in which Hessian Lipschitz is not guaranteed.

%% file: conclusion.tex
\section{Conclusion}\label{sec:conclusion}
This work studies a two-layer regression problem consists of a softmax unit and a ReLU unit inspired by LLM training and attention mechanism. We calculate the Lipschitz constant for Hessian of loss function under specific condition. Then, we add a regularization term to prove the PD of Hessian. Rooted in these two standard properties, we propose a convergent algorithm to find the optimal solution for the regression model based on approximation Newton method.

Continuing our research, we relax the assumption of fixed ReLU state, and we %introduce a technique from \cite{llh+23} and
prove the convergence of Newton method in the sense of loss value.

While this article makes significant progress, it is important to acknowledge that some of the assumptions made still possess a certain degree of strictness. Looking forward, our future work aims to relax these assumptions and explore innovative arguments to further enhance the understanding of convergence. Additionally, an ongoing challenge in this research lies in efficiently calculating and representing the Hessian in a closed-form manner.

Furthermore, we envision an intriguing research direction: extending the results obtained in this paper to more complex architectures such as general LLMs. Examining the applicability of our findings to these broader frameworks could unveil insightful implications for practical implementations and lead to further advancements in the field.

%% file: preli.tex
{\bf Roadmap.} In Section~\ref{sec:preli}, we list some useful lemmas, and we give definitions of basic functions. We derive the gradient of basic functions in Section~\ref{sec:grad}. We calculate the hessian of basic functions in Section~\ref{sec:hess}, and we state a reformulation of $\nabla^2 L(x)$. We prove some bounds for basic functions in Section~\ref{sec:bound}. We calculate the Lipschitz constant of basic functions in Section~\ref{sec:lipschitz_basic}. Using these lemmas, we prove the Lipschitz condition of $\nabla^2 L(x)$ in Section~\ref{sec:lipschitz_hess}. In Section~\ref{sec:psd}, we add a regularization term to $L(x)$ to make its Hessian to be PD. In Section~\ref{sec:newton}, we state the update rule of approximated Newton method under the model of this paper, and we prove its convergence. In Section~\ref{sec:cov_loss}, we prove the convergence in the sense of loss value under looser condition.

\section{Preliminary} \label{sec:preli}

In Section~\ref{sec:notation}, we list some notations used in this paper. In Section~\ref{sec:fac}, we state some useful lemmas in linear algebra and mathematical analysis. In Section~\ref{sec:def}, we give some definitions of basic function discussed in this paper.

\subsection{Notations} \label{sec:notation}
%%%Zhao: The correct label: {sec:preli:notation}
Let $z$ be a scalar, we define
\begin{align*}
 {\bf 1}[ z > 0 ] :=   \begin{cases}
        0, & z \leq 0 \\
        1, & z > 0
    \end{cases}
\end{align*}

Let $x \in \R^n$ be a vector, ${\bf 1}[x]$ denotes a vector $y \in \R^n$ such that $y_i := {\bf 1}[x_i > 0]$.

Let $x \in \R^n$ be a vector, $\exp(x) \in \R^n$ denotes the vector that $\exp(x)_i = \exp(x_i)$.

Let $f: \R^d \to \R^n $, let $t$ be a scalar, $\frac{\d f}{\d t} \in \R^n$ denotes a vector whose $i$-th entry is $\frac{\d f_i}{\d t}$.

Let $x, y$ be two vectors $\in \R^n$, let $\langle x,y \rangle$ be $\sum_{i=1}^n x_i \cdot y_i$.

Let $x, y$ be two vectors $\in \R^n$, let $x \circ y := z \in \R^n$ where $z_i = x_i \cdot y_i$.

Let ${\bf 1}_n$ denotes a vector with n entries and all of them are 1.

Let ${\bf 0}_n$ denotes a vector with n entries and all of them are 0.

Let ${\bf I}_n$ denotes a ${n \times n}$ identity matrix.

Let $[n] := \{1, 2, \cdots n \}$.

Let $\phi(x) := \max \{ 0,x \}$, i.e., the ReLU activation function.

\subsection{Facts} \label{sec:fac}
In this section, we state some useful lemmas which help us in the folowing proofs.
\begin{fact} \label{fac:diff_indicator}
We have $\frac{\d \phi(z)}{\d z} = {\bf 1}[ z > 0 ]$
\end{fact}

\begin{fact} [Hadamard product] \label{fac:hadamard_product_algebra}
For vector $y, z\in \R^n$
\begin{itemize}
    \item $\langle y \circ z, {\bf 1}_n \rangle = \langle y , z \rangle$
    \item $y \circ z = \diag(y) \cdot z$
    \item $y ^\top (z \circ w) = z ^\top (y \circ w)$
\end{itemize}
\end{fact}

\begin{fact}[Scalar algebra]\label{fac:scalar_algebra}
For $x,y$ belong to $\R$
\begin{itemize}
    \item $| \phi(x) - \phi(y) | \leq |x - y|$
\end{itemize}
\end{fact}

\begin{fact} [Vector algebra] \label{fac:vector_norm}
For vectors $x,y$ belong to $ \R^n$
\begin{itemize}
    \item $\| x \|_{\infty} \leq \| x \|_2 \leq \sqrt{n} \| x \|_{\infty}$
    \item $\| x \circ y \|_2 \leq \| x \|_2 \cdot \| y \|_2$
    \item $\| \exp(x) \|_{\infty} \leq \exp(\| x \|_2)$ 
\end{itemize}
\end{fact}

\begin{fact} [Mean value theorem for vector function] \label{fac:mvt}
For vector $x, y \in C \subset \R^n$, vector function $f(x): C \to \R$, $g(x): C \to \R^m$, let $f,g$ be differentiable on open convex domain $C$, we have
\begin{itemize}
    \item Part 1 $f(y) - f(x) = \nabla f(x+t(y-x))^\top (y-x)$
    \item Part 2 $\| g(y) - g(x) \|_2 \leq \| g'(x+t(y-x))\| \cdot \| y - x \|_2$ for some $t \in (0,1)$, where $g'(a)$ denotes a matrix which its $(i,j)$-th term is $\frac{\d g(a)_j}{\d a_i}$.
    \item Part 3 If $\| g'(a) \| \leq M$ for all $a \in C$, then $\| g(y) - g(x) \|_2 \leq M \| y - x \|_2$ for all $x, y \in C$
\end{itemize}
\begin{proof}
{\bf Proof of Part 1}

{\bf Part 1} can be verified by applying Mean Value Theorem of 1-variable function on $\gamma(c) = f(x + c(y-x))$.
\begin{align*}
    f(y) - f(x) = \gamma(1) - \gamma(0) = \gamma'(t) (1-0) = \nabla f(x+t(y-x))^\top (y-x)
\end{align*}
where $t \in (0,1)$.

{\bf Proof of Part 2}

Let $G(c) := (g(y) - g(x))^\top g(c)$, we have
\begin{align*}
    \| g(y) - g(x) \|_2^2 = & ~ G(y) - G(x) \\
    = & ~ \nabla G(x+t(y-x)) ^\top (y-x) \\
    = & ~ (\underbrace{g'(x+t(y-x))}_{d \times n} \cdot \underbrace{(g(y) - g(x))}_{n \times 1})^\top \cdot \underbrace{(y-x)}_{d \times 1} \\
    \leq & ~ \| g'(x+t(y-x)) \| \cdot \| g(y) - g(x) \|_2 \cdot \| y-x \|_2
\end{align*}
the initial step is by basic calculation, the second step is from {\bf Part 1}, the third step uses chain rule, the 4th step is due to Cauchy-Schwartz inequality. Removing $\| g(y) - g(x) \|_2$ on both sides gives the result.

{\bf Proof of Part 3}

{\bf Part 3} directly follows from {\bf Part 2}.
\end{proof}
    
\end{fact}

\begin{fact} [Matrix algebra] \label{fac:matrix_algebra}
Let $A \in \R^{n \times d}$, $B \in \R^{d \times m}$ Let $x, y \in \R^d$. Let $\| \cdot \|$ represent the l2 induced norm of matrix. 
We have
\begin{itemize}
    \item $\| Ax \|_2 \leq \| A \| \cdot \| x \|_2$
    \item $\| A \cdot B \| \leq \| A \| \cdot \| B \|$
    \item $\| A \| \leq \| A \|_F$, where $\| \cdot \|_F$ denotes the Frobenius norm
    \item $\| \phi(A x) - \phi(A y) \|_2 \leq \| A (x - y) \|_2$
    \item $\| x \cdot y ^\top \| \leq \| x \|_2 \cdot \| y \|_2$
    \item $\| \diag(x) \| = \lambda_{\max}(\diag(x)) = \| x \|_\infty \leq \| x \|_2$, where $\lambda_{\max}$ denotes the eigenvalue with the biggest absolute value
\end{itemize}
\begin{proof}
The second part directly follows from Fact~\ref{fac:scalar_algebra}.
\end{proof}
\end{fact}

\begin{fact} [PSD Lemmas] \label{fac:psd}
Let $u, v \in \R^{n}$, we have
\begin{itemize}
    \item $-\| u \|_2^2 \cdot {\bf I}_n \preceq uu^\top \preceq \| u \|_2^2 \cdot {\bf I}_n$
    \item $-\| u \|_2 \cdot {\bf I}_n \preceq \diag(u) \preceq \| u \|_2 \cdot {\bf I}_n$
    \item $-(uu^\top + vv^\top) \preceq uv^\top + vu^\top \preceq uu^\top + vv^\top$
\end{itemize}
    
\end{fact}

\subsection{Definitions} \label{sec:def}
Since $L(x)$ has a complicated formula, we separate and into basic functions to compute its Hessian.
In this section, we define some functions using following notations
\begin{itemize}
\item the matrix $A_1 \in \R^{n \times d}$ 
\item the matrix $A_2 \in \R^{m \times n}$.
\item the vector $b \in \R^m$
\end{itemize}

\begin{definition} \label{def:h}
We define $h: \R^d \rightarrow \R_{\geq 0}^n$ as 
\begin{align*}
    h(x) = \phi(A_1 \cdot x).
\end{align*}
\end{definition}

\begin{definition} \label{def:u}
Suppose $h(x)$ is defined as Definition~\ref{def:h}.

We define $u : \R^d \rightarrow \R_{>0}^m$ to be
\begin{align*}
    u(x) := \exp(A_2 \cdot h(x) )
\end{align*}
\end{definition}

\begin{definition} \label{def:alpha}
Suppose $u(x)$ is defined as Definition~\ref{def:u}.

We define $\alpha: \R^d \rightarrow \R_{>0}$ as
\begin{align*}
    \alpha(x) := \langle u(x), {\bf 1}_m \rangle
\end{align*}
\end{definition}

\begin{definition} \label{def:f}
Suppose $\alpha(x)$, $u(x)$ are defined as Definition~\ref{def:alpha} and Definition~\ref{def:u}.

We define $f: \R^d \rightarrow \R_{>0}^m$ as
\begin{align*}
    f(x) := \alpha(x)^{-1} \cdot u(x)
\end{align*}
\end{definition}

\begin{definition} \label{def:c}
Assume $f(x)$ is defined as Definition~\ref{def:f}.

We define $c: \R^d \rightarrow \R^m$ to be
\begin{align*}
    c(x) := f(x) - b
\end{align*}
\end{definition}

\begin{definition} \label{def:L}
Assume $c(x)$ is defined as Definition~\ref{def:c}.

We define Loss $L : \R^d \rightarrow \R_{>0}$ as
\begin{align*}
   L(x) := \frac{1}{2} \cdot \| c(x) \|_2^2.
\end{align*}
\end{definition}

%% file: gradient.tex
\section{Gradient} \label{sec:grad}

In this section, we derive the gradients of basic functions. Our main goal is use secondary results to compute $\nabla L(x)$

\subsection{Basic derivatives}

%We want to work on

\begin{lemma} \label{lem:basic_derivatives}
Let $A_{1,i}$ denotes the $i$-th column of matrix $A_1 \in \R^{n \times d}$, we have
\begin{itemize}
    \item Part 1.
    \begin{align*}
       \frac{\d \phi(A_1 x)}{\d x_i} = {\bf 1}[A_1 x] \circ A_{1,i}.
    \end{align*}

    \item Part 2
    \begin{align*}
        \frac{\d A_2 \phi(A_1 x)}{\d x_i} = A_2 \cdot ({\bf 1}[A_1 x] \circ A_{1,i}).
    \end{align*}

    \item Part 3. 
    \begin{align*}
        \frac{\d u(x)}{\d x_i} = u(x) \circ (A_2 \cdot ({\bf 1}[A_1 x] \circ A_{1,i}))
    \end{align*}
    
    \item Part 4.
    \begin{align*}
        \frac{\d \alpha(x)}{ \d x_i} =  \langle u(x) , A_2 \cdot ({\bf 1}[A_1 x] \circ A_{1,i}) \rangle
    \end{align*}
    
    \item Part 5. 
    \begin{align*}
        \frac{\d \alpha(x)^{-1}}{\d x_i} = - \alpha(x)^{-1} \cdot \langle f(x) , A_2 \cdot ({\bf 1}[A_1 x] \circ A_{1,i}) \rangle
    \end{align*}

    \item Part 6.
    \begin{align*}
        \frac{\d f(x)}{\d x_i} = f(x) \circ (A_2 \cdot ({\bf 1}[A_1 x] \circ A_{1,i})) - f(x) \cdot \langle f(x) , A_2 \cdot ({\bf 1}[A_1 x] \circ A_{1,i}) \rangle
    \end{align*}

    \item Part 7.
    \begin{align*}
        \frac{\d c(x)}{\d x_i} = f(x) \circ (A_2 \cdot ({\bf 1}[A_1 x] \circ A_{1,i})) - f(x) \cdot \langle f(x) , A_2 \cdot ({\bf 1}[A_1 x] \circ A_{1,i}) \rangle
    \end{align*}
    
    \item Part 8.
    \begin{align*}
        \frac{\d L(x)}{\d x_i} = \langle c(x), f(x) \circ (A_2 \cdot ({\bf 1}[A_1 x] \circ A_{1,i})) \rangle - \langle c(x), f(x) \rangle \cdot \langle f(x) , A_2 \cdot ({\bf 1}[A_1 x] \circ A_{1,i}) \rangle
    \end{align*}
    
    \item Part 9
    \begin{align*}
        \frac{ \d {\bf 1}[A_1 x] }{\d x_i} = {\bf 0}_n
    \end{align*}
     
\end{itemize}
\begin{proof}
Our proof is outlined below.
\paragraph{Proof of Part 1.}
\begin{align*}
    \frac{\d \phi(A_1x)}{\d x_i} = & ~ \frac{\d \phi(y)}{\d y_i} |_{y = A_1 x} \circ \frac{\d A_1 x}{\d x_i} \\
    = & ~ {\bf 1}[A_1 x] \circ \frac{\d A_1 x}{\d x_i} \\
    = & ~ {\bf 1}[A_1 x] \circ A_{1,i}
\end{align*}
where the first step uses chain rule, the 2nd step is by Fact~\ref{fac:diff_indicator}, the 3rd step holds because of the definition of matrix product.

\paragraph{Proof of Part 2.}
\begin{align*}
    \frac{\d A_2 \phi(A_1 x)}{\d x_i} = & ~ A_2 \cdot \frac{\d \phi(A_1 x)}{\d x_i} \\
    = & ~ A_2 \cdot ({\bf 1}[A_1 x] \circ A_{1,i})
\end{align*}
where the first step holds because of the definition of matrix product, the 2nd step is due to {\bf Part 1}.

\paragraph{Proof of Part 3}
\begin{align*}
    \frac{\d u(x)}{\d x_i} = & ~ \frac{\d \exp(A_2 \phi(A_1 x))}{\d x_i} \\
    = & ~ \exp(A_2 \phi(A_1 x)) \circ \frac{\d A_2 \phi(A_1 x)}{\d x_i} \\
    = & ~ u(x) \circ (A_2 \cdot ({\bf 1}[A_1 x] \circ A_{1,i}))
\end{align*}
where the 1st step is from the definition of $u$ (see Definition~\ref{def:u}), the second step uses the chain rule, the third step is derived from {\bf Part 2}.

\paragraph{Proof of Part 4}
\begin{align*}
    \frac{\d \alpha(x)}{\d x_i} = & ~ \frac{\d \langle u(x), {\bf 1}_m \rangle}{\d x_i} \\
    = & ~ \langle \frac{\d u(x)}{\d x_i}, {\bf 1}_m \rangle \\
    = & ~ \langle u(x) \circ (A_2 \cdot ({\bf 1}[A_1 x] \circ A_{1,i})), {\bf 1}_m \rangle \\
    = & ~ \langle u(x) , A_2 \cdot ({\bf 1}[A_1 x] \circ A_{1,i}) \rangle
\end{align*}
where the 1st step is due to the definition of $\alpha$ (see Definition~\ref{def:alpha}), the 2nd step is from sum rule of taking derivative, the third step follows by {\bf Part 3} of this lemma, the fourth step is due to Fact~\ref{fac:hadamard_product_algebra}.

\paragraph{Proof of Part 5}
\begin{align*}
    \frac{\d \alpha(x)^{-1}}{\d x_i} = & ~ - \alpha(x)^{-2} \cdot \frac{\d \alpha(x)}{\d x_i} \\
    = & ~ - \alpha(x)^{-2} \cdot \langle u(x) , A_2 \cdot ({\bf 1}[A_1 x] \circ A_{1,i}) \rangle \\
    = & ~ - \alpha(x)^{-1} \cdot \langle f(x) , A_2 \cdot ({\bf 1}[A_1 x] \circ A_{1,i}) \rangle
\end{align*}
where the first step uses chain rule, the second step is from {\bf Part 4}, the 3rd step is given by definition of $f$ (refer to Definition~\ref{def:f}).

\paragraph{Part 6}
\begin{align*}
    \frac{\d f(x)}{\d x_i} = & ~ \frac{\d ( \alpha(x)^{-1} \cdot u(x)) }{ \d x_i} \\
    = & ~ \alpha(x)^{-1} \cdot \frac{\d u(x) }{ \d x_i} + \frac{\d \alpha(x)^{-1}}{\d x_i} \cdot u(x) \\
    = & ~ \alpha(x)^{-1} \cdot u(x) \circ (A_2 \cdot ({\bf 1}[A_1 x] \circ A_{1,i})) + \frac{\d \alpha(x)^{-1}}{\d x_i} \cdot u(x) \\
    = & ~ \alpha(x)^{-1} \cdot u(x) \circ (A_2 \cdot ({\bf 1}[A_1 x] \circ A_{1,i})) - \alpha(x)^{-1} \cdot \langle f(x) , A_2 \cdot ({\bf 1}[A_1 x] \circ A_{1,i})\rangle \cdot u(x) \\
    = & ~ f(x) \circ (A_2 \cdot ({\bf 1}[A_1 x] \circ A_{1,i})) - f(x) \cdot \langle f(x) , A_2 \cdot ({\bf 1}[A_1 x] \circ A_{1,i}) \rangle
\end{align*}
where the first step holds because the definition of $f$ (see Definition~\ref{def:f}, the 2nd step uses product rule, the third step is given by {\bf Part 3}, the fourth step follows by {\bf Part 4}.

\paragraph{Part 7}
The result is obvious since $c(x)$ and $f(x)$ differ by only one constant term.

\paragraph{Proof of Part 8}
\begin{align*}
    \frac{\d L(x)}{\d x_i} = & ~ \frac{1}{2} \cdot \frac{\d \| c(x) \|_2^2}{\d x_i} \\
    = & ~ \langle c(x), \frac{\d c(x)}{\d x_i} \rangle \\
    = & ~ \langle c(x), f(x) \circ (A_2 \cdot ({\bf 1}[A_1 x] \circ A_{1,i})) - f(x) \cdot \langle f(x) , A_2 \cdot ({\bf 1}[A_1 x] \circ A_{1,i}) \rangle \rangle \\
    = & ~ \langle c(x), f(x) \circ (A_2 \cdot ({\bf 1}[A_1 x] \circ A_{1,i})) \rangle - \langle c(x), f(x) \rangle \cdot \langle f(x) , A_2 \cdot ({\bf 1}[A_1 x] \circ A_{1,i}) \rangle
\end{align*}
where the 1st step is from definition of $L$ (see Definition~\ref{def:L}), the second step uses chain rule, the third step is derived from {\bf Part 6}, the 4th step is inner product calculation.

\paragraph{Proof of Part 9}
The result simply follows by ${\bf 1}[A_1 x]$ is a step function composing by 2 constant functions.

\end{proof}

\end{lemma}

%% file: hessian.tex
\section{Hessian} \label{sec:hess}
This section presents the Hessian of basic functions, including $h(x)$ (see Section~\ref{sec:hess_h}), $u(x)$ (see Section~\ref{sec:hess_u}), $\alpha(x)$ (see Section~\ref{sec:hess_alpha}), $\alpha^{-1}(x)$ (see Section~\ref{sec:hess_alpha_inverse}), $f(x)$ (see Section~\ref{sec:hess_f}). Next, we compute $\nabla^2 L(x)$ based on the result of basic functions (see Section~\ref{sec:hess_L}). 

The direct result in Section~\ref{sec:hess_L} is difficult to analyze, hence we exert a close-form decomposition technique to simplify its expression \cite{ans00,hjs+22}. In Section~\ref{sec:hess_L_decompose_separate}, we simplify each term in $\nabla^2 L(x)$. In Section~\ref{sec:hess_L_decompose}, we summarize the simplifications to get a decomposition of $\nabla^2 L(x)$.

\subsection{Hessian of \texorpdfstring{$h(x)$}{}} \label{sec:hess_h}
In this section, we derive $\nabla^2 h(x)$.

\begin{lemma} [Hessian of $h(x)$] \label{lem:hessian_h} Supposed $h$ is defined as Definition~\ref{def:h},
\begin{itemize}
\item Part 1.
\begin{align*}
    \frac{\d ^2 h(x)}{\d x_i^2} = 0 % {\bf 1}[A_1 x] \circ A_{1,i} \circ A_{1,i}
\end{align*}
\item Part 2.
\begin{align*}
    \frac{\d ^2 h(x)}{\d x_i \d x_j} = 0 % {\bf 1}[A_1 x] \circ A_{1,i} \circ A_{1,j}
\end{align*}
\end{itemize}

\begin{proof}

{\bf Part 1}
\begin{align*}
    \frac{\d ^2 h(x)}{\d x_i^2} = & ~ \frac{\d}{\d x_i} ({\bf 1}[A_1 x] \circ A_{1,i}) \\
    = & ~ 0 % {\bf 1}[A_1 x] \circ A_{1,i} \circ A_{1,i}
\end{align*}
where the first step is given by {\bf Part 1} of Lemma~\ref{lem:basic_derivatives}, the 2nd step uses {\bf Part 9} of Lemma~\ref{lem:basic_derivatives}.

{\bf Proof of Part 2}
\begin{align*}
    \frac{\d ^2 h(x)}{\d x_i \d x_j} = & ~ \frac{\d}{\d x_j} ({\bf 1}[A_1 x] \circ A_{1,i}) \\
    = & ~ 0 % {\bf 1}[A_1 x] \circ A_{1,i} \circ A_{1,j}
\end{align*}
where the initial step is derived from {\bf Part 1} of Lemma~\ref{lem:basic_derivatives}, the second step follows {\bf Part 9} of Lemma~\ref{lem:basic_derivatives}.

\end{proof}
\end{lemma}

\subsection{Hessian of \texorpdfstring{$u(x)$}{}}\label{sec:hess_u}
In this section, we derive $\nabla^2 u(x)$.

\begin{lemma} [Hessian of $u(x)$] \label{lem:hessian_u} Assume $u$ is defined as Definition~\ref{def:u},
\begin{itemize}
\item Part 1.
\begin{align*}
    \frac{\d ^2 u(x)}{\d x_i^2} = u(x) \circ ((A_2 \cdot ({\bf 1}[A_1 x] \circ A_{1,i})) \circ (A_2 \cdot ({\bf 1}[A_1 x] \circ A_{1,i})) 
\end{align*}
\item Part 2.
\begin{align*}
    \frac{\d ^2 u(x)}{\d x_i \d x_j} = u(x) \circ ((A_2 \cdot ({\bf 1}[A_1 x] \circ A_{1,j})) \circ (A_2 \cdot ({\bf 1}[A_1 x] \circ A_{1,i}))  
\end{align*}
\end{itemize}

\begin{proof}
{\bf Part 1}
\begin{align*}
    \frac{\d ^2 u(x)}{\d x_i^2} = & ~ \frac{\d}{\d x_i} (u(x) \circ A_2 \cdot ({\bf 1}[A_1 x] \circ A_{1,i})) \\
    = & ~ \frac{\d u(x)}{\d x_i} \circ (A_2 \cdot ({\bf 1}[A_1 x] \circ A_{1,i})) \\ %+ u(x) \circ \frac{\d A_2 \cdot ({\bf 1}[A_1 x] \circ A_{1,i})}{\d x_i} \\
    = & ~ u(x) \circ (A_2 \cdot ({\bf 1}[A_1 x] \circ A_{1,i})) \circ (A_2 \cdot ({\bf 1}[A_1 x] \circ A_{1,i})) \\%~ + \\
    %& ~ u(x) \circ (A_2 \cdot ({\bf 1}[A_1 x] \circ A_{1,i} \circ A_{1,i})) \\
   % = & ~ u(x) \circ ((A_2 \cdot ({\bf 1}[A_1 x] \circ A_{1,i})) \circ (A_2 \cdot ({\bf 1}[A_1 x] \circ A_{1,i})) %+  A_2 \cdot ({\bf 1}[A_1 x] \circ A_{1,i} \circ A_{1,i}))
\end{align*}
where the initial step is derived by {\bf Part 3} of Lemma~\ref{lem:basic_derivatives}, the 2nd step uses {\bf Part 9} of Lemma~\ref{lem:basic_derivatives}, the 3rd step follows by {\bf Part 3} of Lemma~\ref{lem:basic_derivatives}.

{\bf Proof of Part 2}
\begin{align*}
    \frac{\d ^2 u(x)}{\d x_i \d x_j} = & ~ \frac{\d}{\d x_i} (u(x) \circ A_2 \cdot ({\bf 1}[A_1 x] \circ A_{1,i})) \\
    = & ~ \frac{\d u(x)}{\d x_j} \circ (A_2 \cdot ({\bf 1}[A_1 x] \circ A_{1,i}))  \\
    = & ~ u(x) \circ (A_2 \cdot ({\bf 1}[A_1 x] \circ A_{1,j})) \circ (A_2 \cdot ({\bf 1}[A_1 x] \circ A_{1,i}))  
\end{align*}
where the 1st step is given by {\bf Part 3} of Lemma~\ref{lem:basic_derivatives}, the 2nd step uses {\bf Part 9} of Lemma~\ref{lem:basic_derivatives}, the 3rd step follows by {\bf Part 3} of Lemma~\ref{lem:basic_derivatives}.

\end{proof}
\end{lemma}

\subsection{Hessian of \texorpdfstring{$\alpha(x)$}{}}\label{sec:hess_alpha}
In this section, we derive $\nabla^2 \alpha(x)$.

\begin{lemma} [Hessian of $\alpha(x)$] \label{lem:hessian_alpha} We set $\alpha$ to be defined as Definition~\ref{def:alpha},
\begin{itemize}
\item Part 1.
\begin{align*}
    \frac{\d ^2 \alpha(x)}{\d x_i^2} = \langle u(x) ,(A_2 \cdot ({\bf 1}[A_1 x] \circ A_{1,i})) \circ  (A_2 \cdot ({\bf 1}[A_1 x] \circ A_{1,i}))  \rangle
\end{align*}
\item Part 2.
\begin{align*}
    \frac{\d ^2 \alpha(x)}{\d x_i \d x_j} = \langle u(x) , (A_2 \cdot ({\bf 1}[A_1 x] \circ A_{1,j}) \circ (A_2 \cdot ({\bf 1}[A_1 x] \circ A_{1,i}))  \rangle
\end{align*}
\end{itemize}

\begin{proof}
%\Zhao{Please fix the proofs yourself}
{\bf Part 1}
\begin{align*}
    \frac{\d ^2 \alpha(x)}{\d x_i^2} = & ~ \frac{\d}{\d x_i} (\langle u(x) , A_2 \cdot ({\bf 1}[A_1 x] \circ A_{1,i}) \rangle) \\
    = & ~ \langle \frac{\d}{\d x_i} u(x) , A_2 \cdot ({\bf 1}[A_1 x] \circ A_{1,i}) \rangle \\
    = & ~ \langle u(x) , (A_2 \cdot ({\bf 1}[A_1 x] \circ A_{1,i}) \circ (A_2 \cdot ({\bf 1}[A_1 x] \circ A_{1,i}) )\rangle
\end{align*}
where the 1st step is given by {\bf Part 4} of Lemma~\ref{lem:basic_derivatives}, the 2nd step follows by Lemma~\ref{lem:hessian_h}, the 3th step is derived from Lemma~\ref{lem:hessian_u}.

{\bf Proof of Part 2}
\begin{align*}
    \frac{\d ^2 \alpha(x)}{\d x_i \d x_j} = & ~ \frac{\d}{\d x_j} (\langle u(x) , A_2 \cdot ({\bf 1}[A_1 x] \circ A_{1,i}) \rangle) \\
    = & ~ \langle \frac{\d}{\d x_j} u(x) , A_2 \cdot ({\bf 1}[A_1 x] \circ A_{1,i}) \rangle \\
    = & ~ \langle u(x) , (A_2 \cdot ({\bf 1}[A_1 x] \circ A_{1,j}) \circ (A_2 \cdot ({\bf 1}[A_1 x] \circ A_{1,i}) )\rangle
\end{align*}
where the first step uses {\bf Part 4} of Lemma~\ref{lem:basic_derivatives}, the 2nd step utilizes Lemma~\ref{lem:hessian_h}, the 3th step is derived from Lemma~\ref{lem:hessian_u}.

\end{proof}
\end{lemma}

\subsection{Hessian of \texorpdfstring{$\alpha(x)^{-1}$}{}}\label{sec:hess_alpha_inverse}
In this section, we derive $\nabla^2 \alpha^{-1}(x)$.

\begin{lemma} [Hessian of $\alpha^{-1}(x)$] \label{lem:hessian_alpha_inverse} Suppose $\alpha$ is defined as Definition~\ref{def:alpha},
\begin{itemize}
\item Part 1.
\begin{align*}
    \frac{\d ^2 \alpha^{-1}(x)}{\d x_i^2} = \alpha(x)^{-3} \cdot \langle f(x) , (A_2 \cdot ({\bf 1}[A_1 x] \circ A_{1,i})) \circ (A_2 \cdot ({\bf 1}[A_1 x] \circ A_{1,i}))
\end{align*}
\item Part 2.
\begin{align*}
    \frac{\d ^2 \alpha^{-1}(x)}{\d x_i \d x_j} = \alpha(x)^{-3} \cdot \langle f(x) , (A_2 \cdot ({\bf 1}[A_1 x] \circ A_{1,i})) \circ (A_2 \cdot ({\bf 1}[A_1 x] \circ A_{1,j}))
\end{align*}
\end{itemize}

\begin{proof}

%\Zhao{Please fix the following proofs yourself.}
{\bf Part 1}
\begin{align*}
    & \frac{\d ^2 \alpha^{-1}(x)}{\d x_i^2} \\
   = & ~ \frac{\d}{\d x_i} (- \alpha(x)^{-1} \cdot \langle f(x) , A_2 \cdot ({\bf 1}[A_1 x] \circ A_{1,i}) \rangle) \\
    = & ~ \alpha(x)^{-2} \cdot ((\frac{\d \langle f(x) , A_2 \cdot ({\bf 1}[A_1 x] \circ A_{1,i}) \rangle}{\d x_i}) \cdot \alpha(x)^{-1} -  \frac{\d \alpha(x)^{-1}}{\d x_i} \cdot \langle f(x) , A_2 \cdot ({\bf 1}[A_1 x] \circ A_{1,i}) \rangle) \\
    = & ~ \alpha(x)^{-2} \cdot (Q_1(x) - Q_2(x))
\end{align*}
For the first term $Q_1(x)$, we have
\begin{align*}
    & Q_1(x) \\
    = & ~ (\langle \frac{\d f(x)}{\d x_i} , A_2 \cdot ({\bf 1}[A_1 x] \circ A_{1,i}) \rangle + \langle f(x) , \frac{\d A_2 \cdot ({\bf 1}[A_1 x] \circ A_{1,i})}{\d x_i} \rangle) \cdot \alpha(x)^{-1} \\
    = & ~ \langle \frac{\d f(x)}{\d x_i} , A_2 \cdot ({\bf 1}[A_1 x] \circ A_{1,i}) \rangle \cdot \alpha(x)^{-1} \\
    = & ~ \alpha(x)^{-1} \cdot \langle f(x) \circ (A_2 \cdot ({\bf 1}[A_1 x] \circ A_{1,i})) - f(x) \cdot \langle f(x) , A_2 \cdot ({\bf 1}[A_1 x] \circ A_{1,i}) \rangle , A_2 \cdot ({\bf 1}[A_1 x] \circ A_{1,i}) \rangle \\
    = & ~ \alpha(x)^{-1} \cdot (\langle f(x) \circ (A_2 \cdot ({\bf 1}[A_1 x] \circ A_{1,i})), A_2 \cdot ({\bf 1}[A_1 x] \circ A_{1,i}) \rangle ~- \\
    & ~ \langle f(x), A_2 \cdot ({\bf 1}[A_1 x] \circ A_{1,i}) \rangle \cdot \langle f(x) , A_2 \cdot ({\bf 1}[A_1 x] \circ A_{1,i}) \rangle) \\
    = & ~ \alpha(x)^{-1} \cdot (\langle f(x) , (A_2 \cdot ({\bf 1}[A_1 x] \circ A_{1,i})) \circ (A_2 \cdot ({\bf 1}[A_1 x] \circ A_{1,i})) \rangle ~- \\
    & ~ \langle f(x), A_2 \cdot ({\bf 1}[A_1 x] \circ A_{1,i}) \rangle \cdot \langle f(x) , A_2 \cdot ({\bf 1}[A_1 x] \circ A_{1,i}) \rangle)
\end{align*}
where the initial step is because of differential product rule, the second step is by Lemma~\ref{lem:hessian_h}, the third step is given by {\bf Part 6} of Lemma~\ref{lem:basic_derivatives}, the fourth step is inner product calculation, the last step follows by Fact~\ref{fac:hadamard_product_algebra}.

For the second term $Q_2(x)$, we have
\begin{align*}
    Q_2(x) = & ~ - \alpha(x)^{-1} \cdot \langle f(x) , A_2 \cdot ({\bf 1}[A_1 x] \circ A_{1,i}) \rangle \cdot \langle f(x) , A_2 \cdot ({\bf 1}[A_1 x] \circ A_{1,i}) \rangle
\end{align*}
which follows from {\bf Part 5} of  Lemma~\ref{lem:basic_derivatives}.

Therefore, we have
\begin{align*}
    \frac{\d ^2 \alpha^{-1}(x)}{\d x_i^2}
    = & ~ \alpha(x)^{-2} \cdot (Q_1(x) - Q_2(x)) \\
    = & ~ \alpha(x)^{-2} \cdot (\alpha(x)^{-1} \cdot (\langle f(x) , (A_2 \cdot ({\bf 1}[A_1 x] \circ A_{1,i})) \circ (A_2 \cdot ({\bf 1}[A_1 x] \circ A_{1,i})) \rangle ~- \\
    & ~ \langle f(x), A_2 \cdot ({\bf 1}[A_1 x] \circ A_{1,i}) \rangle \cdot \langle f(x) , A_2 \cdot ({\bf 1}[A_1 x] \circ A_{1,i}) \rangle) ~+  \\
    & ~ \alpha(x)^{-1} \cdot \langle f(x) , A_2 \cdot ({\bf 1}[A_1 x] \circ A_{1,i}) \rangle \cdot \langle f(x) , A_2 \cdot ({\bf 1}[A_1 x] \circ A_{1,i}) \rangle) \\
    = & ~ \alpha(x)^{-3} \cdot \langle f(x) , (A_2 \cdot ({\bf 1}[A_1 x] \circ A_{1,i})) \circ (A_2 \cdot ({\bf 1}[A_1 x] \circ A_{1,i})) \rangle
\end{align*}

{\bf Proof of Part 2}
\begin{align*}
    & \frac{\d ^2 \alpha^{-1}(x)}{\d x_i \d x_j} \\
    = & ~ \frac{\d}{\d x_j} (- \alpha(x)^{-1} \cdot \langle f(x) , A_2 \cdot ({\bf 1}[A_1 x] \circ A_{1,i}) \rangle) \\
    = & ~ \alpha(x)^{-2} \cdot ((\frac{\d \langle f(x) , A_2 \cdot ({\bf 1}[A_1 x] \circ A_{1,i}) \rangle}{\d x_j}) \cdot \alpha(x)^{-1} -  \frac{\d \alpha(x)^{-1}}{\d x_j} \cdot \langle f(x) , A_2 \cdot ({\bf 1}[A_1 x] \circ A_{1,i}) \rangle) \\
    = & ~ \alpha(x)^{-2} \cdot (Q_1(x) - Q_2(x))
\end{align*}
For the first term $Q_1(x)$, we have
\begin{align*}
    & Q_1(x) \\
    = & ~ (\langle \frac{\d f(x)}{\d x_j} , A_2 \cdot ({\bf 1}[A_1 x] \circ A_{1,i}) \rangle + \langle f(x) , \frac{\d A_2 \cdot ({\bf 1}[A_1 x] \circ A_{1,i})}{\d x_j} \rangle) \cdot \alpha(x)^{-1} \\
    = & ~ \langle \frac{\d f(x)}{\d x_j} , A_2 \cdot ({\bf 1}[A_1 x] \circ A_{1,i}) \rangle \cdot \alpha(x)^{-1} \\
    = & ~ \alpha(x)^{-1} \cdot \langle f(x) \circ (A_2 \cdot ({\bf 1}[A_1 x] \circ A_{1,j})) - f(x) \cdot \langle f(x) , A_2 \cdot ({\bf 1}[A_1 x] \circ A_{1,j}) \rangle , A_2 \cdot ({\bf 1}[A_1 x] \circ A_{1,i}) \rangle \\
    = & ~ \alpha(x)^{-1} \cdot (\langle f(x) \circ (A_2 \cdot ({\bf 1}[A_1 x] \circ A_{1,j})), A_2 \cdot ({\bf 1}[A_1 x] \circ A_{1,i}) \rangle ~- \\
    & ~ \langle f(x), A_2 \cdot ({\bf 1}[A_1 x] \circ A_{1,i}) \rangle \cdot \langle f(x) , A_2 \cdot ({\bf 1}[A_1 x] \circ A_{1,j}) \rangle) \\
    = & ~ \alpha(x)^{-1} \cdot (\langle f(x) , (A_2 \cdot ({\bf 1}[A_1 x] \circ A_{1,j})) \circ (A_2 \cdot ({\bf 1}[A_1 x] \circ A_{1,i})) \rangle ~- \\
    & ~ \langle f(x), A_2 \cdot ({\bf 1}[A_1 x] \circ A_{1,i}) \rangle \cdot \langle f(x) , A_2 \cdot ({\bf 1}[A_1 x] \circ A_{1,j}) \rangle)
\end{align*}
where the first step is given by differential product rule, the second step is from Lemma~\ref{lem:hessian_h}, the third step is given by {\bf Part 6} of Lemma~\ref{lem:basic_derivatives}, the 4th step is inner product calculation, the last step follows by Fact~\ref{fac:hadamard_product_algebra}.

For the second term $Q_2(x)$, we have
\begin{align*}
    Q_2(x) = & ~ - \alpha(x)^{-1} \cdot \langle f(x) , A_2 \cdot ({\bf 1}[A_1 x] \circ A_{1,j}) \rangle \cdot \langle f(x) , A_2 \cdot ({\bf 1}[A_1 x] \circ A_{1,i}) \rangle
\end{align*}
which follows from {\bf Part 5} of  Lemma~\ref{lem:basic_derivatives}.

Therefore, we have
\begin{align*}
    \frac{\d ^2 \alpha^{-1}(x)}{\d x_i \d x_j}
    = & ~ \alpha(x)^{-2} \cdot (Q_1(x) - Q_2(x)) \\
    = & ~ \alpha(x)^{-2} \cdot (\alpha(x)^{-1} \cdot (\langle f(x) , (A_2 \cdot ({\bf 1}[A_1 x] \circ A_{1,j})) \circ (A_2 \cdot ({\bf 1}[A_1 x] \circ A_{1,i})) \rangle ~- \\
    & ~ \langle f(x), A_2 \cdot ({\bf 1}[A_1 x] \circ A_{1,i}) \rangle \cdot \langle f(x) , A_2 \cdot ({\bf 1}[A_1 x] \circ A_{1,j}) \rangle) ~+  \\
    & ~ \alpha(x)^{-1} \cdot \langle f(x) , A_2 \cdot ({\bf 1}[A_1 x] \circ A_{1,j}) \rangle \cdot \langle f(x) , A_2 \cdot ({\bf 1}[A_1 x] \circ A_{1,i}) \rangle) \\
    = & ~ \alpha(x)^{-3} \cdot \langle f(x) , (A_2 \cdot ({\bf 1}[A_1 x] \circ A_{1,i})) \circ (A_2 \cdot ({\bf 1}[A_1 x] \circ A_{1,j})) \rangle
\end{align*}

\end{proof}
\end{lemma}

\subsection{Hessian of \texorpdfstring{$f(x)$}{}}\label{sec:hess_f}
In this section, we derive $\nabla^2 f(x)$.

\begin{lemma} [Hessian of $f(x)$] \label{lem:hessian_f} Suppose $f$ is defined as Definition~\ref{def:f},
\begin{itemize}
\item Part 1.
\begin{align*}
    \frac{\d^2 f(x)}{\d x_i^2}
    = & ~ f(x) \circ (A_2 \cdot ({\bf 1}[A_1 x] \circ A_{1,i})) \circ (A_2 \cdot ({\bf 1}[A_1 x] \circ A_{1,i})) ~- \\
    & ~ f(x) \circ (A_2 \cdot ({\bf 1}[A_1 x] \circ A_{1,i})) \cdot \langle f(x) , A_2 \cdot ({\bf 1}[A_1 x] \circ A_{1,i}) \rangle ~- \\
    & ~ f(x) \circ (A_2 \cdot ({\bf 1}[A_1 x] \circ A_{1,i})) \cdot \langle f(x), A_2 \cdot ({\bf 1}[A_1 x] \circ A_{1,i}) \rangle ~+ \\
    & ~ 2f(x) \cdot \langle f(x) , A_2 \cdot ({\bf 1}[A_1 x] \circ A_{1,i}) \rangle \cdot \langle f(x), A_2 \cdot ({\bf 1}[A_1 x] \circ A_{1,i}) \rangle ~- \\
    & ~ f(x) \cdot \langle f(x) \circ (A_2 \cdot ({\bf 1}[A_1 x] \circ A_{1,i})), A_2 \cdot ({\bf 1}[A_1 x] \circ A_{1,i}) \rangle
\end{align*}
\item Part 2.
\begin{align*}
    \frac{\d^2 f(x)}{\d x_i \d x_j}
    = & ~ f(x) \circ (A_2 \cdot ({\bf 1}[A_1 x] \circ A_{1,j})) \circ (A_2 \cdot ({\bf 1}[A_1 x] \circ A_{1,i})) ~- \\
    & ~ f(x) \circ (A_2 \cdot ({\bf 1}[A_1 x] \circ A_{1,i})) \cdot \langle f(x) , A_2 \cdot ({\bf 1}[A_1 x] \circ A_{1,j}) \rangle ~- \\
    & ~ f(x) \circ (A_2 \cdot ({\bf 1}[A_1 x] \circ A_{1,j})) \cdot \langle f(x), A_2 \cdot ({\bf 1}[A_1 x] \circ A_{1,i}) \rangle ~+ \\
    & ~ 2f(x) \cdot \langle f(x) , A_2 \cdot ({\bf 1}[A_1 x] \circ A_{1,j}) \rangle \cdot \langle f(x), A_2 \cdot ({\bf 1}[A_1 x] \circ A_{1,i}) \rangle ~- \\
    & ~ f(x) \cdot \langle f(x) \circ (A_2 \cdot ({\bf 1}[A_1 x] \circ A_{1,j})), A_2 \cdot ({\bf 1}[A_1 x] \circ A_{1,i}) \rangle
\end{align*}
\end{itemize}

\begin{proof}

{\bf Part 1}
\begin{align*}
    \frac{\d ^2 f(x)}{\d x_i^2} = & ~ \frac{\d}{\d x_i}(f(x) \circ (A_2 \cdot ({\bf 1}[A_1 x] \circ A_{1,i})) - f(x) \cdot \langle f(x) , A_2 \cdot ({\bf 1}[A_1 x] \circ A_{1,i}) \rangle) \\
    = & ~ \frac{\d}{\d x_i} (f(x) \circ (A_2 \cdot ({\bf 1}[A_1 x] \circ A_{1,i}))) - \frac{\d}{\d x_i} (f(x) \cdot \langle f(x) , A_2 \cdot ({\bf 1}[A_1 x] \circ A_{1,i}) \rangle) \\
     = & ~ \frac{\d f(x)}{\d x_i} \circ (A_2 \cdot ({\bf 1}[A_1 x] \circ A_{1,i})) - \frac{\d}{\d x_i} (f(x) \cdot \langle f(x) , A_2 \cdot ({\bf 1}[A_1 x] \circ A_{1,i}) \rangle) \\
    = & ~ \frac{\d f(x)}{\d x_i} \circ (A_2 \cdot ({\bf 1}[A_1 x] \circ A_{1,i})) ~-\\
    & ~ \frac{\d f(x)}{\d x_i} \cdot \langle f(x) , A_2 \cdot ({\bf 1}[A_1 x] \circ A_{1,i}) \rangle - f(x) \cdot \frac{\d \langle f(x) , A_2 \cdot ({\bf 1}[A_1 x] \circ A_{1,i}) \rangle}{\d x_i} \\
    = & ~ Q_1(x) - Q_2(x) - Q_3(x)
\end{align*}
where the first step is by {\bf Part 6} in Lemma~\ref{lem:basic_derivatives}, the second step uses differential sum rule, the third step is derived from Lemma~\ref{lem:hessian_h}, the 4th step uses differential product rule.

For the 1st item $Q_1(x)$,
\begin{align*}
    & Q_1(x) \\
    = & ~ (f(x) \circ (A_2 \cdot ({\bf 1}[A_1 x] \circ A_{1,i})) - f(x) \cdot \langle f(x) , A_2 \cdot ({\bf 1}[A_1 x] \circ A_{1,i}) \rangle) \circ (A_2 \cdot ({\bf 1}[A_1 x] \circ A_{1,i})) \\
    = & ~ f(x) \circ (A_2 \cdot ({\bf 1}[A_1 x] \circ A_{1,i})) \circ (A_2 \cdot ({\bf 1}[A_1 x] \circ A_{1,i})) ~- \\
    & ~ f(x) \circ (A_2 \cdot ({\bf 1}[A_1 x] \circ A_{1,i})) \cdot \langle f(x) , A_2 \cdot ({\bf 1}[A_1 x] \circ A_{1,i}) \rangle
\end{align*}
where the 1st step follows by {\bf Part 6} of Lemma~\ref{lem:basic_derivatives}, the 2nd step is Hadamard product calculation.

For the 2nd term $Q_2(x)$
\begin{align*}
    & Q_2(x) \\
    = & ~ (f(x) \circ (A_2 \cdot ({\bf 1}[A_1 x] \circ A_{1,i})) - f(x) \cdot \langle f(x) , A_2 \cdot ({\bf 1}[A_1 x] \circ A_{1,i}) \rangle) \cdot \langle f(x), A_2 \cdot ({\bf 1}[A_1 x] \circ A_{1,i}) \rangle \\
    = & ~ f(x) \circ (A_2 \cdot ({\bf 1}[A_1 x] \circ A_{1,i})) \cdot \langle f(x), A_2 \cdot ({\bf 1}[A_1 x] \circ A_{1,i}) \rangle ~- \\
    & ~ f(x) \cdot \langle f(x) , A_2 \cdot ({\bf 1}[A_1 x] \circ A_{1,i}) \rangle \cdot \langle f(x), A_2 \cdot ({\bf 1}[A_1 x] \circ A_{1,i}) \rangle
\end{align*}
where the 1st step is given by {\bf Part 6} in Lemma~\ref{lem:basic_derivatives}, the second step is a rearrangement.

For the thid term $Q_3(x)$
\begin{align*}
    & Q_3(x) \\
    = & ~ f(x) \cdot \frac{\d \langle f(x) , A_2 \cdot ({\bf 1}[A_1 x] \circ A_{1,i}) \rangle}{\d x_i} \\
    = & ~ f(x) \cdot  \langle \frac{\d f(x)}{\d x_i}, A_2 \cdot ({\bf 1}[A_1 x] \circ A_{1,i}) \rangle \\
    = & ~ f(x) \cdot \langle f(x) \circ (A_2 \cdot ({\bf 1}[A_1 x] \circ A_{1,i})) - f(x) \cdot \langle f(x) , A_2 \cdot ({\bf 1}[A_1 x] \circ A_{1,i}) \rangle , A_2 \cdot ({\bf 1}[A_1 x] \circ A_{1,i}) \rangle \\
    = & ~ f(x) \cdot \langle f(x) \circ (A_2 \cdot ({\bf 1}[A_1 x] \circ A_{1,i})), A_2 \cdot ({\bf 1}[A_1 x] \circ A_{1,i}) \rangle ~- \\
    & ~ f(x) \cdot \langle f(x), A_2 \cdot ({\bf 1}[A_1 x] \circ A_{1,i}) \rangle \cdot \langle f(x) , A_2 \cdot ({\bf 1}[A_1 x] \circ A_{1,i}) \rangle 
\end{align*}
where the first step is given by Lemma~\ref{lem:hessian_h}, the 2nd step uses {\bf Part 6} in Lemma~\ref{lem:basic_derivatives}, the third step is a rearrangement.

Merging $Q_1(x)$, $Q_2(x)$, $Q_3(x)$, we have
\begin{align*}
    \frac{\d^2 f(x)}{\d x_i^2} = & ~ Q_1(x) - Q_2(x) - Q_3(x) \\
    = & ~ f(x) \circ (A_2 \cdot ({\bf 1}[A_1 x] \circ A_{1,i})) \circ (A_2 \cdot ({\bf 1}[A_1 x] \circ A_{1,i})) ~- \\
    & ~ f(x) \circ (A_2 \cdot ({\bf 1}[A_1 x] \circ A_{1,i})) \cdot \langle f(x) , A_2 \cdot ({\bf 1}[A_1 x] \circ A_{1,i}) \rangle ~- \\
    & ~ f(x) \circ (A_2 \cdot ({\bf 1}[A_1 x] \circ A_{1,i})) \cdot \langle f(x), A_2 \cdot ({\bf 1}[A_1 x] \circ A_{1,i}) \rangle ~+ \\
    & ~ f(x) \cdot \langle f(x) , A_2 \cdot ({\bf 1}[A_1 x] \circ A_{1,i}) \rangle \cdot \langle f(x), A_2 \cdot ({\bf 1}[A_1 x] \circ A_{1,i}) \rangle ~- \\
    & ~ f(x) \cdot \langle f(x) \circ (A_2 \cdot ({\bf 1}[A_1 x] \circ A_{1,i})), A_2 \cdot ({\bf 1}[A_1 x] \circ A_{1,i}) \rangle ~+ \\
    & ~ f(x) \cdot \langle f(x), A_2 \cdot ({\bf 1}[A_1 x] \circ A_{1,i}) \rangle \cdot \langle f(x) , A_2 \cdot ({\bf 1}[A_1 x] \circ A_{1,i}) \rangle \\
    = & ~ f(x) \circ (A_2 \cdot ({\bf 1}[A_1 x] \circ A_{1,i})) \circ (A_2 \cdot ({\bf 1}[A_1 x] \circ A_{1,i})) ~- \\
    & ~ f(x) \circ (A_2 \cdot ({\bf 1}[A_1 x] \circ A_{1,i})) \cdot \langle f(x) , A_2 \cdot ({\bf 1}[A_1 x] \circ A_{1,i}) \rangle ~- \\
    & ~ f(x) \circ (A_2 \cdot ({\bf 1}[A_1 x] \circ A_{1,i})) \cdot \langle f(x), A_2 \cdot ({\bf 1}[A_1 x] \circ A_{1,i}) \rangle ~+ \\
    & ~ 2f(x) \cdot \langle f(x) , A_2 \cdot ({\bf 1}[A_1 x] \circ A_{1,i}) \rangle \cdot \langle f(x), A_2 \cdot ({\bf 1}[A_1 x] \circ A_{1,i}) \rangle ~- \\
    & ~ f(x) \cdot \langle f(x) \circ (A_2 \cdot ({\bf 1}[A_1 x] \circ A_{1,i})), A_2 \cdot ({\bf 1}[A_1 x] \circ A_{1,i}) \rangle
\end{align*}
where the last step merges the second term of $Q_2(x)$ and the second term of $Q_3(x)$.

{\bf Proof of Part 2}
\begin{align*}
    \frac{\d ^2 f(x)}{\d x_i \d x_j} = & ~ \frac{\d}{\d x_j}(f(x) \circ (A_2 \cdot ({\bf 1}[A_1 x] \circ A_{1,i})) - f(x) \cdot \langle f(x) , A_2 \cdot ({\bf 1}[A_1 x] \circ A_{1,i}) \rangle) \\
    = & ~ \frac{\d}{\d x_j} (f(x) \circ (A_2 \cdot ({\bf 1}[A_1 x] \circ A_{1,i}))) - \frac{\d}{\d x_j} (f(x) \cdot \langle f(x) , A_2 \cdot ({\bf 1}[A_1 x] \circ A_{1,i}) \rangle) \\
     = & ~ \frac{\d f(x)}{\d x_j} \circ (A_2 \cdot ({\bf 1}[A_1 x] \circ A_{1,i})) - \frac{\d}{\d x_j} (f(x) \cdot \langle f(x) , A_2 \cdot ({\bf 1}[A_1 x] \circ A_{1,i}) \rangle) \\
    = & ~ \frac{\d f(x)}{\d x_j} \circ (A_2 \cdot ({\bf 1}[A_1 x] \circ A_{1,i})) ~-\\
    & ~ \frac{\d f(x)}{\d x_j} \cdot \langle f(x) , A_2 \cdot ({\bf 1}[A_1 x] \circ A_{1,i}) \rangle - f(x) \cdot \frac{\d \langle f(x) , A_2 \cdot ({\bf 1}[A_1 x] \circ A_{1,i}) \rangle}{\d x_j} \\
    = & ~ Q_1(x) - Q_2(x) - Q_3(x)
\end{align*}
where the first step is by {\bf Part 6} in Lemma~\ref{lem:basic_derivatives}, the second step uses differential sum rule, the 3rd step is given by Lemma~\ref{lem:hessian_h}, the 4th step uses differential product rule.

For the 1st item $Q_1(x)$,
\begin{align*}
    & Q_1(x) \\
    = & ~ (f(x) \circ (A_2 \cdot ({\bf 1}[A_1 x] \circ A_{1,j})) - f(x) \cdot \langle f(x) , A_2 \cdot ({\bf 1}[A_1 x] \circ A_{1,j}) \rangle) \circ (A_2 \cdot ({\bf 1}[A_1 x] \circ A_{1,i})) \\
    = & ~ f(x) \circ (A_2 \cdot ({\bf 1}[A_1 x] \circ A_{1,j})) \circ (A_2 \cdot ({\bf 1}[A_1 x] \circ A_{1,i})) ~- \\
    & ~ f(x) \circ (A_2 \cdot ({\bf 1}[A_1 x] \circ A_{1,i})) \cdot \langle f(x) , A_2 \cdot ({\bf 1}[A_1 x] \circ A_{1,j}) \rangle
\end{align*}
where the initial step is by {\bf Part 6} in Lemma~\ref{lem:basic_derivatives}, the second step is Hadamard product calculation.

For the 2nd item $Q_2(x)$,
\begin{align*}
    & Q_2(x) \\
    = & ~ (f(x) \circ (A_2 \cdot ({\bf 1}[A_1 x] \circ A_{1,j})) - f(x) \cdot \langle f(x) , A_2 \cdot ({\bf 1}[A_1 x] \circ A_{1,j}) \rangle) \cdot \langle f(x), A_2 \cdot ({\bf 1}[A_1 x] \circ A_{1,i}) \rangle \\
    = & ~ f(x) \circ (A_2 \cdot ({\bf 1}[A_1 x] \circ A_{1,j})) \cdot \langle f(x), A_2 \cdot ({\bf 1}[A_1 x] \circ A_{1,i}) \rangle ~- \\
    & ~ f(x) \cdot \langle f(x) , A_2 \cdot ({\bf 1}[A_1 x] \circ A_{1,j}) \rangle \cdot \langle f(x), A_2 \cdot ({\bf 1}[A_1 x] \circ A_{1,i}) \rangle
\end{align*}
where the first step is given by {\bf Part 6} in Lemma~\ref{lem:basic_derivatives}, the second step is a rearrangement.

For the thid term $Q_3(x)$
\begin{align*}
    & Q_3(x) \\
    = & ~ f(x) \cdot \frac{\d \langle f(x) , A_2 \cdot ({\bf 1}[A_1 x] \circ A_{1,i}) \rangle}{\d x_j} \\
    = & ~ f(x) \cdot  \langle \frac{\d f(x)}{\d x_j}, A_2 \cdot ({\bf 1}[A_1 x] \circ A_{1,i}) \rangle \\
    = & ~ f(x) \cdot \langle f(x) \circ (A_2 \cdot ({\bf 1}[A_1 x] \circ A_{1,j})) - f(x) \cdot \langle f(x) , A_2 \cdot ({\bf 1}[A_1 x] \circ A_{1,j}) \rangle , A_2 \cdot ({\bf 1}[A_1 x] \circ A_{1,i}) \rangle \\
    = & ~ f(x) \cdot \langle f(x) \circ (A_2 \cdot ({\bf 1}[A_1 x] \circ A_{1,j})), A_2 \cdot ({\bf 1}[A_1 x] \circ A_{1,i}) \rangle ~- \\
    & ~ f(x) \cdot \langle f(x), A_2 \cdot ({\bf 1}[A_1 x] \circ A_{1,i}) \rangle \cdot \langle f(x) , A_2 \cdot ({\bf 1}[A_1 x] \circ A_{1,j}) \rangle 
\end{align*}
where the first step is because of Lemma~\ref{lem:hessian_h}, the second step uses {\bf Part 6} in Lemma~\ref{lem:basic_derivatives}, the 3rd step is a rearrangement.

Merging $Q_1(x)$, $Q_2(x)$, $Q_3(x)$, we have
\begin{align*}
    \frac{\d^2 f(x)}{\d x_i \d x_j}
    = & ~ Q_1(x) - Q_2(x) - Q_3(x) \\
    = & ~ f(x) \circ (A_2 \cdot ({\bf 1}[A_1 x] \circ A_{1,j})) \circ (A_2 \cdot ({\bf 1}[A_1 x] \circ A_{1,i})) ~- \\
    & ~ f(x) \circ (A_2 \cdot ({\bf 1}[A_1 x] \circ A_{1,i})) \cdot \langle f(x) , A_2 \cdot ({\bf 1}[A_1 x] \circ A_{1,j}) \rangle ~- \\
    & ~ f(x) \circ (A_2 \cdot ({\bf 1}[A_1 x] \circ A_{1,j})) \cdot \langle f(x), A_2 \cdot ({\bf 1}[A_1 x] \circ A_{1,i}) \rangle ~+ \\
    & ~ f(x) \cdot \langle f(x) , A_2 \cdot ({\bf 1}[A_1 x] \circ A_{1,j}) \rangle \cdot \langle f(x), A_2 \cdot ({\bf 1}[A_1 x] \circ A_{1,i}) \rangle ~- \\
    & ~ f(x) \cdot \langle f(x) \circ (A_2 \cdot ({\bf 1}[A_1 x] \circ A_{1,j})), A_2 \cdot ({\bf 1}[A_1 x] \circ A_{1,i}) \rangle ~+ \\
    & ~ f(x) \cdot \langle f(x), A_2 \cdot ({\bf 1}[A_1 x] \circ A_{1,i}) \rangle \cdot \langle f(x) , A_2 \cdot ({\bf 1}[A_1 x] \circ A_{1,j}) \rangle \\
    = & ~ f(x) \circ (A_2 \cdot ({\bf 1}[A_1 x] \circ A_{1,j})) \circ (A_2 \cdot ({\bf 1}[A_1 x] \circ A_{1,i})) ~- \\
    & ~ f(x) \circ (A_2 \cdot ({\bf 1}[A_1 x] \circ A_{1,i})) \cdot \langle f(x) , A_2 \cdot ({\bf 1}[A_1 x] \circ A_{1,j}) \rangle ~- \\
    & ~ f(x) \circ (A_2 \cdot ({\bf 1}[A_1 x] \circ A_{1,j})) \cdot \langle f(x), A_2 \cdot ({\bf 1}[A_1 x] \circ A_{1,i}) \rangle ~+ \\
    & ~ 2f(x) \cdot \langle f(x) , A_2 \cdot ({\bf 1}[A_1 x] \circ A_{1,j}) \rangle \cdot \langle f(x), A_2 \cdot ({\bf 1}[A_1 x] \circ A_{1,i}) \rangle ~- \\
    & ~ f(x) \cdot \langle f(x) \circ (A_2 \cdot ({\bf 1}[A_1 x] \circ A_{1,j})), A_2 \cdot ({\bf 1}[A_1 x] \circ A_{1,i}) \rangle
\end{align*}
where the last step merges the second term of $Q_2(x)$ and the second term of $Q_3(x)$.

\end{proof}
\end{lemma}

\subsection{Hessian of \texorpdfstring{$L(x)$}{}}\label{sec:hess_L}
In this section, we derive $\nabla^2 L(x)$.

\begin{lemma} [Hessian of $L(x)$] \label{lem:hessian_L} Let $L$ be defined as Definition~\ref{def:L},
\begin{itemize}
\item Part 1.
\begin{align*}
    & \frac{\d ^2 L(x)}{\d x_i^2} \\
    = & ~ \langle f(x) \circ (A_2 \cdot ({\bf 1}[A_1 x] \circ A_{1,i})), f(x) \circ (A_2 \cdot ({\bf 1}[A_1 x] \circ A_{1,i})) \rangle ~+ \\
    & ~ \langle c(x), f(x) \circ (A_2 \cdot ({\bf 1}[A_1 x] \circ A_{1,i})) \circ (A_2 \cdot ({\bf 1}[A_1 x] \circ A_{1,i})) \rangle ~- \\
    & ~ 2 \cdot \langle c(x)+f(x), f(x) \circ (A_2 \cdot ({\bf 1}[A_1 x] \circ A_{1,i})) \rangle \cdot \langle f(x) , A_2 \cdot ({\bf 1}[A_1 x] \circ A_{1,i}) \rangle ~- \\
    & ~ \langle 2c(x)+f(x), f(x) \rangle \cdot \langle f(x) , A_2 \cdot ({\bf 1}[A_1 x] \circ A_{1,i}) \rangle^2 ~- \\
    & ~ \langle c(x), f(x) \rangle \cdot \langle f(x) \circ (A_2 \cdot ({\bf 1}[A_1 x] \circ A_{1,i})), A_2 \cdot ({\bf 1}[A_1 x] \circ A_{1,i}) \rangle
\end{align*}
\item Part 2.
\begin{align*}
    & \frac{\d ^2 L(x)}{\d x_i \d x_j} \\
    = & ~ \langle f(x) \circ (A_2 \cdot ({\bf 1}[A_1 x] \circ A_{1,i})), f(x) \circ (A_2 \cdot ({\bf 1}[A_1 x] \circ A_{1,j})) \rangle ~+ \\
    & ~ \langle c(x), f(x) \circ (A_2 \cdot ({\bf 1}[A_1 x] \circ A_{1,j})) \circ (A_2 \cdot ({\bf 1}[A_1 x] \circ A_{1,i})) \rangle ~- \\
    & ~ \langle c(x)+f(x), f(x) \circ (A_2 \cdot ({\bf 1}[A_1 x] \circ A_{1,i})) \rangle \cdot \langle f(x) , A_2 \cdot ({\bf 1}[A_1 x] \circ A_{1,j}) \rangle ~- \\
    & ~ \langle c(x)+f(x), f(x) \circ (A_2 \cdot ({\bf 1}[A_1 x] \circ A_{1,j})) \rangle \cdot \langle f(x) , A_2 \cdot ({\bf 1}[A_1 x] \circ A_{1,i}) \rangle ~+ \\
    & ~ \langle 2c(x)+f(x), f(x) \rangle \cdot \langle f(x) , A_2 \cdot ({\bf 1}[A_1 x] \circ A_{1,j}) \rangle \cdot \langle f(x) , A_2 \cdot ({\bf 1}[A_1 x] \circ A_{1,i}) \rangle ~- \\
    & ~ \langle c(x), f(x) \rangle \cdot \langle f(x) \circ (A_2 \cdot ({\bf 1}[A_1 x] \circ A_{1,j})), A_2 \cdot ({\bf 1}[A_1 x] \circ A_{1,i}) \rangle
\end{align*}
\end{itemize}

\begin{proof}

{\bf Proof of Part 1}
\begin{align} \label{eq:hessian_L_1}
    & \frac{\d ^2 L(x)}{\d x_i^2} \notag\\
    = & ~ \frac{\d}{\d x_i}(\langle c(x), f(x) \circ (A_2 \cdot ({\bf 1}[A_1 x] \circ A_{1,i})) \rangle - \langle c(x), f(x) \rangle \cdot \langle f(x) , A_2 \cdot ({\bf 1}[A_1 x] \circ A_{1,i}) \rangle)\notag\\
    = & ~ \frac{\d}{\d x_i}(\langle c(x), f(x) \circ (A_2 \cdot ({\bf 1}[A_1 x] \circ A_{1,i})) \rangle) - \frac{\d}{\d x_i} (\langle c(x), f(x) \rangle \cdot \langle f(x) , A_2 \cdot ({\bf 1}[A_1 x] \circ A_{1,i}) \rangle) \notag\\
    = & ~ Q_1(x) - Q_2(x)
\end{align}
where the first step is by {\bf Part 8} of Lemma~\ref{lem:basic_derivatives}, the second step is because of differential sum rule.

For the 1st item $Q_1(x)$ in Eq.~\eqref{eq:hessian_L_1}
\begin{align} \label{eq:hessian_L_1_Q1}
    Q_1(x) = & ~ \langle \frac{\d c(x)}{\d x_i}, f(x) \circ (A_2 \cdot ({\bf 1}[A_1 x] \circ A_{1,i})) \rangle + \langle c(x),\frac{\d f(x) \circ (A_2 \cdot ({\bf 1}[A_1 x] \circ A_{1,i}))}{\d x_i} \rangle \notag \\
    = & ~ \langle \frac{\d c(x)}{\d x_i}, f(x) \circ (A_2 \cdot ({\bf 1}[A_1 x] \circ A_{1,i})) \rangle + \langle c(x),\frac{\d f(x) }{\d x_i} \circ (A_2 \cdot ({\bf 1}[A_1 x] \circ A_{1,i})) \rangle \notag \\
    = & ~ Q_{1,1}(x) + Q_{1,2}(x)
\end{align}
where the 1st step uses differential product rule, the 2nd step is because of Lemma~\ref{lem:hessian_h}.

For the first term $Q_{1,1}(x)$ of Eq.~\eqref{eq:hessian_L_1_Q1}, we have
\begin{align} \label{eq:hessian_L_1_Q11}
    & Q_{1,1}(x) \notag \\
    = & ~ \langle f(x) \circ (A_2 \cdot ({\bf 1}[A_1 x] \circ A_{1,i})) - f(x) \cdot \langle f(x) , A_2 \cdot ({\bf 1}[A_1 x] \circ A_{1,i}) \rangle, f(x) \circ (A_2 \cdot ({\bf 1}[A_1 x] \circ A_{1,i})) \rangle \notag \\
    = & ~ \langle f(x) \circ (A_2 \cdot ({\bf 1}[A_1 x] \circ A_{1,i})), f(x) \circ (A_2 \cdot ({\bf 1}[A_1 x] \circ A_{1,i})) \rangle ~- \notag \\
    & ~ \langle f(x), f(x) \circ (A_2 \cdot ({\bf 1}[A_1 x] \circ A_{1,i})) \rangle \cdot \langle f(x) , A_2 \cdot ({\bf 1}[A_1 x] \circ A_{1,i}) \rangle
\end{align}
where the first step is given by {\bf Part 7} in Lemma~\ref{lem:basic_derivatives}, the second step is inner product calculation.

For the second term $Q_{1,2}(x)$ of Eq.~\eqref{eq:hessian_L_1_Q1},
\begin{align} \label{eq:hessian_L_1_Q12}
    & Q_{1,2}(x) \notag \\
    = & ~ \langle c(x), (f(x) \circ (A_2 \cdot ({\bf 1}[A_1 x] \circ A_{1,i})) - f(x) \cdot \langle f(x) , A_2 \cdot ({\bf 1}[A_1 x] \circ A_{1,j}) \rangle) \circ (A_2 \cdot ({\bf 1}[A_1 x] \circ A_{1,i})) \rangle \notag \\
    = & ~ \langle c(x), f(x) \circ (A_2 \cdot ({\bf 1}[A_1 x] \circ A_{1,i})) \circ (A_2 \cdot ({\bf 1}[A_1 x] \circ A_{1,i})) \rangle ~- \notag \\
    & ~ \langle c(x), f(x) \circ (A_2 \cdot ({\bf 1}[A_1 x] \circ A_{1,i})) \rangle \cdot \langle f(x) , A_2 \cdot ({\bf 1}[A_1 x] \circ A_{1,i}) \rangle
\end{align}
where the 1st step is given by {\bf Part 7} of Lemma~\ref{lem:basic_derivatives}, the 2nd step is inner product calculation.

For the second item $Q_2(x)$ of Eq.~\eqref{eq:hessian_L_1},
\begin{align} \label{eq:hessian_L_1_Q2}
    Q_2(x) = & ~ \frac{\d \langle c(x), f(x) \rangle}{\d x_i} \cdot \langle f(x) , A_2 \cdot ({\bf 1}[A_1 x] \circ A_{1,i}) \rangle + \langle c(x), f(x) \rangle \cdot \frac{\d \langle f(x) , A_2 \cdot ({\bf 1}[A_1 x] \circ A_{1,i}) \rangle}{\d x_i} \notag\\
    = & ~ \frac{\d \langle c(x), f(x) \rangle}{\d x_i} \cdot \langle f(x) , A_2 \cdot ({\bf 1}[A_1 x] \circ A_{1,i}) \rangle + \langle c(x), f(x) \rangle \cdot \langle \frac{\d f(x)}{\d x_i} , A_2 \cdot ({\bf 1}[A_1 x] \circ A_{1,i}) \rangle \notag \\
    = & ~ Q_{2,1}(x) + Q_{2,2}(x)
\end{align}
where the 1st step uses differential product rule, the 2nd step is by Lemma~\ref{lem:hessian_h}.

For the first term $Q_{2,1}(x)$ of Eq.~\eqref{eq:hessian_L_1_Q2}
\begin{align} \label{eq:hessian_L_1_Q21}
    & Q_{2,1}(x) \notag \\
    = & ~ (\langle \frac{\d c(x)}{\d x_i}, f(x) \rangle + \langle c(x), \frac{\d f(x)}{\d x_i} \rangle) \cdot \langle f(x) , A_2 \cdot ({\bf 1}[A_1 x] \circ A_{1,i}) \rangle \notag \\
    = & ~ \langle c(x)+f(x), f(x) \circ (A_2 \cdot ({\bf 1}[A_1 x] \circ A_{1,i})) - f(x) \cdot \langle f(x) , A_2 \cdot ({\bf 1}[A_1 x] \circ A_{1,i}) \rangle \rangle ~\cdot \notag \\ 
    & ~ \langle f(x) , A_2 \cdot ({\bf 1}[A_1 x] \circ A_{1,i}) \rangle \notag \\
    = & ~ \langle c(x)+f(x), f(x) \circ (A_2 \cdot ({\bf 1}[A_1 x] \circ A_{1,i})) \rangle \cdot\langle f(x) , A_2 \cdot ({\bf 1}[A_1 x] \circ A_{1,i}) \rangle ~-\notag \\
    & ~ \langle c(x)+f(x), f(x) \rangle \cdot \langle f(x) , A_2 \cdot ({\bf 1}[A_1 x] \circ A_{1,i}) \rangle \cdot \langle f(x) , A_2 \cdot ({\bf 1}[A_1 x] \circ A_{1,i}) \rangle
\end{align}
where 1st step uses differential product rule, the 2nd follows by {\bf Part 6} and {\bf Part 7} in Lemma~\ref{lem:basic_derivatives}, the 3rd step is inner product calculation.

For the second term $Q_{2,2}(x)$ of Eq.~\eqref{eq:hessian_L_1_Q2}
\begin{align} \label{eq:hessian_L_1_Q22}
    & Q_{2,2}(x) \notag \\
    = & ~ \langle c(x), f(x) \rangle \cdot \langle f(x) \circ (A_2 \cdot ({\bf 1}[A_1 x] \circ A_{1,i})) - f(x) \cdot \langle f(x) , A_2 \cdot ({\bf 1}[A_1 x] \circ A_{1,i}) \rangle , A_2 \cdot ({\bf 1}[A_1 x] \circ A_{1,i}) \rangle \notag \\
    = & ~ \langle c(x), f(x) \rangle \cdot \langle f(x) \circ (A_2 \cdot ({\bf 1}[A_1 x] \circ A_{1,i})), A_2 \cdot ({\bf 1}[A_1 x] \circ A_{1,i}) \rangle \notag ~- \notag \\
    & ~ \langle c(x), f(x) \rangle \cdot \langle f(x), A_2 \cdot ({\bf 1}[A_1 x] \circ A_{1,i}) \rangle  \cdot \langle f(x) , A_2 \cdot ({\bf 1}[A_1 x] \circ A_{1,i}) \rangle
\end{align}
where the initial step is given by {\bf Part 6} in Lemma~\ref{lem:basic_derivatives}, the 2nd step is inner product calculation.

Now, putting Eq.~\eqref{eq:hessian_L_1_Q11}, Eq.~\eqref{eq:hessian_L_1_Q12}, Eq.~\eqref{eq:hessian_L_1_Q21}, Eq.~\eqref{eq:hessian_L_1_Q22} into Eq.~\eqref{eq:hessian_L_1}, we have
\begin{align*}
    & \frac{\d^2 L(x)}{\d x_i^2} \\
    = & ~ Q_{1,1}(x) + Q_{1,2}(x) - (Q_{2,1}(x) + Q_{2,2}(x)) \\
    = & ~ \langle f(x) \circ (A_2 \cdot ({\bf 1}[A_1 x] \circ A_{1,i})), f(x) \circ (A_2 \cdot ({\bf 1}[A_1 x] \circ A_{1,i})) \rangle ~- \\
    & ~ \langle  f(x), f(x) \circ (A_2 \cdot ({\bf 1}[A_1 x] \circ A_{1,i})) \rangle \cdot \langle f(x) , A_2 \cdot ({\bf 1}[A_1 x] \circ A_{1,i}) \rangle ~+\\
    & ~ \langle c(x), f(x) \circ (A_2 \cdot ({\bf 1}[A_1 x] \circ A_{1,i})) \circ (A_2 \cdot ({\bf 1}[A_1 x] \circ A_{1,i})) \rangle ~- \\
    & ~ \langle  c(x),f(x) \circ (A_2 \cdot ({\bf 1}[A_1 x] \circ A_{1,i})) \rangle \cdot \langle f(x) , A_2 \cdot ({\bf 1}[A_1 x] \circ A_{1,i}) \rangle ~- \\
    & ~ \langle c(x)+f(x), f(x) \circ ( A_2 \cdot ({\bf 1}[A_1 x] \circ A_{1,i})) \rangle \cdot \langle f(x) , A_2 \cdot ({\bf 1}[A_1 x] \circ A_{1,i}) \rangle ~+ \\
    & ~ \langle c(x)+f(x), f(x) \rangle \cdot \langle f(x) , A_2 \cdot ({\bf 1}[A_1 x] \circ A_{1,i}) \rangle \cdot \langle f(x) , A_2 \cdot ({\bf 1}[A_1 x] \circ A_{1,i}) \rangle ~- \\
    & ~ \langle c(x), f(x) \rangle \cdot \langle f(x) \circ (A_2 \cdot ({\bf 1}[A_1 x] \circ A_{1,i})), A_2 \cdot ({\bf 1}[A_1 x] \circ A_{1,i}) \rangle \notag ~+ \\
    & ~ \langle c(x), f(x) \rangle \cdot \langle f(x), A_2 \cdot ({\bf 1}[A_1 x] \circ A_{1,i}) \rangle  \cdot \langle f(x) , A_2 \cdot ({\bf 1}[A_1 x] \circ A_{1,i}) \rangle \\
    = & ~ \langle f(x) \circ (A_2 \cdot ({\bf 1}[A_1 x] \circ A_{1,i})), f(x) \circ (A_2 \cdot ({\bf 1}[A_1 x] \circ A_{1,i})) \rangle ~+ \\
    & ~ \langle c(x), f(x) \circ (A_2 \cdot ({\bf 1}[A_1 x] \circ A_{1,i})) \circ (A_2 \cdot ({\bf 1}[A_1 x] \circ A_{1,i})) \rangle ~- \\
    & ~ 2 \cdot \langle c(x)+f(x), f(x) \circ (A_2 \cdot ({\bf 1}[A_1 x] \circ A_{1,i})) \rangle \cdot \langle f(x) , A_2 \cdot ({\bf 1}[A_1 x] \circ A_{1,i}) \rangle ~- \\
    & ~ \langle 2c(x)+f(x), f(x) \rangle \cdot \langle f(x) , A_2 \cdot ({\bf 1}[A_1 x] \circ A_{1,i}) \rangle^2 ~- \\
    & ~ \langle c(x), f(x) \rangle \cdot \langle f(x) \circ (A_2 \cdot ({\bf 1}[A_1 x] \circ A_{1,i})), A_2 \cdot ({\bf 1}[A_1 x] \circ A_{1,i}) \rangle
\end{align*}
where the 3rd step merges the second term of $Q_{11}(x)$ and the second term of $Q_{12}(x)$. It also merges the second term of $Q_{21}(x)$ and the second term of $Q_{22}(x)$.

{\bf Proof of Part 2}
\begin{align} \label{eq:hessian_L_2}
    & \frac{\d ^2 L(x)}{\d x_i \d x_j} \notag\\
    = & ~ \frac{\d}{\d x_j}(\langle c(x), f(x) \circ (A_2 \cdot ({\bf 1}[A_1 x] \circ A_{1,i})) \rangle - \langle c(x), f(x) \rangle \cdot \langle f(x) , A_2 \cdot ({\bf 1}[A_1 x] \circ A_{1,i}) \rangle)\notag\\
    = & ~ \frac{\d}{\d x_j}(\langle c(x), f(x) \circ (A_2 \cdot ({\bf 1}[A_1 x] \circ A_{1,i})) \rangle) - \frac{\d}{\d x_j} (\langle c(x), f(x) \rangle \cdot \langle f(x) , A_2 \cdot ({\bf 1}[A_1 x] \circ A_{1,i}) \rangle) \notag\\
    = & ~ Q_1(x) - Q_2(x)
\end{align}
where the initial step is because of {\bf Part 8} in Lemma~\ref{lem:basic_derivatives}, the second step is given by differential sum rule.

For the first item $Q_1(x)$ in Eq.~\eqref{eq:hessian_L_2}
\begin{align} \label{eq:hessian_L_2_Q1}
    Q_1(x) = & ~ \langle \frac{\d c(x)}{\d x_j}, f(x) \circ (A_2 \cdot ({\bf 1}[A_1 x] \circ A_{1,i})) \rangle + \langle c(x),\frac{\d f(x) \circ (A_2 \cdot ({\bf 1}[A_1 x] \circ A_{1,i}))}{\d x_j} \rangle \notag \\
    = & ~ \langle \frac{\d c(x)}{\d x_j}, f(x) \circ (A_2 \cdot ({\bf 1}[A_1 x] \circ A_{1,i})) \rangle + \langle c(x),\frac{\d f(x) }{\d x_j} \circ (A_2 \cdot ({\bf 1}[A_1 x] \circ A_{1,i})) \rangle \notag \\
    = & ~ Q_{1,1}(x) + Q_{1,2}(x)
\end{align}
where the first step uses differential product rule, the 2nd step is derived from Lemma~\ref{lem:hessian_h}.

For the first term $Q_{1,1}(x)$ of Eq.~\eqref{eq:hessian_L_2_Q1}
\begin{align} \label{eq:hessian_L_2_Q11}
    & Q_{1,1}(x) \notag \\
    = & ~ \langle f(x) \circ (A_2 \cdot ({\bf 1}[A_1 x] \circ A_{1,j})) - f(x) \cdot \langle f(x) , A_2 \cdot ({\bf 1}[A_1 x] \circ A_{1,j}) \rangle, f(x) \circ (A_2 \cdot ({\bf 1}[A_1 x] \circ A_{1,i})) \rangle \notag \\
    = & ~ \langle f(x) \circ (A_2 \cdot ({\bf 1}[A_1 x] \circ A_{1,i})), f(x) \circ (A_2 \cdot ({\bf 1}[A_1 x] \circ A_{1,j})) \rangle ~- \notag \\
    & ~ \langle f(x), f(x) \circ (A_2 \cdot ({\bf 1}[A_1 x] \circ A_{1,i})) \rangle \cdot \langle f(x) , A_2 \cdot ({\bf 1}[A_1 x] \circ A_{1,j}) \rangle
\end{align}
where the first step is given by {\bf Part 7} in Lemma~\ref{lem:basic_derivatives}, the 2nd step is inner product calculation.

For the second term $Q_{1,2}(x)$ of Eq.~\eqref{eq:hessian_L_2_Q1}
\begin{align} \label{eq:hessian_L_2_Q12}
    & Q_{1,2}(x) \notag \\
    = & ~ \langle c(x), (f(x) \circ (A_2 \cdot ({\bf 1}[A_1 x] \circ A_{1,j})) - f(x) \cdot \langle f(x) , A_2 \cdot ({\bf 1}[A_1 x] \circ A_{1,j}) \rangle) \circ (A_2 \cdot ({\bf 1}[A_1 x] \circ A_{1,i})) \rangle \notag \\
    = & ~ \langle c(x), f(x) \circ (A_2 \cdot ({\bf 1}[A_1 x] \circ A_{1,j})) \circ (A_2 \cdot ({\bf 1}[A_1 x] \circ A_{1,i})) \rangle ~- \notag \\
    & ~ \langle c(x), f(x) \circ (A_2 \cdot ({\bf 1}[A_1 x] \circ A_{1,i})) \rangle \cdot \langle f(x) , A_2 \cdot ({\bf 1}[A_1 x] \circ A_{1,j}) \rangle
\end{align}
where the 1st step is given by {\bf Part 7} of Lemma~\ref{lem:basic_derivatives}, the second step is inner product calculation.

For the second item $Q_2(x)$ of Eq.~\eqref{eq:hessian_L_2}
\begin{align} \label{eq:hessian_L_2_Q2}
    Q_2(x) = & ~ \frac{\d \langle c(x), f(x) \rangle}{\d x_j} \cdot \langle f(x) , A_2 \cdot ({\bf 1}[A_1 x] \circ A_{1,i}) \rangle + \langle c(x), f(x) \rangle \cdot \frac{\d \langle f(x) , A_2 \cdot ({\bf 1}[A_1 x] \circ A_{1,i}) \rangle}{\d x_j} \notag\\
    = & ~ \frac{\d \langle c(x), f(x) \rangle}{\d x_j} \cdot \langle f(x) , A_2 \cdot ({\bf 1}[A_1 x] \circ A_{1,i}) \rangle + \langle c(x), f(x) \rangle \cdot \langle \frac{\d f(x)}{\d x_j} , A_2 \cdot ({\bf 1}[A_1 x] \circ A_{1,i}) \rangle \notag \\
    = & ~ Q_{2,1}(x) + Q_{2,2}(x)
\end{align}
where the 1st step uses differential product rule, the second step is owing to Lemma~\ref{lem:hessian_h}.

For the first term $Q_{2,1}(x)$ of Eq.~\eqref{eq:hessian_L_2_Q2},
\begin{align} \label{eq:hessian_L_2_Q21}
    & Q_{2,1}(x) \notag \\
    = & ~ (\langle \frac{\d c(x)}{\d x_j}, f(x) \rangle + \langle c(x), \frac{\d f(x)}{\d x_j} \rangle) \cdot \langle f(x) , A_2 \cdot ({\bf 1}[A_1 x] \circ A_{1,i}) \rangle \notag \\
    = & ~ \langle c(x)+f(x), f(x) \circ (A_2 \cdot ({\bf 1}[A_1 x] \circ A_{1,j})) - f(x) \cdot \langle f(x) , A_2 \cdot ({\bf 1}[A_1 x] \circ A_{1,j}) \rangle \rangle ~\cdot \notag \\ 
    & ~ \langle f(x) , A_2 \cdot ({\bf 1}[A_1 x] \circ A_{1,i}) \rangle \notag \\
    = & ~ \langle c(x)+f(x), f(x) \circ (A_2 \cdot ({\bf 1}[A_1 x] \circ A_{1,j})) \rangle \cdot\langle f(x) , A_2 \cdot ({\bf 1}[A_1 x] \circ A_{1,i}) \rangle ~-\notag \\
    & ~ \langle c(x)+f(x), f(x) \rangle \cdot \langle f(x) , A_2 \cdot ({\bf 1}[A_1 x] \circ A_{1,j}) \rangle \cdot \langle f(x) , A_2 \cdot ({\bf 1}[A_1 x] \circ A_{1,i}) \rangle
\end{align}
where 1st step uses differential product rule, the 2nd follows by {\bf Part 6} and {\bf Part 7} of Lemma~\ref{lem:basic_derivatives}, the 3rd step is inner product calculation.

For the second term $Q_{2,2}(x)$ of Eq.~\eqref{eq:hessian_L_2_Q2}, we have
\begin{align} \label{eq:hessian_L_2_Q22}
    & Q_{2,2}(x) \notag \\
    = & ~ \langle c(x), f(x) \rangle \cdot \langle f(x) \circ (A_2 \cdot ({\bf 1}[A_1 x] \circ A_{1,j})) - f(x) \cdot \langle f(x) , A_2 \cdot ({\bf 1}[A_1 x] \circ A_{1,j}) \rangle , A_2 \cdot ({\bf 1}[A_1 x] \circ A_{1,i}) \rangle \notag \\
    = & ~ \langle c(x), f(x) \rangle \cdot \langle f(x) \circ (A_2 \cdot ({\bf 1}[A_1 x] \circ A_{1,j})), A_2 \cdot ({\bf 1}[A_1 x] \circ A_{1,i}) \rangle \notag ~- \notag \\
    & ~ \langle c(x), f(x) \rangle \cdot \langle f(x), A_2 \cdot ({\bf 1}[A_1 x] \circ A_{1,i}) \rangle  \cdot \langle f(x) , A_2 \cdot ({\bf 1}[A_1 x] \circ A_{1,j}) \rangle
\end{align}
where the 1st step is given by {\bf Part 6} of Lemma~\ref{lem:basic_derivatives}, the 2nd step is inner product calculation.

Now, putting Eq.~\eqref{eq:hessian_L_2_Q11}, Eq.~\eqref{eq:hessian_L_2_Q12}, Eq.~\eqref{eq:hessian_L_2_Q21}, Eq.~\eqref{eq:hessian_L_2_Q22} into Eq.~\eqref{eq:hessian_L_2}, we have
\begin{align*}
    & \frac{\d^2 L(x)}{\d x_i \d x_j} \\
    = & ~ Q_{1,1}(x) + Q_{1,2}(x) - (Q_{2,1}(x) + Q_{2,2}(x)) \\
    = & ~ \langle f(x) \circ (A_2 \cdot ({\bf 1}[A_1 x] \circ A_{1,i})), f(x) \circ (A_2 \cdot ({\bf 1}[A_1 x] \circ A_{1,j})) \rangle ~- \\
    & ~ \langle f(x), f(x) \circ (A_2 \cdot ({\bf 1}[A_1 x] \circ A_{1,i})) \rangle \cdot \langle f(x) , A_2 \cdot ({\bf 1}[A_1 x] \circ A_{1,j}) \rangle ~+\\
    & ~ \langle c(x), f(x) \circ (A_2 \cdot ({\bf 1}[A_1 x] \circ A_{1,j})) \circ (A_2 \cdot ({\bf 1}[A_1 x] \circ A_{1,i})) \rangle ~- \\
    & ~ \langle c(x), f(x) \circ (A_2 \cdot ({\bf 1}[A_1 x] \circ A_{1,i})) \rangle \cdot \langle f(x) , A_2 \cdot ({\bf 1}[A_1 x] \circ A_{1,j}) \rangle ~- \\
    & ~ \langle c(x)+f(x), f(x) \circ (A_2 \cdot ({\bf 1}[A_1 x] \circ A_{1,j})) \rangle \cdot \langle f(x) , A_2 \cdot ({\bf 1}[A_1 x] \circ A_{1,i}) \rangle ~+ \\
    & ~ \langle c(x)+f(x), f(x) \rangle \cdot \langle f(x) , A_2 \cdot ({\bf 1}[A_1 x] \circ A_{1,j}) \rangle \cdot \langle f(x) , A_2 \cdot ({\bf 1}[A_1 x] \circ A_{1,i}) \rangle ~- \\
    & ~ \langle c(x), f(x) \rangle \cdot \langle f(x) \circ (A_2 \cdot ({\bf 1}[A_1 x] \circ A_{1,j})), A_2 \cdot ({\bf 1}[A_1 x] \circ A_{1,i}) \rangle \notag ~+ \\
    & ~ \langle c(x), f(x) \rangle \cdot \langle f(x), A_2 \cdot ({\bf 1}[A_1 x] \circ A_{1,i}) \rangle  \cdot \langle f(x) , A_2 \cdot ({\bf 1}[A_1 x] \circ A_{1,j}) \rangle \\
    = & ~ \langle f(x) \circ (A_2 \cdot ({\bf 1}[A_1 x] \circ A_{1,i})), f(x) \circ (A_2 \cdot ({\bf 1}[A_1 x] \circ A_{1,j})) \rangle ~+ \\
    & ~ \langle c(x), f(x) \circ (A_2 \cdot ({\bf 1}[A_1 x] \circ A_{1,j})) \circ (A_2 \cdot ({\bf 1}[A_1 x] \circ A_{1,i})) \rangle ~- \\
    & ~ \langle c(x)+f(x), f(x) \circ (A_2 \cdot ({\bf 1}[A_1 x] \circ A_{1,i})) \rangle \cdot \langle f(x) , A_2 \cdot ({\bf 1}[A_1 x] \circ A_{1,j}) \rangle ~- \\
    & ~ \langle c(x)+f(x), f(x) \circ (A_2 \cdot ({\bf 1}[A_1 x] \circ A_{1,j})) \rangle \cdot \langle f(x) , A_2 \cdot ({\bf 1}[A_1 x] \circ A_{1,i}) \rangle ~+ \\
    & ~ \langle 2c(x)+f(x), f(x) \rangle \cdot \langle f(x) , A_2 \cdot ({\bf 1}[A_1 x] \circ A_{1,j}) \rangle \cdot \langle f(x) , A_2 \cdot ({\bf 1}[A_1 x] \circ A_{1,i}) \rangle ~- \\
    & ~ \langle c(x), f(x) \rangle \cdot \langle f(x) \circ (A_2 \cdot ({\bf 1}[A_1 x] \circ A_{1,j})), A_2 \cdot ({\bf 1}[A_1 x] \circ A_{1,i}) \rangle 
\end{align*}
where the 3rd step merges the second term of $Q_{11}(x)$ and the second term of $Q_{12}(x)$. It also merges the second term of $Q_{21}(x)$ and the second term of $Q_{22}(x)$.
    
\end{proof}
\end{lemma}

\subsection{Decomposition of each term in Hessian of \texorpdfstring{$L(x)$}{}} \label{sec:hess_L_decompose_separate}

In this section, we simplify each term in $\nabla L(x)$.

\begin{lemma} [Helpful Lemmas] \label{lem:help} We can simplify $\nabla^2 L_\mathrm{reg}(x)$ as follows
\begin{itemize}
    \item Part 1
    \begin{align*}
        & \langle f(x) \circ (A_2 \cdot ({\bf 1}[A_1 x] \circ A_{1,i})), f(x) \circ (A_2 \cdot ({\bf 1}[A_1 x] \circ A_{1,j})) \rangle \\
        = & ~ A_{1,i}^\top \cdot \diag({\bf 1} [A_1 x]) \cdot A_2^\top \cdot  B_{1,1}(x)  \cdot A_2 \cdot \diag({\bf 1}[A_1 x]) \cdot A_{1,j}
    \end{align*}
    where $B_{1,1}(x) \in \R^{m \times m}$ satisfies
    \begin{align*}
        B_{1,1}(x) = \diag(f(x) \circ f(x)) 
    \end{align*}

    \item Part 2
    \begin{align*}
        &\langle c(x), f(x) \circ (A_2 \cdot ({\bf 1}[A_1 x]  \circ A_{1,i})) \circ (A_2 \cdot ({\bf 1}[A_1 x]  \circ A_{1,j})) \rangle \\
        = & ~ A_{1,i}^\top \cdot \diag({\bf 1}[A_1 x]) \cdot A_2^\top \cdot B_{1,2}(x) \cdot A_2 \cdot \diag({\bf 1}[A_1 x]) \cdot A_{1,j}
    \end{align*}
    where $B_{1,2}(x) \in \R^{m \times m}$ satisfies
    \begin{align*}
        B_{1,2}(x) = \diag(f(x) \circ c(x))
    \end{align*}

    \item Part 3
    \begin{align*}
        & \langle c(x) + f(x), f(x) \circ (A_2 \cdot ({\bf 1}[A_1 x]  \circ A_{1,i})) \rangle \cdot \langle f(x) , A_2 \cdot ({\bf 1}[A_1 x] \circ A_{1,j}) \rangle \\
        = & ~  A_{1,i}^\top \cdot \diag({\bf 1}[A_1 x]) \cdot  A_2^\top \cdot B_3(x) \cdot A_2 \cdot \diag({\bf 1}[A_1 x]) A_{1,j}
    \end{align*}
    where $B_3(x) \in \R^{m \times m}$ satisfies
    \begin{align*}
        B_2(x) = \diag(f(x)) \cdot (c(x)+f(x)) \cdot f(x)^\top
    \end{align*}

    \item Part 4
    \begin{align*}
         & \langle c(x) + f(x), f(x) \circ (A_2 \cdot ({\bf 1}[A_1 x]  \circ A_{1,j})) \rangle \cdot \langle f(x) , A_2 \cdot ({\bf 1}[A_1 x] \circ A_{1,i}) \rangle \\
         = & ~  A_{1,i}^\top \cdot \diag({\bf 1}[A_1 x]) \cdot A_2^\top \cdot B_3(x) \cdot A_2 \cdot \diag({\bf 1}[A_1 x])\cdot A_{1,j}
    \end{align*}
    where $B_3(x) \in \R^{m \times m}$ satisfies
    \begin{align*}
        B_3(x) =  f(x) \cdot (c(x) + f(x))^\top \cdot \diag(f(x)) 
    \end{align*}

    \item Part 5
    \begin{align*}
        & \langle 2c(x)+f(x), f(x) \rangle \cdot \langle f(x) , A_2 \cdot ({\bf 1}[A_1 x] \circ A_{1,j}) \rangle \cdot \langle f(x) , A_2 \cdot ({\bf 1}[A_1 x] \circ A_{1,i}) \rangle \\
        = & ~ A_{1,i}^\top \cdot \diag({\bf 1}[A_1 x]) \cdot A_2^\top \cdot B_4(x) \cdot A_2 \cdot \diag({\bf 1}[A_1 x]) \cdot A_{1,j}
    \end{align*}
    where $B_4(x) \in \R^{m \times m}$ satisfies
    \begin{align*}
        B_4(x) = \langle 2c(x)+f(x), f(x) \rangle \cdot  f(x) \cdot  f(x)^\top
    \end{align*}

    \item Part 6
    \begin{align*}
        & \langle c(x), f(x) \rangle \cdot \langle f(x) \circ (A_2 \cdot ({\bf 1}[A_1 x] \circ A_{1,j})), A_2 \cdot ({\bf 1}[A_1 x] \circ A_{1,i}) \rangle \\
        = & ~ A_{1,i}^\top \cdot \diag({\bf 1}[A_1 x]) \cdot A_2^\top \cdot B_5(x) \cdot A_2 \cdot \diag({\bf 1}[A_1 x]) \cdot A_{1,j}
    \end{align*}
    where $B_5(x) \in \R^{m \times m}$ satisfies
    \begin{align*}
        B_5(x) = \langle c(x), f(x) \rangle \cdot  \diag(f(x))
    \end{align*}
    
\end{itemize}

\begin{proof}
{\bf Verification of Part 1}
\begin{align*}
    & \langle f(x) \circ (A_2 \cdot ({\bf 1}[A_1 x] \circ A_{1,i})), f(x) \circ (A_2 \cdot ({\bf 1}[A_1 x] \circ A_{1,j})) \rangle \\
    = & ~ (f(x) \circ (A_2 \cdot ({\bf 1}[A_1 x] \circ A_{1,i})))^\top (f(x) \circ (A_2 \cdot ({\bf 1}[A_1 x] \circ A_{1,j}))) \\
    = & ~ (\diag(f(x)) \cdot A_2 \cdot \diag({\bf 1}[A_1 x]) \cdot A_{1,i})^\top \cdot (\diag(f(x)) \cdot A_2 \cdot \diag({\bf 1}[A_1 x]) \cdot A_{1,j}) \\
    = & ~ \underbrace{ A_{1,i}^\top }_{ 1 \times n} \cdot \underbrace{ \diag({\bf 1} [A_1 x]) }_{ n \times n } \cdot \underbrace{ A_2^\top }_{n \times m} \cdot \underbrace{ \diag(f(x)) }_{m \times m} \cdot \underbrace{\diag(f(x)) }_{m \times m} \cdot \underbrace{A_2}_{m \times n} \cdot \underbrace{\diag({\bf 1}[A_1 x])}_{n \times n} \cdot \underbrace{A_{1,j}}_{n \times 1} \\
    = & ~ \underbrace{ A_{1,i}^\top }_{ 1 \times n} \cdot \underbrace{ \diag({\bf 1} [A_1 x]) }_{ n \times n } \cdot \underbrace{ A_2^\top }_{n \times m} \cdot \underbrace{ \diag(f(x) \circ f(x)) }_{m \times m} \cdot \underbrace{A_2}_{m \times n} \cdot \underbrace{\diag({\bf 1}[A_1 x])}_{n \times n} \cdot \underbrace{A_{1,j}}_{n \times 1}
\end{align*}
where the 1st step is due to the definition of inner product, the 2nd step uses Fact~\ref{fac:hadamard_product_algebra}
, the 3rd step is a rearrangement, the last step is given by Fact~\ref{fac:hadamard_product_algebra}.

{\bf Proof of Part 2}
\begin{align*}
    & \langle c(x), f(x) \circ ((A_2 \cdot ({\bf 1}[A_1 x]  \circ A_{1,i})) \circ (A_2 \cdot ({\bf 1}[A_1 x]  \circ A_{1,j})) \rangle \\
    = & ~ \langle A_2 \cdot ({\bf 1}[A_1 x]  \circ A_{1,i}) , f(x) \circ c(x) \circ (A_2 \cdot ({\bf 1}[A_1 x]  \circ A_{1,j})) \rangle \\
    = & ~ (A_2 \cdot ({\bf 1}[A_1 x]  \circ A_{1,i}))^\top \cdot (f(x) \circ c(x) \circ (A_2 \cdot ({\bf 1}[A_1 x]  \circ A_{1,j})) ) \\
    = & ~ (A_2 \cdot \diag({\bf 1}[A_1 x]) \cdot A_{1,i})^\top \cdot \diag(f(x) \circ c(x)) \cdot A_2 \cdot \diag({\bf 1}[A_1 x]) \cdot A_{1,j} \\
    = & ~ \underbrace{A_{1,i}^\top}_{1 \times n} \cdot \underbrace{\diag({\bf 1}[A_1 x])}_{n \times n} \cdot \underbrace{A_2^\top}_{n \times m} \cdot \underbrace{\diag(f(x) \circ c(x))}_{m \times m} \cdot \underbrace{A_2}_{m \times n} \cdot \underbrace{\diag({\bf 1}[A_1 x])}_{n \times n} \cdot \underbrace{A_{1,j}}_{n \times 1}
\end{align*}
where the 1st step is from Fact~\ref{fac:hadamard_product_algebra}, the 2nd step follows by definition of inner product, the 3rd step is from Fact~\ref{fac:hadamard_product_algebra}, the 4th step is due to property of matrix transpose.

{\bf Proof of Part 3}
\begin{align*}
    & \langle c(x)+f(x), f(x) \circ (A_2 \cdot ({\bf 1}[A_1 x]  \circ A_{1,i})) \rangle \cdot \langle f(x) , A_2 \cdot ({\bf 1}[A_1 x] \circ A_{1,j}) \rangle \\
    = & ~ (f(x) \circ (A_2 \cdot ({\bf 1}[A_1 x]  \circ A_{1,i})))^\top \cdot (c(x)+f(x)) \cdot f(x)^\top \cdot (A_2 \cdot ({\bf 1}[A_1 x] \circ A_{1,j})) \\
    = & ~ (\diag(f(x)) \cdot A_2 \cdot \diag({\bf 1}[A_1 x])  \cdot A_{1,i})^\top \cdot (c(x)+f(x)) \cdot f(x)^\top \cdot (A_2 \cdot \diag({\bf 1}[A_1 x]) \cdot A_{1,j}) \\
    = & ~  \underbrace{A_{1,i}^\top}_{1 \times n} \cdot \underbrace{\diag({\bf 1}[A_1 x])}_{n \times n} \cdot  \underbrace{A_2^\top}_{n \times m} \cdot \underbrace{\diag(f(x))}_{m \times m} \cdot \underbrace{(c(x)+f(x))}_{m \times 1} \cdot \underbrace{f(x)^\top}_{1 \times m} \cdot \underbrace{A_2 }_{m \times n} \cdot \underbrace{\diag({\bf 1}[A_1 x])}_{n \times n} \cdot \underbrace{A_{1,j}}_{n \times 1}
\end{align*}
where the first step is given by the definition of inner product, the second step is derived by Fact~\ref{fac:hadamard_product_algebra}, the third step is from property of matrix transpose.

{\bf Proof of Part 4}
\begin{align*}
    & \langle c(x) + f(x), f(x) \circ (A_2 \cdot ({\bf 1}[A_1 x]  \circ A_{1,j})) \rangle \cdot \langle f(x) , A_2 \cdot ({\bf 1}[A_1 x] \circ A_{1,i}) \rangle \\
    = & ~ (A_2 \cdot ({\bf 1}[A_1 x] \circ A_{1,i}))^\top \cdot f(x) \cdot (c(x) + f(x))^\top \cdot (f(x) \circ (A_2 \cdot ({\bf 1}[A_1 x]  \circ A_{1,j}))) \\
    = & ~ (A_2 \cdot \diag({\bf 1}[A_1 x]) \cdot A_{1,i})^\top \cdot f(x) \cdot (c(x) + f(x))^\top \cdot \diag(f(x)) \cdot A_2 \cdot \diag({\bf 1}[A_1 x]) \cdot A_{1,j} \\
    = & ~ \underbrace{A_{1,i}^\top}_{1 \times n} \cdot \underbrace{\diag({\bf 1}[A_1 x])}_{n \times n} \cdot \underbrace{A_2^\top}_{n \times m} \cdot \underbrace{f(x)}_{m \times 1} \cdot \underbrace{(c(x) + f(x))^\top}_{1 \times m} \cdot \underbrace{\diag(f(x))}_{m \times m} \cdot \underbrace{A_2}_{m \times n} \cdot \underbrace{\diag({\bf 1}[A_1 x])}_{n \times n} \cdot \underbrace{A_{1,j}}_{n \times 1}
\end{align*}
where the 1st step is due to the definition of inner product, the 2nd step is from Fact~\ref{fac:hadamard_product_algebra}, the 3rd step follows by property of matrix transpose.

{\bf Proof of Part 5}
\begin{align*}
    & \langle 2c(x)+f(x), f(x) \rangle \cdot \langle f(x) , A_2 \cdot ({\bf 1}[A_1 x] \circ A_{1,j}) \rangle \cdot \langle f(x) , A_2 \cdot ({\bf 1}[A_1 x] \circ A_{1,i}) \rangle \\
    = & ~ (A_2 \cdot ({\bf 1}[A_1 x] \circ A_{1,i})^\top \cdot f(x) \cdot (2c(x)+f(x))^\top f(x) \cdot f(x)^\top \cdot (A_2 \cdot ({\bf 1}[A_1 x] \circ A_{1,j}) \\
    = & ~ (A_2 \cdot \diag({\bf 1}[A_1 x]) \cdot A_{1,i})^\top \cdot f(x) \cdot (2c(x)+f(x))^\top f(x) \cdot f(x)^\top \cdot A_2 \cdot \diag({\bf 1}[A_1 x]) \cdot A_{1,j} \\
    = & ~ \underbrace{A_{1,i}^\top}_{1 \times n} \cdot \underbrace{\diag({\bf 1}[A_1 x])}_{n \times n} \cdot \underbrace{A_2^\top}_{n \times m} \cdot \underbrace{f(x)}_{m \times 1} \cdot \underbrace{(2c(x)+f(x))^\top f(x)}_{\mathrm{scalar}} \cdot \underbrace{f(x)^\top}_{1 \times m} \cdot \underbrace{A_2}_{m \times n} \cdot \underbrace{\diag({\bf 1}[A_1 x])}_{n \times n} \cdot \underbrace{A_{1,j}}_{n \times 1} \\
\end{align*}
where the initial step is given by the definition of inner product, the second step is  by Fact~\ref{fac:hadamard_product_algebra}, the third step holds by property of matrix transpose.

{\bf Proof of Part 6}
\begin{align*}
    & \langle c(x), f(x) \rangle \cdot \langle f(x) \circ (A_2 \cdot ({\bf 1}[A_1 x] \circ A_{1,j})), A_2 \cdot ({\bf 1}[A_1 x] \circ A_{1,i}) \rangle \\
    = & ~ (f(x) \circ (A_2 \cdot ({\bf 1}[A_1 x] \circ A_{1,j})))^\top \cdot c(x)^\top f(x) \cdot A_2 \cdot ({\bf 1}[A_1 x] \circ A_{1,i}) \\
    = & ~ (\diag(f(x)) \cdot A_2 \cdot \diag({\bf 1}[A_1 x]) \cdot A_{1,j})^\top \cdot c(x)^\top f(x) \cdot A_2 \cdot \diag({\bf 1}[A_1 x]) \cdot A_{1,i} \\
    = & ~ A_{1,j}^\top \cdot \diag({\bf 1}[A_1 x]) \cdot A_2^\top \cdot \diag(f(x)) \cdot c(x)^\top f(x) \cdot A_2 \cdot \diag({\bf 1}[A_1 x]) \cdot A_{1,i} \\
    = & ~ \underbrace{A_{1,i}^\top}_{1 \times n} \cdot \underbrace{\diag({\bf 1}[A_1 x])}_{n \times n} \cdot \underbrace{A_2^\top}_{n \times m} \cdot \underbrace{\diag(f(x))}_{m \times m} \cdot \underbrace{c(x)^\top f(x)}_{ \mathrm{scalar} } \cdot \underbrace{A_2}_{m \times n} \cdot \underbrace{\diag({\bf 1}[A_1 x])}_{n \times n} \cdot \underbrace{A_{1,j}}_{n \times 1}
\end{align*}
where the 1st step is due to the definition of inner product, the 2nd step is from Fact~\ref{fac:hadamard_product_algebra}, the 3rd step follows by property of matrix transpose, the 4th step holds by taking transpose of the formula in step 3.

\end{proof}
\end{lemma}

\subsection{Decomposition of Hessian of \texorpdfstring{$L(x)$}{}} \label{sec:hess_L_decompose}

Now, we summarize Lemma~\ref{lem:help} and state the decomposition of $\nabla^2 L(x)$.
\begin{lemma} [Decomposition of Hessian of $L(x)$] \label{lem:decomposition_L}
\begin{align*}
    \frac{\d^2 L(x)}{\d x_i \d x_j} = \underbrace{A_{1,i}^\top}_{1 \times n} \cdot \underbrace{ \diag({\bf 1} [A_1 x]) }_{n \times n} \cdot \underbrace{ A_2^\top }_{n \times m} \cdot \underbrace{ B(x) }_{m \times m} \cdot \underbrace{ A_2 }_{m \times n} \cdot \underbrace{ \diag({\bf 1}[A_1 x]) }_{n \times n} \cdot \underbrace{A_{1,j}}_{n \times 1}
\end{align*}
where $B(x) \in \R^{m \times m}$ satisfies
\begin{itemize}
    \item $B(x) = B_1(x) - B_2(x) - B_3(x) + B_4(x) - B_5(x)$
    \item $B_1(x) = B_{1,1}(x) + B_{1,2}(x) = \diag(f(x) \circ (f(x)+c(x))) $, 
    \item $B_2(x) =  \diag(f(x)) \cdot (c(x)+f(x)) \cdot f(x)^\top $
    \item $B_3(x) =  f(x) \cdot (c(x) + f(x))^\top \cdot \diag(f(x))  $
    \item $B_4(x) = \langle 2c(x)+f(x), f(x) \rangle \cdot f(x) \cdot  f(x)^\top$
    \item $B_5(x) = \langle c(x), f(x) \rangle \cdot \diag(f(x))$ 
\end{itemize}
\end{lemma}

%% file: bound.tex
\section{Bounds for basic functions} \label{sec:bound}
In the following three sections, we aim to verify the Lipschitz continuity $\nabla^2 L(x)$. Again, we prove the Lipschitz of some auxiliary functions at first, and then combine them to generate the final result. The core idea of this proof agrees with previous analysis on recurrent neural network (RNN) and transformer-inspired regression models \cite{azls19a,lsz23,gsx23}. 

In this section, we derive some bounds for basic functions, including upper bound on $\|h(x)\|_2$ (see Section~\ref{sec:upper_bound_h}), upper bound on $\|u(x)\|_2$ (see Section~\ref{sec:upper_bound_u}), upper bound on gradient of $u(x)$, lower bound on $\beta$ (see Section~\ref{sec:lower_bound_beta}), upper bound on $\|f(x)\|_2, \|c(x)\|_2$ (see Section~\ref{sec:upper_bound_f_c}), and upper bound on $\| B(x) \|$ (see Section~\ref{sec:upper_bound_B}). The bounds will be used to derive the Lipschitz condition of $L(x)$.

\subsection{Upper bound on \texorpdfstring{$ \|h(x)\|_2 $}{}} \label{sec:upper_bound_h}
In this section, we give an upper bound on $\| h(x) \|_2$.

\begin{lemma} \label{lem:upper_bound_h}
If the following condition holds,
\begin{itemize}
    \item Let $A_1 \in \R^{n \times d}$ be a matrix with $\| A_1 \| \leq R$
    \item Let $x \in \R^n$ be a vector with $\| x \|_2 \leq R$
    \item Suppose $h$ is defined as Definition~\ref{def:h}
\end{itemize}
we have,
\begin{align*}
    \| h(x) \|_2 \leq R^2
\end{align*}

\begin{proof}
\begin{align*}
    \| h(x) \|_2 = & ~ \| \phi(A_1x) \|_2 \\
    \leq & ~ \| A_1 x \|_2 \\
    \leq & ~ \| A_1 \| \cdot \| x \|_2 \\
    \leq & ~ R^2
\end{align*}
where the 1st step is given by definition of $h(x)$ (see Definition~\ref{def:h}), the 2nd step and the third step are from Fact~\ref{fac:matrix_algebra}, the last step follows by assumptions of this lemma.
\end{proof}

\end{lemma}

\subsection{Upper bound on \texorpdfstring{$\|u(x)\|_2$}{}} \label{sec:upper_bound_u}
In this section, we state an upper bound on $\| u(x) \|_2$.

\begin{lemma} \label{lem:upper_bound_u}
If the following condition holds,
\begin{itemize}
    \item Let $A_1 \in \R^{n \times d}$ be a matrix with $\| A_1 \| \leq R$
    \item Let $x \in \R^n$ be a vector with $\| x \|_2 \leq R$
    \item Suppose $u$ is defined as Definition~\ref{def:h}
\end{itemize}
Then we have,
\begin{align*}
    \| u(x) \|_2 \leq \sqrt{m} \exp(R^3)
\end{align*}

\begin{proof}
\begin{align*}
    \| u(x) \|_2 = & ~ \| \exp(A_2 \cdot h(x)) \|_2 \\
    \leq & ~ \sqrt{m} \| \exp(A_2 \cdot h(x)) \|_\infty \\
    \leq & ~ \sqrt{m} \exp( \|A_2 \cdot h(x) \|_2) \\
    \leq & ~ \sqrt{m} \exp( \|A_2 \| \cdot \| h(x) \|_2) \\
    \leq & ~ \sqrt{m} \exp( R \cdot \| h(x) \|_2) \\
    \leq & ~ \sqrt{m} \exp( R^3 )
\end{align*}
where the 1st step is due to definition of $u(x)$ (see Definition~\ref{def:u}), the 2nd step and the 3rd step are from Fact~\ref{fac:vector_norm}, the 4th step follows by Fact~\ref{fac:matrix_algebra}, the 5th step holds since $\| A_2 \| \leq R$ the last step follows by Lemma~\ref{lem:upper_bound_h}.
\end{proof}

\end{lemma}

\subsection{Upper bound on gradient of \texorpdfstring{$\|u(x)\|_2$}{}} \label{sec:upper_bound_grad_u}
In this section, we compute an upper bound on gradient of $\| u(x) \|_2$.

\begin{lemma} \label{lem:upper_bound_grad_u}
If the following condition holds,
\begin{itemize}
    \item Let $A_1 \in \R^{n \times d}, A_2 \in \R^{m \times n}$ be a matrix with $\| A_1 \| \leq R, \| A_2 \| \leq R$
    \item Let $x \in \R^n$ be a vector with $\| x \|_2 \leq R$
    \item Let $u(x)$ be defined as Definition~\ref{def:u}
    \item Let $u'(x)$ be a matrix which its $(i,j)$-th term is $\frac{\d u(x)_j}{\d x_i}$.
\end{itemize}
Then we have,
\begin{itemize}
    \item Part 1 $\| \frac{\d u(x)}{\d x_i} \|_2 \leq \sqrt{mn} R^2 \exp(R^3)$
    \item Part 2 $\| u'(x) \| \leq \sqrt{mnd} R^2 \exp(R^3)$
\end{itemize}

\begin{proof}
{\bf Proof of Part 1}
\begin{align*}
    \| \frac{\d u(x)}{\d x_i} \|_2 = & ~ \| u(x) \circ (A_2 \cdot ( {\bf 1}[A_1 x] \circ A_{1,i})) \|_2 \\
    \leq & \| u(x) \|_2 \cdot \| A_2  \| \cdot \| {\bf 1}[A_1 x] \|_2 \cdot \| A_{1,i}\|_2 \\
    \leq & \sqrt{m} \exp(R^3) \cdot \| A_2  \| \cdot \| {\bf 1}[A_1 x] \|_2 \cdot \| A_{1,i} \|_2 \\
    \leq & \sqrt{m} \exp(R^3) R^2 \cdot \| {\bf 1}[A_1 x] \|_2 \\
    \leq & \sqrt{mn} \exp(R^3) R^2 \cdot \| {\bf 1}[A_1 x] \|_\infty \\
    \leq & \sqrt{mn} R^2 \exp(R^3)
\end{align*}
where the 1st step is derived by {\bf Part 3} in Lemma~\ref{lem:basic_derivatives}, the 2nd step is from Fact~\ref{fac:vector_norm}, the 3rd step is due to Lemma~\ref{lem:upper_bound_u}, the 4th step holds since $\| A_2 \| \leq R$, the 5th step is from Fact~\ref{fac:vector_norm}, the last step follows by $\| {\bf 1}[A_1 x] \|_\infty \leq 1$.
\end{proof}

{\bf Part 2}
\begin{align*}
    \| u'(x) \| \leq & ~ \| u'(x) \|_F \\
    = & ~ (\sum_{i = 1}^d \| \frac{\d u(x)}{\d x_i} \|_2^2)^{\frac{1}{2}} \\
    \leq & ~ \sqrt{d} \cdot \sqrt{mn} R^2 \exp(R^3)
\end{align*}
where the first step holds since Fact~\ref{fac:matrix_algebra}, the second step is by the definition of Frobenius norm, the 3rd step is by {\bf Part 1}.

\end{lemma}

\subsection{Lower bound on \texorpdfstring{$\beta$}{}} \label{sec:lower_bound_beta}
We present an lower bound on $\beta$ in this section, which is the infimum of $\| u(x) \|_2$.

\begin{lemma} \label{lem:lower_bound_beta}
If the following condition holds,
\begin{itemize}
    \item Let $A_1 \in \R^{n \times d}, A_2 \in \R^{m \times n}$ satisfy $\| A_1 \| \leq R, \| A_2 \| \leq R$
    \item Let $x \in \R^n$ be a vector with $\| x \|_2 \leq R$
    \item Suppose $\alpha$ is defined as Definition~\ref{def:alpha}
    \item We use $\beta$ to denote the greatest lower bound of $\alpha(x)$
\end{itemize}
we have,
\begin{align*}
    \beta \geq \exp(-R^3)
\end{align*}

\begin{proof}
    The proof is similar to the proof of Lemma 8.9 in \cite{dls23}.
\end{proof}
\end{lemma}

\subsection{Upper bound on \texorpdfstring{$\| f(x) \|_2$}{} and \texorpdfstring{$\|c(x)\|_2$}{}} \label{sec:upper_bound_f_c}
In this section, we state an upper bound on $\| f(x) \|_2$ and $\|c(x)\|_2$.

\begin{lemma} [Lemma 6.5 and 6.7 in \cite{gsx23}] \label{lem:upper_bound_f_c}
If the following condition holds,
\begin{itemize}
    \item Suppose $f$ is defined as Definition~\ref{def:f}
    \item Suppose $c$ is defined as Definition~\ref{def:c}
    \item $\| b \|_2 \leq 1$
\end{itemize}
we have,
\begin{align*}
    \| f(x) \|_2 \leq & ~ 1 \\
    \| c(x) \|_2 \leq & ~ 2
\end{align*}
\end{lemma}

\subsection{Upper bound on \texorpdfstring{$\| B(x) \|_2$}{} } \label{sec:upper_bound_B}
In this section, we state an upper bound on $\| B(x) \|_2$. We verify the result by bounding each term in $B(x)$

\begin{lemma} \label{lem:upper_bound_B}
Under following conditions,
\begin{itemize}
    \item Suppose $B(x)$ is defined as in Lemma~\ref{lem:decomposition_L}
    \item $\| b \|_2 \leq 1$
\end{itemize}
we have,
\begin{align*}
    \| B(x) \| \leq & ~ 16 
\end{align*}

\begin{proof}
We bound each term in $B(x)$ separately.

{\bf Term 1}
\begin{align*}
    \| B_1(x) \| = \| \diag(f(x) \circ (f(x)+c(x))) \| \leq & ~ \| f(x) \circ (f(x)+c(x)) \|_2 \\
    \leq & ~ \| f(x) \|_2 \cdot \| f(x)+c(x) \|_2 \\
    \leq & ~ 3
\end{align*}
the initial step is by Fact~\ref{fac:matrix_algebra}, the second step is owing to Fact~\ref{fac:vector_norm}, the third step follows by Lemma~\ref{lem:upper_bound_f_c}.

{\bf Term 2}
\begin{align*}
    \| B_2(x) \| = \| \diag(f(x)) \cdot (c(x)+f(x)) \cdot f(x)^\top \| \leq & ~ \| \diag(f(x)) \| \cdot \| c(x)+f(x) \|_2 \cdot  \| f(x) \|_2 \\
    \leq & ~ \| f(x) \|_2 \cdot \| c(x)+f(x) \|_2 \cdot  \| f(x) \|_2 \\
    \leq & ~ 3
\end{align*}
where the first step and the second step holds because of Fact~\ref{fac:matrix_algebra}, the last step uses Lemma~\ref{lem:upper_bound_f_c}. 

{\bf Term 3}
The third term has the same bound as the second term since they are symmetric.

{\bf Term 4}
\begin{align*}
    \| B_4(x) \| = \| \langle 2c(x)+f(x), f(x) \rangle \cdot f(x) \cdot  f(x)^\top \| \leq & ~ | \langle 2c(x)+f(x), f(x) \rangle | \cdot \| f(x) \|_2 \cdot \| f(x)\|_2 \\
    \leq & ~ \| 2c(x)+f(x) \|_2 \cdot \| f(x) \|_2 \cdot \| f(x) \|_2 \cdot \| f(x)\|_2 \\
    \leq & ~ 5
\end{align*}
the initial step is given by Fact~\ref{fac:matrix_algebra}, the second step is by Cauchy-Schwartz inequality, the 3rd step uses Lemma~\ref{lem:upper_bound_f_c}.

{\bf Term 5}
\begin{align*}
    \| B_5(x) \| = \| \langle c(x), f(x) \rangle \cdot \diag(f(x)) \| = & ~ | \langle c(x), f(x) \rangle| \cdot \| \diag(f(x)) \| \\
    \leq & ~ \| c(x) \|_2 \cdot \| f(x) \|_2 \cdot \| \diag(f(x)) \| \\
    \leq & ~ \| c(x) \|_2 \cdot \| f(x) \|_2 \cdot \| f(x) \|_2 \\
    \leq & ~ 2
\end{align*}
where the 2nd step is from Cauchy-Schwartz inequality, the 3rd step is given by Fact~\ref{fac:matrix_algebra}, the 4th step is due to Lemma~\ref{lem:upper_bound_f_c}.

Using triangle inequality, we can compute an upper bound for $\| B(x) \|$
\begin{align*}
    \| B(x) \| \leq \| B_1(x) \| + \| B_2(x) \| + \| B_3(x) \| + \| B_4(x) \| + \| B_5(x) \| \leq 16
\end{align*}
\end{proof}
\end{lemma}

%% file: lip.tex
\section{Lipschitz of basic functions} \label{sec:lipschitz_basic}
We calculate the Lipschitz condition for basic functions in this section, including $h(x)$ (see Section~\ref{sec:lipschitz_h}), $u(x)$ (see Section~\ref{sec:lipschitz_u}), $\alpha(x)$ (see Section~\ref{sec:lipschitz_alpha}), $\alpha^{-1}(x)$ (see Section~\ref{sec:lipschitz_alpha_inverse}), and $f(x), c(x)$ (see Section~\ref{sec:lipschitz_f_c}). The main technique we utilized in this section is Mean Value Theorem for vector function (Fact~\ref{fac:mvt}) and triangle inequality.

\paragraph{$Remark$} In this section we assume that
\begin{itemize}
    \item Let $A_1 \in \R^{n \times d}, A_2 \in \R^{m \times n}$ be matrices with $\| A_1 \| \leq R$, $\| A_2 \| \leq R$
    \item Let $x,y \in \R^d, b \in \R^m$ be vectors with $\| x \|_2 \leq R, \| y \|_2 \leq R, \|b\|_2 \leq 1$, $R>4$
\end{itemize}
which is Assumption~\ref{ass:bounded_parameters} in Section~\ref{sec:cov_variable:informal}.

\subsection{Lipschitz of function \texorpdfstring{$h(x)$}{}} \label{sec:lipschitz_h}
We state the Lipschitz constant of $h(x)$ in this section.

\begin{lemma} \label{lem:lipschitz_h}
We define $h(x)$ to be Definition~\ref{def:h},
\begin{align*}
    \| h(x) - h(y) \|_2 \leq R \cdot \| x - y \|_2
\end{align*}

\begin{proof}
\iffalse
We consider the $i$-th term of $\| h(x) - h(y) \|_2^2$. Let $a_i$ denotes the $i$-th row of matrix $A_1$. Let $h(x)_i$ denotes the $i$-th term of $h(x)$. Notice that $h(x) = \phi (A_1 x)$ by Definition~\ref{def:h}.

If $a_ix >0, a_i y > 0$, we have
\begin{align*}
    | h(x)_i - h(y)_i | = | a_i x - a_i y |
\end{align*}

If $a_ix > 0, a_i y \leq 0$, we have
\begin{align*}
    | h(x)_i - h(y)_i | = | a_i x - 0 | \leq | a_ix - a_iy|
\end{align*}

If $a_ix \leq 0, a_i y > 0$, we have
\begin{align*}
    | h(x)_i - h(y)_i | = | 0 - a_iy | \leq | a_ix - a_iy|
\end{align*}

If $a_ix \leq 0, a_i y \leq 0$, we have
\begin{align*}
    | h(x)_i - h(y)_i | = |0 - 0| \leq |a_ix - a_iy|
\end{align*}

Therefore, for all $i \in [n]$, we have
\begin{align*}
    | h(x)_i - h(y)_i | \leq |a_ix - a_iy|
\end{align*}
\fi
We have
\begin{align*}
    \| h(x) - h(y) \|_2 = & ~ \| \phi(A_1 x) - \phi(A_1 y) \|_2 \\
    \leq & ~ \| A_1(x-y) \|_2 \\
    \leq & ~ \| A_1 \| \cdot \| x - y \|_2 \\
    \leq & ~ R \cdot \| x - y \|_2
\end{align*}
where the 1st step is given by the definition of $h(x)$ (see Definition~\ref{def:h}), the second step and the 3rd step follow by Fact~\ref{fac:matrix_algebra}, the last step holds because $\| A_1 \| \leq R$.
\end{proof}
\end{lemma}

\subsection{Lipschitz of function \texorpdfstring{$u(x)$}{}}\label{sec:lipschitz_u}
We state the Lipschitz constant of $u(x)$ in this section.

\begin{lemma} \label{lem:lipschitz_u}
Suppose $u(x)$ is defined as Definition~\ref{def:u},
Then we have,
\begin{align*}
    \| u(x) - u(y) \|_2 \leq R^2 \sqrt{mnd} \exp(R^3) \cdot \| x - y \|_2
\end{align*}

\begin{proof}
The result directly from Mean Value Theorem (see Fact~\ref{fac:mvt} and upper bound of $u'(x)$ (see Lemma~\ref{lem:upper_bound_grad_u}).
   
\end{proof}

\end{lemma}

\subsection{Lipschitz of function \texorpdfstring{$\alpha(x)$}{}}\label{sec:lipschitz_alpha}
We compute the Lipschitz constant of $\alpha(x)$ in this section.

\begin{lemma} \label{lem:lipschitz_alpha}
Suppose $\alpha(x)$ is defined as Definition~\ref{def:alpha},
\begin{align*}
    | \alpha(x) - \alpha(y) | \leq R^2 m \sqrt{nd} \exp(R^3) \cdot \| x - y \|_2
\end{align*}

\begin{proof}
\begin{align*}
    | \alpha(x) - \alpha(y) | = & ~ \| \langle u(x), {\bf 1}_m \rangle - \langle u(y) , {\bf 1}_m \rangle \|_2 \\
    \leq & ~ \sqrt{m} \cdot \| u(x) - u(y) \|_2 \\
    \leq & ~ R^2 m \sqrt{nd} \exp(R^3) \cdot \| x - y \|_2
\end{align*}
the initial step is given by the definition of $\alpha$ (see Definition~\ref{def:alpha}), the second step uses Cauchy-Schwartz inequality, the 3rd step is from Lemma~\ref{lem:lipschitz_u}.
\end{proof}
\end{lemma}

\subsection{Lipschitz of function \texorpdfstring{$\alpha(x)^{-1}$}{}}\label{sec:lipschitz_alpha_inverse}
In this section, we give the Lipschitz constant of $\alpha^{-1}(x)$.

\begin{lemma} \label{lem:lipschitz_alpha_inverse}
Suppose $\alpha(x)$ is defined as Definition~\ref{def:alpha}. We use $\beta$ to denote the greatest lower bound of $\alpha(x)$, then it holds that
\begin{align*}
    \| \alpha(x)^{-1} - \alpha(y)^{-1} \|_2 \leq R^2 \beta^{-2} m \sqrt{nd} \exp(R^3) \cdot \| x - y \|_2
\end{align*}
\begin{proof}
Using Lemma 7.2 from \cite{dls23}, we have
\begin{align*}
    | \alpha(x)^{-1} - \alpha(y)^{-1} | \leq \beta^{-2} | \alpha(x) - \alpha(y) |
\end{align*}
Combining with Lemma~\ref{lem:lipschitz_alpha} leads to the result.
\end{proof}
\end{lemma}

\subsection{Lipschitz of function \texorpdfstring{$f(x)$}{} and function \texorpdfstring{$c(x)$}{}}\label{sec:lipschitz_f_c}
We provide the Lipschitz constant of $f(x)$ and $c(x)$ in this section.

\begin{lemma} \label{lem:lipschitz_f_c}
Suppose $f(x), c(x)$ is defined as Definition~\ref{def:f} and Definition~\ref{def:c}. We use $\beta$ to denote the greatest lower bound of $\alpha(x)$, then it holds that
\begin{align*}
    \| f(x) - f(y) \|_2 \leq & ~ 2R^2 m^{1.5} \sqrt{nd} \exp(4 R^3) \cdot \| x - y \|_2 \\\\
    \| c(x) - c(y) \|_2 \leq & ~ 2R^2 m^{1.5} \sqrt{nd} \exp(4 R^3) \cdot \| x - y \|_2
\end{align*}
\begin{proof}
\begin{align*}
    \| f(x) - f(y) \|_2 = & ~ \| \alpha(x)^{-1}(x) \cdot u(x) - \alpha(y)^{-1} \cdot u(y) \|_2 \\
    \leq & ~ | \alpha(x)^{-1} | \cdot \| u(x) - u(y) \|_2 + | \alpha(x)^{-1} - \alpha(y)^{-1} | \cdot \| u(y) \|_2 \\
    \leq & ~ \beta^{-1} \cdot \| u(x) - u(y) \|_2 + | \alpha(x)^{-1} - \alpha(y)^{-1} | \cdot \| u(y) \|_2 \\
    \leq & ~ \beta^{-1} R^2 \sqrt{mnd} \exp(R^3) \cdot \| x - y \|_2 + | \alpha(x)^{-1} - \alpha(y)^{-1} | \cdot \| u(y) \|_2 \\
    \leq & ~ \beta^{-1} R^2 \sqrt{mnd} \exp(R^3) \cdot \| x - y \|_2 + \beta^{-2} R^2 m \sqrt{nd} \exp(R^3) \cdot \| x - y \|_2 \cdot \| u(y) \|_2 \\
    \leq & ~ \beta^{-1} R^2 \sqrt{mnd} \exp(R^3) \cdot \| x - y \|_2 + \beta^{-2} R^2 m^{1.5} \sqrt{nd} \exp(2 R^3) \cdot \| x - y \|_2 \\
    \leq & ~ 2R^2 m^{1.5} \sqrt{nd} \exp(4 R^3) \cdot \| x - y \|_2
\end{align*}
where the first step is given by the definition of $f(x)$ (see Definition~\ref{def:f}), the 2nd step is given by triangle inequality, the 3rd step is by the definition of $\beta$, the fourth step is due to Lemma~\ref{lem:lipschitz_u}, the 5th step uses Lemma~\ref{lem:lipschitz_alpha_inverse}, the 6th step is derived from Lemma~\ref{lem:upper_bound_u}, the last step is because of Lemma~\ref{lem:lower_bound_beta} and $R > 4$.

The Lipschitz number of $c(x)$ equals to that of $f(x)$ since they only differ by constant term.
\end{proof}
\end{lemma}

\section{Lipschitz of \texorpdfstring{$\nabla^2 L(x)$}{}} \label{sec:lipschitz_hess}

We prove the Lipschitz of $\nabla^2 L(x)$ under the condition that ${\bf 1}[A_1 x] = {\bf 1}[A_1 y]$, i.e., the ReLU function does not change state. We give the lipschitz constant for each term in $B(x)$ in section~\ref{sec:lipschitz_B}. We compute the Lipschitz constant of $\nabla^2 L(x)$ in Section~\ref{sec:lipschitz_hess_L}.

\paragraph{$Remark$} In this section we assume that
\begin{itemize}
    \item Let $A_1 \in \R^{n \times d}, A_2 \in \R^{m \times n}$ be matrices with $\| A_1 \| \leq R$, $\| A_2 \| \leq R$, $R>4$
    \item Let $x,y \in \R^d, b \in \R^m$ be vectors with $\| x \|_2 \leq R, \| y \|_2 \leq R, \|b\|_2 \leq 1$
    \item ${\bf 1}[A_1 x] = {\bf 1}[A_1 y]$
\end{itemize}
which is Assumption~\ref{ass:bounded_parameters} and Assumption~\ref{ass:fixed_ReLU} in Section~\ref{sec:cov_variable:informal}.

\subsection{Lipschitz of function \texorpdfstring{$B(x)$}{}}\label{sec:lipschitz_B}
We calculate the Lipschitz constant of $B(x)$ in this section.

\begin{lemma} \label{lem:lipschitz_B}
\begin{itemize}
Recall the definition for $B(x)$ in Section~\ref{sec:hess},
    \item Part 1
    \begin{align*}
        \| B_1(x) - B_1(y) \| \leq 8R^2 m^{1.5} \sqrt{nd} \exp(4 R^3) \cdot \| x - y \|_2
    \end{align*}

    \item Part 2
    \begin{align*}
        \| B_2(x) - B_2(y) \| \leq 16R^2 m^{1.5} \sqrt{nd} \exp(4 R^3) \cdot \| x - y \|_2
    \end{align*}

    \item Part 3
    \begin{align*}
        \| B_3(x) - B_3(y) \| \leq 16R^2 m^{1.5} \sqrt{nd} \exp(4 R^3) \cdot \| x - y \|_2
    \end{align*}

    \item Part 4
    \begin{align*}
        \| B_4(x) - B_4(y) \| \leq 36 R^2 m^{1.5} \sqrt{nd} \exp(4 R^3) \cdot \| x - y \|_2
    \end{align*}

    \item Part 5
    \begin{align*}
        \| B_5(x) - B_5(y) \| \leq  10R^2 m^{1.5} \sqrt{nd} \exp(4 R^3) \cdot \| x - y \|_2
    \end{align*}

    \item Part 6
    \begin{align*}
        \| B(x) - B(y) \| \leq  86 R^2 m^{1.5} \sqrt{nd} \exp(4 R^3) \cdot \| x - y \|_2
    \end{align*}
\end{itemize}

\begin{proof}
{\bf Part 1}
\begin{align*}
    \| B_1(x) - B_1(y) \|
    = & ~ \| \diag(f(x) \circ (f(x) + c(x))) - \diag(f(y) \circ (f(y) + c(y)) \| \\
    \leq & ~ \| f(x) \circ (f(x)+c(x)) - f(y) \circ (f(y)+c(y)) \|_2 \\
    \leq & ~ \| f(x) \circ (f(x)+c(x)) - f(x) \circ (f(y)+c(y)) \|_2 ~ + \\
    & ~ \| f(x) \circ (f(y)+c(y)) - f(y) \circ (f(y)+c(y)) \|_2 \\
    := & ~ D_1 + D_2
\end{align*}
where the first step is given by Lemma~\ref{lem:help}, the second step is given by Fact~\ref{fac:matrix_algebra}, the 3rd step uses triangle inequality.

For the 1st item $D_1$,
\begin{align*}
    D_1 = & ~ \| f(x) \circ (f(x)+c(x) - (f(y)+c(y))) \|_2 \\
    \leq & ~ \| f(x) \|_2 \cdot \| f(x)+c(x) - (f(y)+c(y)) \|_2 \\
    \leq & ~ \| f(x) \|_2 \cdot (\| f(x) - f(y) \|_2 + \| c(x) - c(y) \|_2) \\
    \leq & ~ \| f(x) - f(y) \|_2 + \| c(x) - c(y) \|_2 \\
    \leq & ~ 4R^2 m^{1.5} \sqrt{nd} \exp(4 R^3) \cdot \| x - y \|_2
\end{align*}
the first step is given by property of Hadamard product, the second step is by Fact~\ref{fac:vector_norm}, the third step uses triangle inequality, the fourth step uses Lemma~\ref{lem:upper_bound_f_c}, the fifth step is because of Lemma~\ref{lem:lipschitz_f_c}.

For the 2nd item $D_2$,
\begin{align*}
    D_2 = & ~ \| (f(x) - f(y)) \circ (f(y)+c(y)) \|_2 \\
    \leq & ~ \| f(x) - f(y) \|_2 \cdot \| f(y)+c(y) \|_2 \\
    \leq & ~ 2 \cdot \| f(x) - f(y) \|_2 \\
    \leq & ~ 4R^2 m^{1.5} \sqrt{nd} \exp(4 R^3) \cdot \| x - y \|_2
\end{align*}
where the first step is given by  property of Hadamard product, the second step is by Fact~\ref{fac:vector_norm}, the third step is given by  Lemma~\ref{lem:upper_bound_f_c}, the fourth step is because of Lemma~\ref{lem:lipschitz_f_c}.

Merging $D_1$ and $D_2$ yields
\begin{align*}
    \| B_1(x) - B_1(y) \| \leq & ~ D_1 + D_2 \\
    \leq & ~ 8R^2 m^{1.5} \exp(4 R^3) \cdot \| x - y \|_2
\end{align*}

{\bf Proof of Part 2}
\begin{align*}
     \| B_2(x) - B_2(y) \| = & ~ \| \diag(f(x)) \cdot (c(x)+f(x)) \cdot f(x)^\top - \diag(f(y)) \cdot (c(y)+f(y)) \cdot f(y)^\top\| \\
     \leq & ~ \| \diag(f(x)) \cdot (c(x)+f(x)) \cdot f(x)^\top - \diag(f(x)) \cdot (c(x)+f(x)) \cdot f(y)^\top\| ~ + \\
     & ~ \| \diag(f(x)) \cdot (c(x)+f(x)) \cdot f(y)^\top - \diag(f(x)) \cdot (c(y)+f(y)) \cdot f(y)^\top\| ~+ \\
     & ~ \| \diag(f(x)) \cdot (c(y)+f(y)) \cdot f(y)^\top - \diag(f(y)) \cdot (c(y)+f(y)) \cdot f(y)^\top\| \\
     := & ~ D_1 + D_2 + D_3
\end{align*}

For the 1st item $D_1$,
\begin{align*}
    D_1 = & ~ \| \diag(f(x)) \cdot (c(x)+f(x)) \cdot (f(x) - f(y))^\top \| \\
    \leq & ~ \| \diag(f(x)) \| \cdot \| c(x)+f(x) \|_2 \cdot \|f(x) - f(y) \|_2 \\
    \leq & ~ \| f(x) \|_2 \cdot \| c(x)+f(x) \|_2 \cdot \|f(x) - f(y) \|_2 \\
    \leq & ~ 3 \cdot \|f(x) - f(y) \|_2 \\
    \leq & ~ 6R^2 m^{1.5} \sqrt{nd} \exp(4 R^3) \cdot \| x - y \|_2
\end{align*}
where the 2nd step and the 3rd step use Fact~\ref{fac:matrix_algebra}, the 4th step is by Lemma~\ref{lem:upper_bound_f_c}, the 5th step uses Lemma~\ref{lem:lipschitz_f_c}.

For the 2nd item $D_2$,
\begin{align*}
    D_2 = & ~ \| \diag(f(x)) \cdot ((c(x)+f(x)) - (c(y) + f(y)) \cdot  f(y)^\top \| \\
    \leq & ~ \| \diag(f(x)) \| \cdot \| (c(x)+f(x)) - (c(y) + f(y))\|_2 \cdot \| f(y) \|_2 \\
    \leq & ~ \| f(x) \|_2 \cdot \| (c(x)+f(x)) - (c(y) + f(y))\|_2 \cdot \| f(y) \|_2 \\
    \leq & ~ \|(c(x)+f(x)) - (c(y) + f(y)) \|_2 \\
    \leq & ~ \| c(x)- c(y) \|_2 + \| f(x) - f(y) \|_2 \\
    \leq & ~ 4R^2 m^{1.5} \sqrt{nd} \exp(4 R^3) \cdot \| x - y \|_2
\end{align*}
where the 2nd step and the 3rd step use Fact~\ref{fac:matrix_algebra}, the 4th step is by Lemma~\ref{lem:upper_bound_f_c}, the 5th step is derived from triangle inequality, the 6th step uses Lemma~\ref{lem:lipschitz_f_c}.

For the 3rd item $D_3$,
\begin{align*}
    D_3 = & ~ \| \diag(f(x) - f(y)) \cdot (c(y)+f(y)) \cdot f(y)^\top \| \\
    \leq & ~ \| \diag(f(x) - f(y)) \| \cdot \| c(y)+f(y) \|_2 \cdot \| f(y) \|_2 \\
    \leq & ~ \| f(x) - f(y) \|_2 \cdot \| c(y)+f(y) \|_2 \cdot \| f(y) \|_2 \\
    \leq & ~ 3 \cdot \|f(x) - f(y) \|_2 \\
    \leq & ~ 6R^2 m^{1.5} \sqrt{nd} \exp(4 R^3) \cdot \| x - y \|_2
\end{align*}
where the 2nd step and the 3rd step use Fact~\ref{fac:matrix_algebra}, the 4th step is by Lemma~\ref{lem:upper_bound_f_c}, the 5th step uses Lemma~\ref{lem:lipschitz_f_c}.

Adding up $D_1$, $D_2$ and $D_3$, we have
\begin{align*}
    \| B_2(x) - B_2(y) \| \leq & ~ D_1 + D_2 + D_3 \\
    \leq & ~ 16R^2 m^{1.5} \sqrt{nd} \exp(4 R^3) \cdot \| x - y \|_2
\end{align*}

{\bf Proof of Part 3}

The result is obvious since $B_3(x)$ is the transpose of matrix $B_2(x)$.

{\bf Proof of Part 4}
\begin{align*}
    \| B_4(x) - B_4(y) \| \leq & ~ \| \langle 2c(x)+f(x), f(x) \rangle \cdot f(x) \cdot  f(x)^\top - \langle 2c(y)+f(y), f(y) \rangle \cdot f(y) \cdot  f(y)^\top \|\\
    \leq & ~ \| \langle 2c(x)+f(x), f(x) \rangle \cdot f(x) \cdot  f(x)^\top - \langle 2c(x)+f(x), f(x) \rangle \cdot f(x) \cdot  f(y)^\top \| ~+\\
    & ~ \| \langle 2c(x)+f(x), f(x) \rangle \cdot f(x) \cdot  f(y)^\top - \langle 2c(x)+f(x), f(x) \rangle \cdot f(y) \cdot  f(y)^\top \| ~+\\
    & ~ \| \langle 2c(x)+f(x), f(x) \rangle \cdot f(y) \cdot  f(y)^\top - \langle 2c(x)+f(x), f(y) \rangle \cdot f(y) \cdot  f(y)^\top \| ~+\\
    & ~ \| \langle 2c(x)+f(x), f(y) \rangle \cdot f(y) \cdot  f(y)^\top - \langle 2c(y)+f(y), f(y) \rangle \cdot f(y) \cdot  f(y)^\top \| \\
    := & ~ D_1 +D_2 +D_3 +D_4
\end{align*}
where the first step is given by Lemma~\ref{lem:help}, the second step is given by triangle inequality.

For the 1st term $D_1$, there is
\begin{align*}
    D_1 = & ~ \| \langle 2c(x)+f(x), f(x) \rangle \cdot f(x) \cdot  (f(x) - f(y))^\top \| \\
    \leq & ~ | \langle 2c(x)+f(x), f(x) \rangle | \cdot \| f(x) \|_2 \cdot \|f(x) - f(y) \|_2 \\
    \leq & ~ \| 2c(x)+f(x) \|_2 \cdot \| f(x) \|_2 \cdot \| f(x) \|_2 \cdot \|f(x) - f(y) \|_2 \\
    \leq & ~ 5 \cdot \|f(x) - f(y) \|_2 \\
    \leq & ~ 10 R^2 m^{1.5} \sqrt{nd} \exp(4 R^3) \cdot \| x - y \|_2
\end{align*}
where the 2nd step is from Fact~\ref{fac:matrix_algebra}, the 3rd step uses Cauchy-Schwartz inequality, the 4th step is by Lemma~\ref{lem:upper_bound_f_c}, the 5th step follows by Lemma~\ref{lem:lipschitz_f_c}.

For the second term $D_2$, there is
\begin{align*}
    D_2 = & ~ \| \langle 2c(x)+f(x), f(x) \rangle \cdot (f(x) - f(y)) \cdot  f(y)^\top \| \\
    \leq & ~ | \langle 2c(x)+f(x), f(x) \rangle | \cdot \| f(x) - f(y) \|_2 \cdot \|f(y) \|_2 \\
    \leq & ~ \| 2c(x)+f(x) \|_2 \cdot \| f(x) \|_2 \cdot \| f(x) - f(y) \|_2 \cdot \|f(y) \|_2 \\
    \leq & ~ 5 \cdot \|f(x) - f(y) \|_2 \\
    \leq & ~ 10 R^2 m^{1.5} \sqrt{nd} \exp(4 R^3) \cdot \| x - y \|_2
\end{align*}
where the 2nd step is from Fact~\ref{fac:matrix_algebra}, the 3rd step uses Cauchy-Schwartz inequality, the fourth step is by Lemma~\ref{lem:upper_bound_f_c}, the 5th step follows by Lemma~\ref{lem:lipschitz_f_c}.

For the 3rd term $D_3$, there is
\begin{align*}
    D_3 = & ~ \| \langle 2c(x)+f(x), f(x) - f(y) \rangle \cdot f(y) \cdot f(y)^\top \| \\
    \leq & ~ | \langle 2c(x)+f(x), f(x) - f(y) \rangle | \cdot \| f(y) \|_2 \cdot \| f(y) \|_2 \\
    \leq & ~ \| 2c(x)+f(x) \|_2 \cdot \| f(x) - f(y) \|_2 \cdot \| f(y) \|_2 \cdot \|f(y) \|_2 \\
    \leq & ~ 5 \cdot \|f(x) - f(y) \|_2 \\
    \leq & ~ 10 R^2 m^{1.5} \sqrt{nd} \exp(4 R^3) \cdot \| x - y \|_2
\end{align*}
where the 2nd step is from Fact~\ref{fac:matrix_algebra}, the 3rd step uses Cauchy-Schwartz inequality, the fourth step is by Lemma~\ref{lem:upper_bound_f_c}, the 5th step follows by Lemma~\ref{lem:lipschitz_f_c}.

For the last term $D_4$, there is
\begin{align*}
    D_4 = & ~ \| \langle 2c(x)+f(x) - (2c(y)+f(y)), f(y) \rangle \cdot f(y) \cdot f(y)^\top \| \\
    \leq & ~ | \langle 2c(x)+f(x) - (2c(y)+f(y)), f(y) \rangle | \cdot \| f(y) \|_2 \cdot \|f(y) \|_2 \\
    \leq & ~ \| 2c(x)+f(x) - (2c(y)+f(y)) \|_2 \cdot \| f(y) \|_2 \cdot \| f(y) \|_2 \cdot \| f(y) \|_2 \\
    \leq & ~ \| 2c(x)+f(x)- (2c(y)+f(y)) \|_2 \\
    \leq & ~ 2 \cdot \| c(x) - c(y) \|_2 + \| f(x) - f(y) \|_2 \\
    \leq & ~ 6 R^2 m^{1.5} \sqrt{nd} \exp(4 R^3) \cdot \| x - y \|_2
\end{align*}
where the 2nd step is from Fact~\ref{fac:matrix_algebra}, the 3rd step uses Cauchy-Schwartz inequality, the fourth step is by Lemma~\ref{lem:upper_bound_f_c}, the 5th step follows by triangle inequality, the 6th step is due to Lemma~\ref{lem:lipschitz_f_c}.

Summing up $D_1$ to $D_4$, the result is
\begin{align*}
    \| B_4(x) - B_4(y) \| \leq & ~ D_1 + D_2 + D_3 + D_4 \\
    \leq & ~ 36 R^2 m^{1.5} \sqrt{nd} \exp(4 R^3) \cdot \| x - y \|_2
\end{align*}

{\bf Proof of Part 5}
\begin{align*}
    \| B_5(x) - B_5(y) \| = & ~ \| \langle c(x), f(x) \rangle \cdot \diag(f(x)) - \langle c(y), f(y) \rangle \cdot \diag(f(y)) \| \\
    \leq & ~ \| \langle c(x), f(x) \rangle \cdot \diag(f(x)) - \langle c(x), f(x) \rangle \cdot \diag(f(y)) \| ~+ \\
    & ~ \| \langle c(x), f(x) \rangle \cdot \diag(f(y)) - \langle c(x), f(y) \rangle \cdot \diag(f(y)) \| ~+ \\
    & ~ \| \langle c(x), f(y) \rangle \cdot \diag(f(y)) - \langle c(y), f(y) \rangle \cdot \diag(f(y)) \| \\
    := & ~ D_1 + D_2 + D_3
\end{align*}

For the 1st item $D_1$,
\begin{align*}
    D_1 = & ~ \| \langle c(x), f(x) \rangle \cdot \diag(f(x)-f(y)) \| \\
    = & ~ | \langle c(x), f(x) \rangle | \cdot \| \diag(f(x)-f(y)) \| \\
    \leq & ~ \| c(x) \|_2 \cdot \| f(x) \|_2 \cdot \| \diag(f(x)-f(y)) \| \\
    \leq & ~ \| c(x) \|_2 \cdot \| f(x) \|_2 \cdot \| f(x)-f(y) \|_2 \\
    \leq & ~ 2 \cdot \| f(x) - f(y) \|_2 \\
    \leq & ~ 4R^2 m^{1.5} \sqrt{nd} \exp(4 R^3) \cdot \| x - y \|_2
\end{align*}
the initial step and the 2nd step are some rearrangement, the 3rd step uses Cauchy-Schwartz inequality, the fourth step is by Fact~\ref{fac:matrix_algebra}, the 5th step is given by Lemma~\ref{lem:upper_bound_f_c}, the 6th step is derived from Lemma~\ref{lem:lipschitz_f_c}.

For the second item $D_2$, we have
\begin{align*}
    D_2 = & ~ \| \langle c(x), f(x) - f(y) \rangle \cdot \diag(f(y)) \| \\
    = & ~ | \langle c(x), f(x) - f(y) \rangle | \cdot \| \diag(f(x)-f(y)) \| \\
    \leq & ~ \| c(x) \|_2 \cdot \| f(x) - f(y) \|_2 \cdot \| \diag(f(y)) \| \\
    \leq & ~ \| c(x) \|_2 \cdot \| f(x)-f(y) \|_2 \cdot \| f(y) \|_2 \\
    \leq & ~ 2 \cdot \| f(x) - f(y) \|_2 \\
    \leq & ~ 4R^2 m^{1.5} \sqrt{nd} \exp(4 R^3) \cdot \| x - y \|_2
\end{align*}
the first step and the 2nd step are some rearrangement, the 3rd step uses Cauchy-Schwartz inequality, the fourth step is by Fact~\ref{fac:matrix_algebra}, the 5th step is given by Lemma~\ref{lem:upper_bound_f_c}, the 6th step is derived from Lemma~\ref{lem:lipschitz_f_c}.

For the first item $D_3$, we have
\begin{align*}
    D_3 = & ~ \| \langle c(x) - c(y), f(y) \rangle \cdot \diag(f(y)) \| \\
    = & ~ | \langle c(x) - c(y), f(y) \rangle | \cdot \| \diag(f(y)) \| \\
    \leq & ~ \| c(x) - c(y) \|_2 \cdot \| f(y) \|_2 \cdot \| \diag(f(y)) \| \\
    \leq & ~ \| c(x) - c(y) \|_2 \cdot \| f(y) \|_2 \cdot \| f(y) \|_2 \\
    \leq & ~ \| c(x) - c(y) \|_2 \\
    \leq & ~ 2R^2 m^{1.5} \sqrt{nd} \exp(4 R^3) \cdot \| x - y \|_2
\end{align*}
the first and the 2nd step are some rearrangement, the 3rd step uses Cauchy-Schwartz inequality, the fourth step is by  Fact~\ref{fac:matrix_algebra}, the 5th step is given by Lemma~\ref{lem:upper_bound_f_c}, the 6th step is derived from Lemma~\ref{lem:lipschitz_f_c}.

Merging $D_1$, $D_2$, and $D_3$ yields
\begin{align*}
    \| B_5(x) - B_5(y) \| \leq & D_1 + D_2 + D_3 \\
    \leq & ~ 10R^2 m^{1.5} \sqrt{nd} \exp(4 R^3) \cdot \| x - y \|_2
\end{align*}

{\bf Proof of Part 6}
\begin{align*}
    \| B_(x) - B_(y) \| \leq & ~ \| B_1(x) - B_1(y) \|+ \| B_2(x) - B_2(y) \| + \| B_3(x) - B_3(y) \| ~ + \\
    & ~ \| B_4(x) - B_4(y) \| + \| B_5(x) - B_5(y) \| \\
    \leq & ~ 86R^2 m^{1.5} \sqrt{nd} \exp(4 R^3) \cdot \| x - y \|_2
\end{align*}
\end{proof}
\end{lemma}

\subsection{Lipschitz of function \texorpdfstring{$\nabla^2 L(x)$}{}} \label{sec:lipschitz_hess_L}
We state the Lipschitz constant of $\nabla^2 L(x)$ below.

\begin{lemma} \label{lem:lipschitz_hessian_L}
Let $M = m^{1.5} \sqrt{nd} \exp(5 R^3)$,
\begin{align*}
    \| \nabla^2 L(x) - \nabla^2 L(y) \| \leq M \cdot \| x - y \|_2
\end{align*}

\begin{proof}
\begin{align} \label{eq:hessian_each_entry}
    \| \nabla^2 L(x) - \nabla^2 L(y) \| = & ~ \| A_1^\top \cdot \diag({\bf 1} [A_1 x]) \cdot A_2^\top \cdot  (B(x) - B(y))  \cdot A_2 \cdot \diag({\bf 1}[A_1 x]) \cdot A_1 \| \notag \\
    \leq & ~ \| A_1^\top \| \cdot \| \diag({\bf 1} [A_1 x]) \|^2 \cdot \| A_2^\top \| \cdot \| B(x) - B(y) \| \cdot \| A_2 \| \cdot \| A_1 \| \notag \\
    = & ~ \| A_1^\top \| \cdot \| \diag({\bf 1} [A_1 x]) \|^2 \cdot \| B(x) - B(y) \| \cdot \| A_2 \|^2 \cdot \| A_1 \| \notag \\
    \leq & ~ \| A_1^\top \| \cdot \| {\bf 1} [A_1 x] \|_\infty^2 \cdot \| B(x) - B(y) \| \cdot \| A_2 \|^2 \cdot \| A_1 \| \notag \\
    \leq & ~ \| A_1^\top \| \cdot \| B(x) - B(y) \| \cdot \| A_2 \|^2 \cdot \| A_1 \| \notag \\
    \leq & ~ R^4 \cdot \| B(x) - B(y) \| \notag \\
    \leq & ~ 86R^6 m^{1.5} \sqrt{nd} \exp(4 R^3) \cdot \| x - y \|_2 \notag \\
    \leq & ~ m^{1.5} \sqrt{nd} \exp(5 R^3) \cdot \| x - y \|_2
\end{align}
where the 1st step is from Lemma~\ref{lem:decomposition_L}, the 2nd step is due to Fact~\ref{fac:matrix_algebra}, the 3rd step is because $\| A \| = \| A^\top \|$, the 4th step is given by Fact~\ref{fac:matrix_algebra}, the 5th step holds since $\| {\bf 1}[A_1 x] \|_\infty \leq 1$, the 6th step is because $\| A_1 \| \leq R$, the 7th step is due to Lemma~\ref{lem:lipschitz_B}, the 8th step holds since $R > 4$.
\end{proof}
\end{lemma}

%% file: psd.tex
\section{PD Lower Bound/Strongly Convex} \label{sec:psd}
We need the objective function to be PD so that we can guarantee the correctness of the approximation Newton method. Therefore, in this section, we introduce a regularization term to achieve the PD property. We give the PSD bounds for $B(x)$ in Section~\ref{sec:psd_B}. In Section~\ref{sec:R}, we describe the regularization term. In Section~\ref{sec:psd_L_reg}, we prove the PD property of $\nabla^2 L_{\mathrm{reg}}(x)$.

\subsection{PSD Bounds for \texorpdfstring{$B(x)$}{}} \label{sec:psd_B}
In this section, we prove the PSD bounds for $B(x)$.

\begin{lemma} \label{lem:psd_B}
Under following conditions,
\begin{itemize}
    \item Suppose $f(x), c(x)$ are defined as Definition~\ref{def:f} and Definition~\ref{def:c}
    \item Let $\| b \|_2 \leq 1$
    \item Let $B(x)$ be the function derived in Lemma~\ref{lem:decomposition_L}
\end{itemize}
we have,
\begin{itemize}
    \item Part 1 $-3 {\bf I}_m \preceq B_1(x) \preceq 3 {\bf I}_m$
    \item Part 2 $-10 {\bf I}_m \preceq B_2(x) + B_3(x) \preceq 10 {\bf I}_m$
    \item Part 3 $-5 {\bf I}_m \preceq B_4(x) \preceq 5 {\bf I}_m$
    \item Part 4 $-2 {\bf I}_m \preceq B_5(x) \preceq 2 {\bf I}_m$
    \item Part 5 $-20 {\bf I}_m \preceq B(x) \preceq 20 {\bf I}_m$
\end{itemize}
\end{lemma}
    
\begin{proof}
{\bf Proof of Part 1}
\begin{align*}
    B_1(x) = & ~ \diag(f(x) \circ (f(x)+c(x))) \\
    \preceq & ~ \| f(x) \circ (c(x)+f(x)) \|_2 \cdot {\bf I}_m \\
    \preceq & ~ \| f(x) \|_2 \cdot \| c(x)+f(x) \|_2 \cdot {\bf I}_m \\
    \preceq & ~ 3 {\bf I}_m
\end{align*}
the initial step is by Fact~\ref{fac:psd}, the 2nd step is given by Fact~\ref{fac:vector_norm}, the third step follows by Lemma~\ref{lem:upper_bound_f_c}.

Symmetrically, we get the PSD lower bound for $B_1(x)$.

{\bf Proof of Part 2}
\begin{align*}
    B_2(x) + B_3(x) = & ~ \diag(f(x)) \cdot (c(x)+f(x)) \cdot f(x)^\top + f(x) \cdot (c(x) + f(x))^\top \cdot \diag(f(x)) \\
    \preceq & ~ \diag(f(x)) \cdot (c(x)+f(x)) \cdot (c(x) + f(x))^\top \cdot \diag(f(x)) + f(x) \cdot f(x)^\top \\
    \preceq & ~ \| \diag(f(x)) \cdot (c(x)+f(x)) \|_2^2 \cdot {\bf I}_m + \| f(x) \|_2^2 \cdot {\bf I}_m \\
    \preceq & ~ \| f(x) \|_2^2 \cdot \| c(x)+f(x) \|_2^2 \cdot {\bf I}_m + \| f(x) \|_2^2 \cdot {\bf I}_m \\
    \preceq & ~ 10{\bf I}_m
\end{align*}
where the 1st step and the 2nd step use Fact~\ref{fac:psd}, the third step holds because of Fact~\ref{fac:matrix_algebra}, the fourth step is due to Lemma~\ref{lem:upper_bound_f_c}.

Symmetrically, we get the PSD lower bound for $B_2(x) + B_3(x)$.

{\bf Proof of Part 3}
\begin{align*}
    B_4(x) = \langle 2c(x)+f(x), f(x) \rangle \cdot f(x) \cdot  f(x)^\top \preceq & ~ |\langle 2c(x)+f(x), f(x) \rangle| \cdot \| f(x) \|_2^2 \cdot {\bf I}_m \\
    \preceq & ~ \| 2c(x)+f(x) \|_2 \cdot \| f(x) \|_2 \cdot \| f(x) \|_2^2 \cdot {\bf I}_m \\
    \preceq & ~ 5 {\bf I}_m
\end{align*}
the initial step is by Fact~\ref{fac:psd}, the second step uses Cauchy-Schwartz inequality, the 3rd step is given by Lemma~\ref{lem:upper_bound_f_c}.

Symmetrically, we get the PSD lower bound for $B_4(x)$.

{\bf Proof of Part 4}
\begin{align*}
    B_5(x) = \langle c(x), f(x) \rangle \cdot \diag(f(x)) \preceq & ~ \langle c(x), f(x) \rangle \cdot \| f(x) \|_2 \cdot {\bf I}_m \\
    \preceq & ~ \| c(x) \|_2 \cdot \| f(x) \|_2 \cdot \| f(x) \|_2 \cdot {\bf I}_m \\
    \preceq & ~ 2 {\bf I}_m
\end{align*}
the initial step is due to Fact~\ref{fac:psd}, the second step is derived from Cauchy-Schwartz inequality, the 3rd step uses Lemma~\ref{lem:upper_bound_f_c}.

Symmetrically, we get the PSD lower bound for $B_5(x)$.

{\bf Proof of Part 5}

The result follows from $B(x) = B_1(x) - B_2(x) - B_3(x) + B_4(x) -B_5(x)$.

\end{proof}

\subsection{Regularization of Loss function} \label{sec:R}
In this section, we define the regularization term.
%\Zhao{Imagine $n \gg \max\{ m, d\}$}
\begin{definition} \label{def:R}
For $A_1 \in \R^{n \times d}, A_2 \in \R^{m \times n}, x \in \R^d$. Let 
\begin{align*}
C := \underbrace{ A_2 }_{m \times n} \cdot \diag({\bf 1}[A_1 x]) \cdot \underbrace{ A_1 }_{n \times d}.
\end{align*}

We define regularization function $R : \R^d \to R_{>0}$ as
\begin{align*}
    R(x) = \frac{1}{2} \cdot \| WCx \|_2^2
\end{align*}
where $W = \diag(w)$ for some vector $w \in \R^m$.
\end{definition}

We have Hessian of $R(x)$ to be as follows,
\begin{lemma} \label{lem:Hessian_R}
Under following definitions
\begin{itemize}
    \item Suppose $R(x)$ is defined as Definition~\ref{def:R}
    \item Suppose $W^2$ is the matrix which each diagonal entry is the square of that of $W$
\end{itemize}
we have
\begin{itemize}
    \item $\nabla R(x) = C^\top W^2 Cx$
    \item $\nabla^2 R(x) = C^\top W^2 C$
\end{itemize}
\end{lemma}

\begin{definition} \label{def:L_reg}
    We define $L_{\mathrm{reg}}(x) = L(x) + R(x)$. (See Definition~\ref{def:L} for $L(x)$, Definition~\ref{def:R} for $R(x)$.)
\end{definition}

\subsection{Hessian of \texorpdfstring{$L_{\mathrm{reg}}(x)$}{} is PD} \label{sec:psd_L_reg}
In this section, we give the PSD bounds for $L_{\mathrm{reg}}(x)$. An extra assumption about the rank of matrix $C$ is made to assist our discussion. Let the data setting satisfy $A_1, A_2$ have full rank and $n \gg \max\{ m, d\}$ in some sense. Then the rank of $C$ has little dependency on $\diag({\bf 1}[A_1 x])$ which implies $C$ only has non-zero singular value (see following for more rigorous proof). %\Zhao{We should be able to relax this assumption}

\begin{lemma} [$L_{\mathrm{reg}}(x)$ is PD] \label{lem:pd_L_reg}
Under following assumptions
\begin{itemize}
    \item Suppose $C, W, R(x), L_{\mathrm{reg}}(x)$ are defined as in Section~\ref{sec:R}
    \item Let $W^2$ be the matrix which each diagonal entry is the square of that of $W$
    \item Recall $A_1 \in \R^{n \times d}, A_2 \in \R^{m \times n}$. Let $A_1, A_2$ have full rank. Let $n \geq \xi \cdot \max\{m, d\}$, where $\xi > 1$.
    \item Let $\| {\bf 1}[A_1 x] \|_1 \geq \theta n$, where $1 > \theta > \frac{1}{\xi}$
    \item Let $l > 0$ be a scalar
    \item For all $i \in [m]$, $w_i^2 \geq 20 + \frac{l}{\sigma_{\min}(C)^2}$
\end{itemize}
we have 
\begin{align*}
    \nabla^2 L_{\mathrm{reg}}(x) \succeq l \cdot {\bf I}_d
\end{align*}

\begin{proof}
Let $D = B(x) + W^2$, so that $\nabla^2 L_{\mathrm{reg}}(x) = C^\top D C$ by Lemma~\ref{lem:decomposition_L} and Definition~\ref{def:L_reg}.

 Due to the 3rd and 4th lemma assumption, we have
 \begin{align*}
     \mathrm{Rank}(\diag({\bf 1}[A_1 x])) \geq \theta n \geq \theta\xi \cdot \max \{m,d \} > \frac{1}{\xi} \xi \max \{m,d \} = \max \{m,d \}
 \end{align*}
 
 Then, $\mathrm{Rank}(C) = \min \{m, \mathrm{Rank}(\diag({\bf 1}[A_1 x])), d \} = \min \{m,d \}$, i.e., $C$ has full rank. Therefore, $\sigma_{\min}(C)^2 > 0$, where $\sigma_{\min}(C)$ denotes the singular value of $C$ with smallest absolute value.

Then, we have
\begin{align*}
    D \succeq & ~ -20{\bf I}_n + w_{\min}^2 {\bf I}_n \\
    \succeq & ~ \frac{l}{\sigma_{\min}(C)^2} \cdot {\bf I}_n 
\end{align*}
where the 1st step is due to the PSD lower bound for $B(x)$ (see Lemma~\ref{lem:psd_B}), the 2nd step is from the assumption of $w$.

Since $D$ is PD, we have
\begin{align*}
    C^\top D C \succeq \frac{l}{\sigma_{\min}(C)^2} \cdot \sigma_{\min}(C)^2 \cdot {\bf I}_d  \succeq l \cdot {\bf I}_d
\end{align*}
This step can be easily verified by the definition of PD.

Therefore, $\nabla^2 L_{\mathrm{reg}}(x)$ is PD, and hence $L_{\mathrm{reg}}(x)$ is strongly convex.
\end{proof}
\end{lemma}

\iffalse
The assumption above may not hold when ${\bf 1}[A_1 x]$ has a zero term, but we can still prove the PSD of the hessian.

\begin{lemma} [$L_{\mathrm{reg}}(x)$ is PSD] \label{lem:psd_L_reg}
If the following conditions hold
\begin{itemize}
    \item Let $C, W, R(x), L_{\mathrm{reg}}(x)$ be defined as in Section~\ref{sec:R}
    \item Let $W^2$ be the matrix which each diagonal entry is the square of that of $W$
    \item For all $i \in [m]$, $w_i^2 \geq 20$
\end{itemize}
we have 
\begin{align*}
    \nabla^2 L_{\mathrm{reg}}(x) \succeq 0
\end{align*}
\begin{proof}
Similar as Lemma~\ref{lem:pd_L_reg}, we have
\begin{align*}
    D \succeq & ~ -20{\bf I}_n + w_{\min}^2 {\bf I}_n \\
    \succeq & ~ 0
\end{align*}

Since $D$ is PSD, $C^\top D C$ is also PSD. This can be simply verified by the definition of PSD.

Therefore $\nabla^2 L_{\mathrm{reg}}(x)$ is PSD, and hence $L_{\mathrm{reg}}(x)$ is convex.
\end{proof}
\end{lemma}

\fi

%% file: newton.tex
\section{Newton Method} \label{sec:newton}

We use an approximated Newton method to implement the Softmax-ReLU regression. In Section~\ref{sec:approx_psd}, we state the update rule with an approximation method. In Section~\ref{sec:main_recult}, we proof the convergence rate of the algorithm.

\subsection{Approximation of Hessian} \label{sec:approx_psd}

In this section, we introduce an approximation method for the Hessian of $L_{\mathrm{reg}}(x)$ for efficient calculation. See more detailed discussion in \cite{dsw22}. See a similar application in \cite{dls23}.

\begin{lemma} [Approximating PD matrix, Lemma 4.5 of \cite{dsw22}] \label{lem:approx_hessian}
Under the following conditions:
\begin{itemize}
\item Consider a constant precision parameter $\epsilon_0 = 0.01$
\item Let $C$ be a real matrix of size $m \times d$
\item Let $D$ be a positive definite matrix of size $m \times m$
\end{itemize}
there exists an algorithm that can be executed in time
\begin{align}
O(( \mathrm{nnz}(C) + d^\omega)\poly(\log(m/\delta)))
\end{align}
The algorithm outputs a sparse diagonal matrix $\Tilde{D}$ of size $m \times m$, and it satisfies the condition
\begin{align}
(1 - \epsilon_0) C^\top D C \preceq C^\top \Tilde{D} C \preceq (1 + \epsilon_0) C^\top D C
\end{align}
Here, $\omega$ represents the exponent of matrix multiplication, which is currently estimated to be $2.372$ \cite{wil12,lg14,aw21,dwz22}.
\end{lemma}

\begin{definition}[Update rule] \label{def:H_g}
Based on previous definitions, we have
\begin{itemize}
    \item Let $g(x) := \langle c(x_t), f(x_t) \circ (A_2 \cdot ({\bf 1}[A_1 x_t] \circ A_{1,i})) \rangle - \langle c(x_t), f(x_t) \rangle \cdot \langle f(x_t) , A_2 \cdot ({\bf 1}[A_1 x_t] \circ A_{1,i}) \rangle + A_1^\top W A_1 x_t$ which is the gradient of $L_\mathrm{reg}(x)$
    \item Let $H(x) := C ^\top D C$, which is the Hessian of $L_\mathrm{reg}(x)$
    \item We update $x_t$ by $x_{t+1} = x_t - \wt{H}(x_t)^{-1} \cdot g(x_t)$ where $\wt{H}$ is the matrix given by approximation in Lemma~\ref{lem:approx_hessian}.
\end{itemize}

\end{definition}

\subsection{Main result} \label{sec:main_recult}
We proof the convergence of the algorithm in this section.
\begin{theorem} \label{thm:main}
We have vector $x \in S := \{ u \in \R^d, \| u \|_2 \leq R \}$ (independent variable), vector $ b \in \R^m, w \in \R^m$ , matrix $A_1 \in \R^{n \times d}, A_2 \in \R^{m \times n}$.

If the following condition holds,
\begin{itemize}
    \item Suppose $C, W, W^2$, and $R(x)$ is defined as in Definition~\ref{def:R}
    \item Suppose $L_\mathrm{reg}(x)$ is defined as Definition~\ref{def:L_reg}
    \item We use $x^*$ to denote the optimal solution for $\min_{x \in S} L_\mathrm{reg}(x)$
    \item We assume $x^* \in S, \|x^* \|_2 \leq R - 0.1l/M$, $\nabla L(x^*) = {\bf 0}_d$
    \item Let $\| A_1 \| \leq R, \| A_2 \| \leq R, \| b \|_2 \leq 1$, $R > 4$
    \item Let ${\bf 1}[A_1 x]$ be the same for all $x \in S$
    \item Let $A_1, A_2$ have full rank. Let $n \geq \xi \cdot \max\{m, d\}$, where $\xi > 1$. Let $\| {\bf 1}[A_1 x] \|_1 \geq \theta n$, where $1 > \theta > \frac{1}{\xi}$
    \item Let $l > 0$ be a scalar. For all $i \in [n]$, $w_i^2 \geq 20 + \frac{l}{\sigma_{\min}(C)}$
    \item let $M = m^{1.5} \sqrt{nd}  \exp(5R^3)$
    \item We choose an initial point $x_0$ such that $M \cdot \|x_0 - x^*\|_2 \leq 0.1l$
\end{itemize}
Then for any accuracy parameter $\epsilon$ in the range of (0, 0.1) and a failure probability $\delta$ in the range of (0, 0.1), an algorithm (refer to Algorithm~\ref{alg:main_algorithm}) can be employed. This algorithm guarantees, with a probability of at least $1 - \delta$, that it will execute $T = O(\log(|x_0 - x^*|_2 / \epsilon))$ iterations and produce a vector $\Tilde{x} \in \mathbb{R}^d$ satisfying $|\Tilde{x} - x^*|_2 \leq \epsilon$.

The execution time for each iteration is
\begin{align*}
    O(( \mathrm{nnz}(C) + d^\omega)\poly(\log(m/\delta)))
\end{align*}
Here $\omega$ represents the exponent of matrix multiplication, which is currently estimated to be $\omega \approx 2.372$ \cite{wil12,lg14,aw21,dwz22}.

\begin{proof}
With Lemma~\ref{lem:lipschitz_hessian_L} (Lipschitz of Hessian) and Lemma~\ref{lem:pd_L_reg} (PD Hessian), we can apply Lemma 6.10 in \cite{lsz23} to have below shrinking lemma.
\begin{align*}
    \| x_k - x^* \|_2 \leq 0.4 \cdot \| x_{k-1} - x^* \|_2
\end{align*} 

Therefore, the distance between $x_k$ and $x^*$ is shrinking in each step, which ensures that $\|x_k\|_2 \leq R$ for every $k$. Hence, we can iterately apply this lemma.

After $T$ iterations, we reach
\begin{align*}
    \| x_T - x^* \|_2 \leq 0.4^T \cdot \| x_0 - x^* \|_2
\end{align*}
See Section 6 of \cite{lsz23} for detailed proof for the convergence rate.

By choosing proper $T$, we arrive at the convergence result. The cost of Approximation Hessian in each iteration is from Lemma~\ref{lem:approx_hessian}. 

\end{proof}
\end{theorem}

\begin{algorithm}[!ht]\caption{Approximation Newton method}\label{alg:main_algorithm}
\begin{algorithmic}[1]
\Procedure{IterativeSoftmaxReLURegression}{$x_0$} \Comment{Theorem~\ref{thm:main}} 
    \State
    We choose an initial point $x_0$ satifying Assumption~\ref{ass:good_initial_pt} (good initial point).
    \State
    Let $T = \eta^{-1} \log(\frac{\| x_0 - x^* \|_2}{\epsilon})$ execution time of the algorithm.
    \For{$t=0 \rightarrow T$}
        \State
        $g \leftarrow \nabla L_\mathrm{reg}(x)$
        \State
        $H \leftarrow \nabla^2 L_\mathrm{reg}(x)$
        \State
        $\wt{H} \leftarrow \mathrm{subsample}(H)$
        \Comment{Lemma~\ref{lem:approx_hessian}}
        \State
        $x_{t+1} \leftarrow x_t - \eta \cdot \wt{H}^{-1} g$
    \EndFor
    \State
    $\Tilde{x} \leftarrow x_{T+1}$
    \State
    \Return $\Tilde{x}$
\EndProcedure
\end{algorithmic}
\end{algorithm}

%% file: sophia_related.tex
\section{Convergence in Loss} \label{sec:cov_loss}
In this section, we relax the condition in Section~\ref{sec:newton} that ${\bf 1}[A_1 x]$ remains the same for all $x \in \R^d$. We modify the analysis of convergence in \cite{llh+23} (see Section~\ref{sec:preli_sophia}). Then we use it to prove the convergence of Newton method in the sense of accuracy of Loss function's value (see Section~\ref{sec:cov_sophia}).

\subsection{Preliminary} \label{sec:preli_sophia}

In this section, we exert some lemmas from \cite{llh+23} and we modify their statements and proofs. Let $L(x): \R^d \to \R$ be the loss function.
Let $x^*$ be the optimal solution for question $\min_{x \in \R^d} L(x)$. We define $L_{\min} := L(x^*)$.

\begin{assumption} \label{ass:L}
$L(x)$ is twice continuously differentiable almost everywhere and strictly convex.
\end{assumption}

\begin{assumption} \label{ass:lip_hess}
$\forall x, y \in \R^d$, we have $\| \nabla^2 L(x)^{-1} \nabla^2 L(y) \| \leq N$
\end{assumption}
If Assumption~\ref{ass:L} and Assumption~\ref{ass:lip_hess} hold, then we have following lemmas.

\begin{lemma}[Lemma E.4 in \cite{llh+23}] \label{lem:ode}
For any $z \in \R^d$, the following differential equation has at least one solution on interval $[0, 1]$:
\begin{align} \label{eq:ode}
    \frac{\d x(t)}{\d t} = & ~ -(\nabla^2 L(x(t)))^{-1} \nabla L(z), \\
    x(0) = & ~ z \notag
\end{align}
and the solution satisfies that $\nabla L(x(t)) = (1 - t) \nabla L(z)$ and $x(1) = x^*$
\end{lemma}

\begin{lemma}[Modified Lemma E.5 in \cite{llh+23}] \label{lem:bound_hess}
For any $z \in \R^d$, it holds that
\begin{align*}
    \frac{2}{N} \cdot (L(z) - L_{\min}) \leq \nabla L(z)^\top (\nabla^2 L(z))^{-1} \nabla L(z) \leq 2N \cdot (L(z) - L_{\min})
\end{align*}

\begin{proof}
Let $x(t)$ be the solution of Eq.~\eqref{eq:ode}. Note that by Lemma~\ref{lem:ode}, we have $x(0) = z$, $x(1) = x^*$.

By Assumption~\ref{ass:lip_hess}, $\forall t \in [0, 1]$, we have
\begin{align} \label{eq:psd_bound}
    N(\nabla^2 L(z))^{-1} \succeq \nabla^2 L(x(t))^{-1} \succeq \frac{1}{N}(\nabla^2 L(z))^{-1}
\end{align}

Using chain rule, we have
\begin{align} \label{eq:diff_Lxt}
    \frac{\d L(x(t))}{\d t} = & ~ (\nabla L(x(t))^\top \frac{\d x(t)}{\d t} \notag \\
    = & ~ - (\nabla L(x(t))^\top (\nabla^2 L(x(t)))^{-1} \nabla L(z)
\end{align}
where the 2nd step is due to Eq.~\eqref{eq:ode}.

Next, we have
\begin{align} \label{eq:int_Lxt}
    L(z) - L_{\min} = L(x(0)) - L(x(1)) = & ~ \int_0^1 (\nabla L(x(t))^\top (\nabla^2 L(x(t)))^{-1} \nabla L(z) \d t  \notag \\
    = & ~ \int_0^1 (1-t)(\nabla L(z))^\top (\nabla^2 L(x(t)))^{-1} \nabla L(z) \d t 
\end{align}
where the initial step follows by Eq.~\eqref{eq:diff_Lxt}, the second step uses Lemma~\ref{lem:ode}.

Inputting Eq.~\eqref{eq:psd_bound} into Eq.~\eqref{eq:int_Lxt}, we have
\begin{align*}
    \frac{1}{N} \cdot (L(z) - L_{\min}) \leq (\int_0^1 (1-t) \d t) \cdot \nabla L(z)^\top (\nabla^2 L(z))^{-1} \nabla L(z) \leq N \cdot (L(z) - L_{\min})
\end{align*}

This directly lead to the result since $\int_0^1 (1-t) \d t = 2$.
\end{proof}
\end{lemma}

\begin{lemma} [Modified Lemma E.10 in \cite{llh+23}] \label{lem:descent} We define the update rule as
\begin{align*}
    x_{\new} = x - \eta \cdot (\nabla^2 L(x))^{-1} \nabla L(x)
\end{align*}
then we have
\begin{align*}
    L(x_{\new}) - L(x) \leq -(\eta - \frac{N}{2} \eta^2) \cdot \nabla L(x) (\nabla^2 L(x))^{-1} \nabla L(x)
\end{align*}

\begin{proof}
We define $f(t) := L(tx_{\new} + (1-t)x)$ where $t \in \R$, $u := (\nabla^2 L(x))^{-1} \nabla L(x)$. Notice that $f(t) = L(x - \eta tu)$.

Utilizing chain rule, we have
\begin{align*}
    f'(t) = & ~ - \eta u^\top \nabla L(x - \eta tu) \\
    f''(t) = & ~ \eta^2 u^\top \nabla L^2 (x - \eta tu) u
\end{align*}

By Assumption~\ref{ass:lip_hess}, $f''(t) \leq N \cdot f''(0)$ for $\forall t \in [0,1]$. Therefore, we have
\begin{align*}
    L(x_{\new}) - L(x) = f(1) - f(0) = & ~ f'(0) + \int_{s=0}^1 \int_{t=0}^s f''(t) \d t \d s \\
    \leq & ~ f'(0) + \int_{s=0}^1 \int_{t=0}^s N \cdot f''(0) \d t \d s \\
    = & ~ f'(0) + \frac{N}{2} f''(0) \\
    = & ~ -(\eta - \frac{N}{2} \eta^2) \cdot \nabla L(x) (\nabla^2 L(x))^{-1} \nabla L(x)
\end{align*}
where the 1st step is derived by definition of integral, the 3rd step is integral calculation, the 4th step is by plugging in $f'(0)$ and $f''(0)$.
\end{proof}
\end{lemma}

\begin{lemma} [Modified Lemma E.11 in \cite{llh+23}] \label{lem:shrinking_loss}
We define the update rule as
\begin{align*}
    x_{t+1} = x_t - \eta \cdot (\nabla^2 L(x_t))^{-1} \nabla L(x_t)
\end{align*}
Then for $t, T \in \mathbb{N}$, $t < T$, we have
\begin{align*}
    L(x_T) - L_{\min} \leq (\eta^2 - \frac{2}{N} \eta + 1)^{T-t} \cdot (L(x_t) - L_{\min})
\end{align*}
If we take $\eta = \frac{1}{N}$, then
\begin{align*}
    L(x_T) - L_{\min} \leq (1 - \frac{1}{N^2})^{T-t} \cdot (L(x_t) - L_{\min})
\end{align*}

\begin{proof}
\begin{align*}
    L(x_{t+1}) - L(x_t) \leq & ~ -(\eta - \frac{N}{2} \eta^2) \cdot \nabla L(x_t) (\nabla^2 L(x_t))^{-1} \nabla L(x_t) \\
    \leq & ~ -\frac{2}{N} \cdot (\eta - \frac{N}{2} \eta^2) \cdot (L(x_t) - L_{\min}) \\
    = & ~ (\eta^2 - \frac{2}{N} \eta) \cdot (L(x_t) - L_{\min})
\end{align*}
where the 1st step is given by Lemma~\ref{lem:descent}, the second step uses Lemma~\ref{lem:bound_hess}.

By some rearrangement of above equation, we have
\begin{align*}
    L(x_{t+1}) - L_{\min} \leq (\eta^2 - \frac{2}{N} \eta + 1) \cdot (L(x_t) - L_{\min})
\end{align*}
It directly leads to the result.
\end{proof}
\end{lemma}

\subsection{Convergence of Softmax-ReLU regression algorithm} \label{sec:cov_sophia}
In this section, we use above lemmas to prove the convergence of Newton method.
\begin{lemma} [Upper bound for Hessian]
Under following assumption
\begin{itemize}
    \item Suppose $C, W, R(x), L_{\mathrm{reg}}(x)$ are defined as in Section~\ref{sec:R}
    \item Let $W^2$ be the matrix which each diagonal entry is the square of that of $W$
    \item Recall $A_1 \in \R^{n \times d}, A_2 \in \R^{m \times n}, b \in \R^m$. Let $A_1, A_2$ have full rank. Let $n \geq \xi \cdot \max\{m, d\}$, where $\xi > 1$
    \item Let $\|A_1\| \leq R$, $\|A_2\| \leq R$, $\|b\|_2 \leq 1$
    \item Let $\| {\bf 1}[A_1 x] \|_1 \geq \theta n$, where $1 > \theta > \frac{1}{\xi}$
    \item Let $f, l > 0$ be scalars
    \item For all $i \in [m]$, $f \geq w_i^2 \geq 20 + \frac{l}{\sigma_{\min}(C)^2}$
\end{itemize}
we have
\begin{itemize}
    \item Part 1 
    \begin{align*}
        \forall x \in \R^d, \|\nabla^2 L_\mathrm{reg}(x)^{-1}\| \leq \frac{1}{l}
    \end{align*}
    
    \item Part 2
    \begin{align*}
        \forall x \in \R^d, \| \nabla^2 L_\mathrm{reg}(x) \| \leq R^4 \cdot (16 + f)
    \end{align*}

    \item Part 3
    \begin{align*}
        \forall x, y \in \R^d, \| \nabla^2 L_\mathrm{reg}(x)^{-1} \nabla^2 L_\mathrm{reg}(y)\| \leq \frac{1}{l} \cdot R^4 \cdot (16 + f)
    \end{align*}
\end{itemize}

\begin{proof}
{\bf Proof of Part 1}

By Lemma~\ref{lem:pd_L_reg}, $\nabla^2 L_\mathrm{reg}(x) \succeq l \cdot {\bf I}_n$. Therefore, $\nabla^2 L_\mathrm{reg}(x) \preceq \frac{1}{l} \cdot {\bf I}_n$, which completes the proof.

{\bf Proof of Part 2}
\begin{align*}
     \| \nabla^2 L_\mathrm{reg}(x) \| = \| C^\top (B(x)+W^2)C \| \leq & ~ \| C \|^2 \cdot (\| B(x) \| + \| W^2 \|) \\
     = & ~ \| A_2 \cdot \diag({\bf 1}[A_1 x]) \cdot A_1 \|^2 \cdot (\| B(x) \| + \| W^2 \|) \\
     \leq & ~ \| A_2 \| \cdot \| \diag({\bf 1}[A_1 x]) \| \cdot \| A_1 \|^2 \cdot (\| B(x) \| + \| W^2 \|) \\
     \leq & ~ (\| A_2 \| \cdot \| {\bf 1}[A_1 x] \|_\infty \cdot \| A_1 \|)^2 \cdot (\| B(x) \| + \| W^2 \|) \\
     \leq & ~ (\| A_2 \| \cdot \| A_1 \|)^2 \cdot (\| B(x) \| + \| W^2 \|) \\
     \leq & ~ R^4 \cdot (\| B(x) \| + \| W^2 \|) \\
     \leq & ~ R^4 \cdot (16 + \| W^2 \|) \\
     \leq & ~ R^4 \cdot (16 + f)
\end{align*}
where the first step is from Fact~\ref{fac:matrix_algebra}, the 2nd step is given by definition of $C$ (see Definition~\ref{def:R}), the 3rd step and the 4th step is from Fact~\ref{fac:matrix_algebra}, the 5th step holds because $\| {\bf 1}[A_1 x] \|_\infty \leq 1$, the 6th step is from lemma assumption, the 7th step is given by Lemma~\ref{lem:upper_bound_B}, the 8th step holds since $\forall i \in [m]$, $w_i^2 \leq f$.

{\bf Proof of Part 3}

It is proved by combining {\bf Part 1} and {\bf Part 2}.

\end{proof}
\end{lemma}

%\Shenghao{Do we need to add the approximation hessian in Section~\ref{sec:approx_psd} to following lemma?}
\begin{theorem} [Convergence of Newton method] \label{lem:cov_L}
Under following conditions
\begin{itemize}
    \item Suppose $C, W, R(x), L_{\mathrm{reg}}(x)$ are defined as in Section~\ref{sec:R}
    \item Let $W^2$ be the matrix which each diagonal entry is the square of that of $W$
    \item Recall $A_1 \in \R^{n \times d}, A_2 \in \R^{m \times n}, b \in \R^m$. Let $A_1, A_2$ have full rank. Let $n \geq \xi \cdot \max\{m, d\}$, where $\xi > 1$
    \item Let $\|A_1\| \leq R$, $\|A_2\| \leq R$, $\|b\|_2 \leq 1$
    \item Let $\| {\bf 1}[A_1 x] \|_1 \geq \theta n$, where $1 > \theta > \frac{1}{\xi}$
    \item Let $f, l > 0$ be scalars
    \item For all $i \in [m]$, $f \geq w_i^2 \geq 20 + \frac{l}{\sigma_{\min}(C)^2}$
    \item Let $x_0$ be the initial point of the algorithm. Let $x^*$ be the optimal point of $\min_{x \in \R^d} L_\mathrm{reg}(x)$, we define $L_{\min} := L_\mathrm{reg}(x^*)$
    \item we define the update rule as $x_{t+1} = x_t - \eta \cdot (\nabla^2 L(x_t))^{-1} \nabla L(x_t)$
    \item Let $N = \frac{1}{l}\cdot R^4 \cdot (16+f)$, we take $\eta = \frac{1}{N}$
\end{itemize}   
For any accuracy parameter $\epsilon > 0$, the algorithm outputs a vector $\wt{x} \in \R^d$ with $L_\mathrm{reg}(\wt{x}) - L_{\min} \leq \epsilon$ in $T = T = O(  N^2 \log( (L_{\mathrm{reg}}(x_0) - L_{\min}) / \epsilon) )$ iterations.

\begin{proof}
The proof is given by a combination of Lemma~\ref{lem:pd_L_reg}, Lemma~\ref{lem:bound_hess}, and Lemma~\ref{lem:shrinking_loss}.

{\bf Proof of strictly convex (Assumption~\ref{ass:L})}

This part follows by Lemma~\ref{lem:pd_L_reg}.

{\bf Proof of bounded Hessian (Assumption~\ref{ass:lip_hess})}

This part is from Lemma~\ref{lem:bound_hess}.

{\bf Proof of $T$ step shrinking loss}

Since Assumption~\ref{ass:L} and Assumption~\ref{ass:lip_hess} are satisfied, we can directly use Lemma~\ref{lem:shrinking_loss}:
\begin{align*}
    L_\mathrm{reg}(x_T) - L_{\min} \leq & ~ (1 - \frac{1}{N^2})^{T} \cdot (L_\mathrm{reg}(x_0) - L_{\min}) \\
    \leq & ~ \epsilon
\end{align*}
where last step follows from $T = O(  N^2 \log( (L_{\mathrm{reg}}(x_0) - L_{\min}) / \epsilon) )$

This completes the proof for convergent rate.
\end{proof}
\end{theorem}

%% file: main.bbl
\newcommand{\etalchar}[1]{$^{#1}$}
\begin{thebibliography}{ADH{\etalchar{+}}19b}

\bibitem[ADH{\etalchar{+}}19a]{ADH+19b}
Sanjeev Arora, Simon Du, Wei Hu, Zhiyuan Li, Ruslan Salakhutdinov, and Ruosong
  Wang.
\newblock On exact computation with an infinitely wide neural net.
\newblock {\em NeurIPS}, 2019.

\bibitem[ADH{\etalchar{+}}19b]{ADH+19a}
Sanjeev Arora, Simon Du, Wei Hu, Zhiyuan Li, and Ruosong Wang.
\newblock Fine-grained analysis of optimization and generalization for
  overparameterized two-layer neural networks.
\newblock {\em International Conference on Machine Learning}, 2019.

\bibitem[AG23]{ag23}
Sanjeev Arora and Anirudh Goyal.
\newblock A theory for emergence of complex skills in language models.
\newblock {\em arXiv preprint arXiv:2307.15936}, 2023.

\bibitem[ALS{\etalchar{+}}22]{als+22}
Josh Alman, Jiehao Liang, Zhao Song, Ruizhe Zhang, and Danyang Zhuo.
\newblock Bypass exponential time preprocessing: Fast neural network training
  via weight-data correlation preprocessing.
\newblock {\em arXiv preprint arXiv:2211.14227}, 2022.

\bibitem[Ans20]{ans00}
Kurt~M Anstreicher.
\newblock The volumetric barrier for semidefinite programming.
\newblock {\em Mathematics of Operations Research}, 2020.

\bibitem[AS23]{as23}
Josh Alman and Zhao Song.
\newblock Fast attention requires bounded entries.
\newblock {\em arXiv preprint arXiv:2302.13214}, 2023.

\bibitem[ASA{\etalchar{+}}22]{asa+22}
Ekin Akyürek, Dale Schuurmans, Jacob Andreas, Tengyu Ma, and Denny Zhou.
\newblock What learning algorithm is in-context learning? investigations with
  linear models.
\newblock {\em arXiv preprint arXiv:2211.15661}, 2022.

\bibitem[AZLS19a]{azls19a}
Zeyuan Allen-Zhu, Yuanzhi Li, and Zhao Song.
\newblock A convergence theory for deep learning via over-parameterization.
\newblock {\em Proceedings of the 36th International Conference on Machine
  Learning}, 2019.

\bibitem[AZLS19b]{azls19b}
Zeyuan Allen-Zhu, Yuanzhi Li, and Zhao Song.
\newblock On the convergence rate of training recurrent neural networks.
\newblock {\em NeurIPS}, 2019.

\bibitem[BCE{\etalchar{+}}23]{bce+23}
Sebastien Bubeck, Varun Chandrasekaran, Ronen Eldan, Johannes Gehrke, Eric
  Horvitz, Ece Kamar, Peter Lee, Yin~Tat Lee, Yuanzhi Li, Scott Lundberg,
  Harsha Nori, Hamid Palangi, Marco~Tulio Ribeiro, and Yi~Zhang.
\newblock Sparks of artificial general intelligence: Early experiments with
  gpt-4.
\newblock {\em arXiv preprint arXiv:2305.12712}, 2023.

\bibitem[BFJ89]{b90}
John~S Bridle, Soulié F.F., and Hérault J.
\newblock Probabilistic interpretation of feedforward classification network
  outputs, with relationships to statistical pattern recognition.
  neurocomputing: Algorithms, architectures and applications.
\newblock {\em NATO ASI Series (Series F: Computer and Systems Sciences). 68.
  Berlin, Heidelberg: Springer. pp. 227–236}, 1989.

\bibitem[Bis06]{b06}
C.~M. Bishop.
\newblock Pattern recognition and machine learning.
\newblock {\em Springer}, 2006.

\bibitem[BMR{\etalchar{+}}20]{bmr+20}
Tom Brown, Benjamin Mann, Nick Ryder, Melanie Subbiah, Jared~D Kaplan, Prafulla
  Dhariwal, Arvind Neelakantan, Pranav Shyam, Girish Sastry, Amanda Askell,
  et~al.
\newblock Language models are few-shot learners.
\newblock {\em Advances in neural information processing systems},
  33:1877--1901, 2020.

\bibitem[Bol68]{b68}
Ludwig Boltzmann.
\newblock "studien über das gleichgewicht der lebendigen kraft zwischen
  bewegten materiellen punkten" [studies on the balance of living force between
  moving material points].
\newblock {\em Wiener Berichte. 58: 517–560}, 1868.

\bibitem[BPSW21]{BPSW21}
Jan van~den Brand, Binghui Peng, Zhao Song, and Omri Weinstein.
\newblock Training (over- parametrized) neural networks in near-linear time.
\newblock {\em 12th Innovations in Theoretical Computer Science Conference
  (ITCS)}, 2021.

\bibitem[Bra20]{Bra20}
Jan van~den Brand.
\newblock A deterministic linear program solver in current matrix
  multiplication time.
\newblock {\em Proceedings of the Fourteenth Annual ACM-SIAM Symposium on
  Discrete Algorithms (SODA)}, 2020.

\bibitem[BS23]{bs23}
Jan den~van Brand and Zhao Song.
\newblock An $\sqrt{n}$ passes streaming algorithm for solving bipartite
  matching exactly.
\newblock {\em Manuscript}, 2023.

\bibitem[BSF06]{bsf06}
Y.~Bengio, P.~Simard, and P.~Frasconi.
\newblock Learning deep architectures for ai.
\newblock {\em Foundations and Trends in Machine Learning, 2(1), 1-27}, 2006.

\bibitem[BSZ23]{bsz23}
Jan van~den Brand, Zhao Song, and Tianyi Zhou.
\newblock Algorithm and hardness for dynamic attention maintenance in large
  language models.
\newblock {\em arXiv preprint arXiv:2304.02207}, 2023.

\bibitem[CCLY19]{ccly19}
Michael~B Cohen, Ben Cousins, Yin~Tat Lee, and Xin Yang.
\newblock A near-optimal algorithm for approximating the john ellipsoid.
\newblock In {\em Conference on Learning Theory}, pages 849--873. PMLR, 2019.

\bibitem[CG19]{CG19}
Yuan Cao and Quanquan Gu.
\newblock Generalization bounds of stochastic gradient descent for wide and
  deep neural networks.
\newblock {\em NeurIPS}, 2019.

\bibitem[CGH{\etalchar{+}}19]{CGH+19}
Tianle Cai, Ruiqi Gao, Jikai Hou, Siyu Chen, Dong Wang, Di~He, Zhihua Zhang,
  and Liwei Wang.
\newblock Gram-gauss-newton method: Learning overparameterized neural networks
  for regression problems.
\newblock {\em arXiv preprint arXiv:1905.11675}, 2019.

\bibitem[Cha22]{cha22}
ChatGPT.
\newblock Optimizing language models for dialogue.
\newblock {\em OpenAI Blog}, November 2022.

\bibitem[CLMY21]{clmy21}
HanQin Cai, Yuchen Lou, Daniel Mckenzie, and Wotao Yin.
\newblock A zeroth-order block coordinate descent algorithm for huge-scale
  black-box optimization.
\newblock {\em arXiv preprint arXiv:2102.10707}, 2021.

\bibitem[CLP{\etalchar{+}}21]{clp+21}
Beidi Chen, Zichang Liu, Binghui Peng, Zhaozhuo Xu, Jonathan~Lingjie Li, Tri
  Dao, Zhao Song, Anshumali Shrivastava, and Re.Mongoose Christopher.
\newblock A learnable lsh framework for efficient neural network training.
\newblock {\em International Conference on Learning Representation}, 2021.

\bibitem[CLS19]{cls19}
Michael~B Cohen, Yin~Tat Lee, and Zhao Song.
\newblock Solving linear programs in the current matrix multiplication time.
\newblock {\em STOC}, 2019.

\bibitem[CND{\etalchar{+}}22]{cnd+22}
Aakanksha Chowdhery, Sharan Narang, Jacob Devlin, Maarten Bosma, Gaurav Mishra,
  Adam Roberts, Paul Barham, Hyung~Won Chung, Charles Sutton, Sebastian
  Gehrmann, et~al.
\newblock Palm: Scaling language modeling with pathways.
\newblock {\em arXiv preprint arXiv:2204.02311}, 2022.

\bibitem[DCLT18]{dclt18}
Jacob Devlin, Ming-Wei Chang, Kenton Lee, and Kristina Toutanova.
\newblock Bert: Pre-training of deep bidirectional transformers for language
  understanding.
\newblock {\em arXiv preprint arXiv:1810.04805}, 2018.

\bibitem[DLMS23]{dlms23}
Yichuan Deng, Zhihang Li, Sridhar Mahadevan, and Zhao Song.
\newblock Zero-th order algorithm for softmax attention optimization.
\newblock {\em arXiv preprint arXiv:2307.08352}, 2023.

\bibitem[DLS23]{dls23}
Yichuan Deng, Zhihang Li, and Zhao Song.
\newblock Attention scheme inspired softmax regression.
\newblock {\em arXiv preprint arXiv:2304.10411}, 2023.

\bibitem[DSW22]{dsw22}
Yichuan Deng, Zhao Song, and Omri Weinstein.
\newblock Discrepancy minimization in input sparsity time.
\newblock {\em arXiv preprint arXiv:2210.12468}, 2022.

\bibitem[DWZ22]{dwz22}
Ran Duan, Hongxun Wu, and Renfei Zhou.
\newblock Faster matrix multiplication via symmetric hashing.
\newblock {\em arXiv preprint arXiv:2210.10173}, 2022.

\bibitem[DZPS19]{DZPS19}
Simon~S Du, Xiyu Zhai, Barnabas Poczos, and Aarti Singh.
\newblock Gradient descent provably optimizes over-parameterized neural
  networks.
\newblock {\em ICLR}, 2019.

\bibitem[Fuk69]{f69}
Kunihiko Fukushima.
\newblock Visual feature extraction by a multilayered network of analog
  threshold elements.
\newblock {\em IEEE Transactions on Systems Science and Cybernetics},
  5(4):322--333, 1969.

\bibitem[Fuk80]{f80}
Kunihiko Fukushima.
\newblock Neocognitron: A self-organizing neural network model for a mechanism
  of pattern recognition unaffected by shift in position.
\newblock {\em Biological cybernetics}, 36(4):193--202, 1980.

\bibitem[GBB11]{gbb11}
Xavier Glorot, Antoine Bordes, and Yoshua Bengio.
\newblock Deep sparse rectifier neural networks.
\newblock In {\em Proceedings of the fourteenth international conference on
  artificial intelligence and statistics}, pages 315--323. JMLR Workshop and
  Conference Proceedings, 2011.

\bibitem[GBC16]{gbc16}
I.~Goodfellow, Y.~Bengio, and A.~Courville.
\newblock {\em Deep Learning}.
\newblock MIT Press, 2016.

\bibitem[Gib02]{g02}
Josiah~Willard Gibbs.
\newblock {\em Elementary Principles in Statistical Mechanics}.
\newblock Charles Scribner's Sons, 1902.

\bibitem[GMS23]{gms23}
Yeqi Gao, Sridhar Mahadevan, and Zhao Song.
\newblock An over-parametrized exponential regression.
\newblock {\em arXiv preprint arXiv:2303.16504}, 2023.

\bibitem[GQSW22]{gqsw22}
Yeqi Gao, Lianke Qin, Zhao Song, and Yitan Wang.
\newblock A sublinear adversarial training algorithm.
\newblock {\em arXiv preprint arXiv:2208.05395}, 2022.

\bibitem[GS22]{gs22}
Yuzhou Gu and Zhao Song.
\newblock A faster small treewidth sdp solver.
\newblock {\em arXiv preprint arXiv:2211.06033}, 2022.

\bibitem[GSX23]{gsx23}
Yeqi Gao, Zhao Song, and Shenghao Xie.
\newblock In-context learning for attention scheme: from single softmax
  regression to multiple softmax regression via a tensor trick.
\newblock {\em arXiv preprint arXiv:2307.02419}, 2023.

\bibitem[GSY23]{gsy23}
Yeqi Gao, Zhao Song, and Junze Yin.
\newblock An iterative algorithm for rescaled hyperbolic functions regression.
\newblock {\em arXiv preprint arXiv:2305.00660}, 2023.

\bibitem[GSYZ23]{gsyz23}
Yeqi Gao, Zhao Song, Xin Yang, and Ruizhe Zhang.
\newblock Fast quantum algorithm for attention computation.
\newblock {\em arXiv preprint arXiv:2307.08045}, 2023.

\bibitem[GSZ23]{gsz23}
Yuzhou Gu, Zhao Song, and Lichen Zhang.
\newblock A nearly-linear time algorithm for structured support vector
  machines.
\newblock {\em arXiv preprint arXiv:2307.07735}, 2023.

\bibitem[GTLV22]{gtlv22}
Shivam Garg, Dimitris Tsipras, Percy Liang, and Gregory Valiant.
\newblock What can transformers learn in-context? a case study of simple
  function classes.
\newblock {\em arXiv preprint arXiv:2208.01066}, 2022.

\bibitem[HJS{\etalchar{+}}22]{hjs+22}
Baihe Huang, Shunhua Jiang, Zhao Song, Runzhou Tao, and Ruizhe Zhang.
\newblock Solving sdp faster: A robust ipm framework and efficient
  implementation.
\newblock {\em 2022 IEEE 63rd Annual Symposium on Foundations of Computer
  Science (FOCS)}, 2022.

\bibitem[HLSY20]{HLSY21}
Baihe Huang, Xiaoxiao Li, Zhao Song, and Xin Yang.
\newblock Fl-ntk: A neural tangent kernel-based framework for federated
  learning convergence analysis.
\newblock {\em ICML}, 2020.

\bibitem[HS00]{hs00}
Richard Hahnloser and H~Sebastian Seung.
\newblock Permitted and forbidden sets in symmetric threshold-linear networks.
\newblock {\em Advances in neural information processing systems}, 13, 2000.

\bibitem[HSM{\etalchar{+}}00]{hsm+00}
Richard~HR Hahnloser, Rahul Sarpeshkar, Misha~A Mahowald, Rodney~J Douglas, and
  H~Sebastian Seung.
\newblock Digital selection and analogue amplification coexist in a
  cortex-inspired silicon circuit.
\newblock {\em nature}, 405(6789):947--951, 2000.

\bibitem[HWL21]{hwl21}
Weihua He, Yongyun Wu, and Xiaohua Li.
\newblock Attention mechanism for neural machine translation: A survey.
\newblock In {\em 2021 IEEE 5th Information Technology, Networking, Electronic
  and Automation Control Conference (ITNEC)}, volume~5, pages 1485--1489. IEEE,
  2021.

\bibitem[JGH18]{JGH18}
Arthur Jacot, Franck Gabriel, and Clement Hongler.
\newblock Neural tangent kernel: Convergence and generalization in neural
  networks.
\newblock {\em NeurIPS}, 2018.

\bibitem[JLSZ23]{jlsz23}
Haotian Jiang, Yin~Tat Lee, Zhao Song, and Lichen Zhang.
\newblock Convex minimization with integer minima in $\wt{O}(n^4)$ time.
\newblock {\em arXiv preprint arXiv:2304.03426, 2023}, 2023.

\bibitem[JSWZ21]{jswz21}
Shunhua Jiang, Zhao Song, Omri Weinstein, and Hengjie Zhang.
\newblock Faster dynamic matrix inverse for faster lps.
\newblock {\em STOC}, 2021.

\bibitem[JT20]{JT20}
Ziwei Ji and Matus Telgarsky.
\newblock Polylogarithmic width suffices for gradient descent to achieve
  arbitrarily small test error with shallow relu network.
\newblock {\em ICLR}, 2020.

\bibitem[KKL20]{kkl20}
Nikita Kitaev, Lukasz Kaiser, and Anselm Levskaya.
\newblock Reformer: The efficient transformer.
\newblock {\em arXiv preprint arXiv: 2001.04451}, 2020.

\bibitem[LBH15]{lbh15}
Y.~LeCun, Y.~Bengio, and G.~Hinton.
\newblock Deep learning.
\newblock {\em Nature, 521(7553), 436-444}, 2015.

\bibitem[LBOM02]{lbom02}
Yann LeCun, L{\'e}on Bottou, Genevieve~B Orr, and Klaus-Robert M{\"u}ller.
\newblock Efficient backprop.
\newblock In {\em Neural networks: Tricks of the trade}, pages 9--50. Springer,
  2002.

\bibitem[LG14]{lg14}
Francois Le~Gall.
\newblock Powers of tensors and fast matrix multiplication.
\newblock {\em Proceedings of 39th international symposium on symbolic and
  algebraic computation}, 2014.

\bibitem[LL18]{LL18}
Yuanzhi Li and Yingyu Liang.
\newblock Learning overparameterized neural networks via stochastic gradient
  descent on structured data.
\newblock {\em NeurIPS}, 2018.

\bibitem[LLH{\etalchar{+}}23]{llh+23}
Hong Liu, Zhiyuan Li, David Hall, Percy Liang, and Tengyu Ma.
\newblock Sophia: A scalable stochastic second-order optimizer for language
  model pre-training.
\newblock {\em arXiv preprint arXiv:2305.14342}, 2023.

\bibitem[LSS{\etalchar{+}}20]{LSS+20}
Jason~D Lee, Ruoqi Shen, Zhao Song, Mengdi Wang, and Zheng Yu.
\newblock Generalized leverage score sampling for neural networks.
\newblock {\em NeurIPS}, 2020.

\bibitem[LSZ19]{lsz19}
Yin~Tat Lee, Zhao Song, and Qiuyi Zhang.
\newblock Solving empirical risk minimization in the current matrix
  multiplication time.
\newblock {\em Conference on Learning Theory (COLT)}, 2019.

\bibitem[LSZ23a]{lsz23}
Zhihang Li, Zhao Song, and Tianyi Zhou.
\newblock Solving regularized exp, cosh and sinh regression problems.
\newblock {\em arXiv preprint, 2303.15725}, 2023.

\bibitem[LSZ{\etalchar{+}}23b]{lsz+23}
S~Cliff Liu, Zhao Song, Hengjie Zhang, Lichen Zhang, and Tianyi Zhou.
\newblock Space-efficient interior point method, with applications to linear
  programming and maximum weight bipartite matching.
\newblock In {\em ICALP}, 2023.

\bibitem[MGN{\etalchar{+}}23]{mgn+23}
Sadhika Malladi, Tianyu Gao, Eshaan Nichani, Alex Damian, Jason~D Lee, Danqi
  Chen, and Sanjeev Arora.
\newblock Fine-tuning language models with just forward passes.
\newblock {\em arXiv preprint arXiv:2305.17333}, 2023.

\bibitem[MWY{\etalchar{+}}23]{mwy+23}
Sadhika Malladi, Alexander Wettig, Dingli Yu, Danqi Chen, and Sanjeev Arora.
\newblock A kernel-based view of language model fine-tuning.
\newblock In {\em International Conference on Machine Learning}, pages
  23610--23641. PMLR, 2023.

\bibitem[OM20]{OS20}
Samet Oymak and Soltanolkotabi Mahdi.
\newblock Toward moderate overparameterization: Global convergence guarantees
  for training shallow neural networks.
\newblock {\em IEEE Journal on Selected Areas in Information Theory}, 2020.

\bibitem[Ope23]{o23}
OpenAI.
\newblock Gpt-4 technical report.
\newblock {\em arXiv preprint arXiv:2303.08774}, 2023.

\bibitem[PMXA23]{pmxa23}
Abhishek Panigrahi, Sadhika Malladi, Mengzhou Xia, and Sanjeev Arora.
\newblock Trainable transformer in transformer.
\newblock {\em arXiv preprint arXiv:2307.01189}, 2023.

\bibitem[PSZA23]{psza23}
Abhishek Panigrahi, Nikunj Saunshi, Haoyu Zhao, and Sanjeev Arora.
\newblock Task-specific skill localization in fine-tuned language models.
\newblock {\em arXiv preprint arXiv:2302.06600}, 2023.

\bibitem[QSY23]{qsy23}
Lianke Qin, Zhao Song, and Yuanyuan Yang.
\newblock Efficient sgd neural network training via sublinear activated neuron
  identification.
\newblock {\em arXiv preprint arXiv:2307.06565}, 2023.

\bibitem[QSZZ23]{qszz23}
Lianke Qin, Zhao Song, Lichen Zhang, and Danyang Zhuo.
\newblock An online and unified algorithm for projection matrix vector
  multiplication with application to empirical risk minimization.
\newblock In {\em International Conference on Artificial Intelligence and
  Statistics}, pages 101--156. PMLR, 2023.

\bibitem[RNS{\etalchar{+}}18]{rns+18}
Alec Radford, Karthik Narasimhan, Tim Salimans, Ilya Sutskever, et~al.
\newblock Improving language understanding by generative pre-training.
\newblock {\em .}, 2018.

\bibitem[RSM{\etalchar{+}}23]{rsm+23}
Rafael Rafailov, Archit Sharma, Eric Mitchell, Stefano Ermon, Christopher
  D.Manning, and Chelsea Finn.
\newblock Direct preference optimization: Your language model is secretly a
  reward model.
\newblock {\em arXiv preprint arXiv:2305.18290}, 2023.

\bibitem[RWC{\etalchar{+}}19]{rwc+19}
Alec Radford, Jeffrey Wu, Rewon Child, David Luan, Dario Amodei, Ilya
  Sutskever, et~al.
\newblock Language models are unsupervised multitask learners.
\newblock {\em OpenAI blog}, 1(8):9, 2019.

\bibitem[SHT23]{sht23}
Clayton Sanford, Daniel Hsu, and Telgarsky.
\newblock Representational strengths and limitations of transformers.
\newblock {\em arXiv preprint arXiv:2306.02896}, 2023.

\bibitem[Son19]{Son19}
Zhao Song.
\newblock Matrix theory: optimization, concentration, and algorithms.
\newblock {\em The University of Texas at Austin}, 2019.

\bibitem[SY19]{SY19}
Zhao Song and Xin Yang.
\newblock Quadratic suffices for over-parametrization via matrix chernoff
  bound.
\newblock {\em arXiv preprint arXiv:1906.03593}, 2019.

\bibitem[SY21]{sy21}
Zhao Song and Zheng Yu.
\newblock Oblivious sketching-based central path method for linear programming.
\newblock {\em ICML}, 2021.

\bibitem[SY23]{sy23}
Zhao Song and Mingquan Ye.
\newblock Efficient asynchronize stochastic gradient algorithm with structured
  data.
\newblock {\em arXiv preprint arXiv:2305.08001}, 2023.

\bibitem[SYYZ22]{syyz22}
Zhao Song, Xin Yang, Yuanyuan Yang, and Tianyi Zhou.
\newblock Faster algorithm for struc- tured john ellipsoid computation.
\newblock {\em arXiv preprint arXiv:2211.14407}, 2022.

\bibitem[SYYZ23]{syyz23}
Zhao Song, Xin Yang, Yuanyuan Yang, and Lichen Zhang.
\newblock Sketching meets differential privacy: Fast algorithm for dynamic
  kronecker projection maintenance.
\newblock In {\em International Conference on Machine Learning}, pages
  32418--32462. PMLR, 2023.

\bibitem[SYZ21]{syz21}
Zhao Song, Shuo Yang, and Ruizhe Zhang.
\newblock Does preprocessing help training over-parameterized neural networks?
\newblock {\em 35th Conference on Neural Information Processing Systems}, 2021.

\bibitem[SZKS21]{szks21}
Charlie Snell, Ruiqi Zhong, Dan Klein, and Jacob Steinhardt.
\newblock Approximating how single head attention learns.
\newblock {\em arXiv preprint arXiv:2103.07601}, 2021.

\bibitem[SZZ21]{SZZ21}
Zhao Song, Lichen Zhang, and Ruizhe Zhang.
\newblock Training multi-layer over-parametrized neural network in subquadratic
  time.
\newblock {\em arXiv preprint arXiv:2112.07628}, 2021.

\bibitem[TL59]{tl59}
A.W. Tucker and R.D. Luce.
\newblock {\em Contributions to the Theory of Games}.
\newblock Princeton University Press, 1959.

\bibitem[UAS{\etalchar{+}}20]{uas20}
Mohd Usama, Belal Ahmad, Enmin Song, M~Shamim Hossain, Mubarak Alrashoud, and
  Ghulam Muhammad.
\newblock Attention-based sentiment analysis using convolutional and recurrent
  neural network.
\newblock {\em Future Generation Computer Systems}, 113:571--578, 2020.

\bibitem[VSP{\etalchar{+}}17]{vsp+17}
Ashish Vaswani, Noam Shazeer, Niki Parmar, Jakob Uszkoreit, Llion Jones,
  Aidan~N Gomez, {\L}ukasz Kaiser, and Illia Polosukhin.
\newblock Attention is all you need.
\newblock {\em Advances in neural information processing systems}, 30, 2017.

\bibitem[WA21]{aw21}
Virginia~Wassilevska Williams and Josh Alman.
\newblock A refined laser method and faster matrix multiplication.
\newblock {\em Proceedings of the 2021 ACM-SIAM Symposium on Discrete
  Algorithms (SODA)}, 2021.

\bibitem[Wil12]{wil12}
Virginia~Wassilevska Williams.
\newblock Multiplying matrices faster than coppersmith-winograd.
\newblock {\em Proceedings of the forty-fouth annual ACM symposium on Theory of
  computing}, 2012.

\bibitem[WZW23]{wzw23}
Xinyi Wang, Wanrong Zhu, and William~Yang Wang.
\newblock Large language models are implicitly topic models: Explaning and
  finding good demonstrations for in-context learning.
\newblock {\em arXiv preprint arXiv:2301.11916}, 2023.

\bibitem[ZFB23]{zfb23}
Ruiqi Zhang, Spencer Frei, and Peter~L Bartlett.
\newblock Trained transformers learn linear models in-context.
\newblock {\em arXiv preprint arXiv:2306.09927}, 2023.

\bibitem[ZG19]{ZG19}
Difan Zou and Quanquan Gu.
\newblock An improved analysis of training over-parameterized deep neural
  networks.
\newblock {\em NeurIPS}, 2019.

\bibitem[Zha22]{Zha22}
Lichen Zhang.
\newblock Speeding up optimizations via data structures: Faster search, sample
  and maintenance.
\newblock {\em Master’s thesis, Carnegie Mellon University}, 2022.

\bibitem[ZHDK23]{zhdk23}
Amir Zandieh, Insu Han, Majid Daliri, and Amin Karbasi.
\newblock Kdeformer: Accelerating transformers via kernel density estimation.
\newblock {\em arXiv preprint arXiv: 2302.02451}, 2023.

\bibitem[ZMG19]{ZMG19}
Guodong Zhang, James Martens, and Roger~B Grosse.
\newblock Fast convergence of natu- ral gradient descent for over-parameterized
  neural networks.
\newblock {\em NeurIPS}, 2019.

\bibitem[ZPD{\etalchar{+}}20]{ZPD+20}
Yi~Zhang, Orestis Plevrakis, Simon~S Du, Xingguo Li, Zhao Song, and Sanjeev
  Arora.
\newblock Over-parameterized adversarial training: An analysis overcoming the
  curse of dimen- sionality.
\newblock {\em arXiv preprint arXiv:2002.06668}, 2020.

\bibitem[ZPGA23]{zpga23}
Haoyu Zhao, Abhishek Panigrahi, Rong Ge, and Sanjeev Arora.
\newblock Do transformers parse while predicting the masked word?
\newblock {\em arXiv preprint arXiv:2303.08117}, 2023.

\bibitem[ZRG{\etalchar{+}}22]{zrg+22}
Susan Zhang, Stephen Roller, Naman Goyal, Mikel Artetxe, Moya Chen, Shuohui
  Chen, Christopher Dewan, Mona Diab, Xian Li, Xi~Victoria Lin, et~al.
\newblock Opt: Open pre-trained transformer language models.
\newblock {\em arXiv preprint arXiv:2205.01068}, 2022.

\end{thebibliography}
